\documentclass[twoside,11pt]{article}

\usepackage{jmlrclean}

\usepackage{amsmath}
\usepackage{bm}
\usepackage{epsfig}
\usepackage{hhline}
\usepackage{multirow}
\usepackage{rotating}
\usepackage{subfigure}

\def\tr{\mathop{\rm tr}}

\def\EP{\mathrm{EP}}
\def\MCMC{\mathrm{MCMC}}

\def\erm{\mathrm{e}}

\def\facpower{D}

\def\balpha{\bm{\alpha}}
\def\bbeta{\bm{\beta}}
\def\bgamma{\bm{\gamma}}

\def\blambda{\bm{\lambda}}
\def\bLambda{\bm{\Lambda}}
\def\bmu{\bm{\mu}}
\def\bpi{\bm{\pi}}
\def\bSigma{\bm{\Sigma}}
\def\btau{\bm{\tau}}
\def\btheta{\bm{\theta}}

\def\0{\mathbf{0}}

\def\D{\mathbf{D}}
\def\I{\mathbf{I}}
\def\J{\mathbf{J}}
\def\k{\mathbf{k}}
\def\K{\mathbf{K}}

\def\s{\mathbf{s}}

\def\x{\mathbf{x}}
\def\y{\mathbf{y}}
\def\z{\mathbf{z}}
\def\beq{\begin{equation}}
\def\eeq{\end{equation}}
\def\beqa{\begin{eqnarray}}
\def\eeqa{\end{eqnarray}}
\def\beqas{\begin{eqnarray}}
\def\eeqas{\end{eqnarray}}
\def\beqas{\begin{eqnarray*}}
\def\eeqas{\end{eqnarray*}}

\def\cov{\mathop{\rm cov}}
\def\<{ \left\langle }
\def\>{ \right\rangle }

\def\Ncal{\mathcal{N}}

\def\wo{\backslash}

\def\oneh{\frac{1}{2}}

\newcommand{\E}[1]{{ \langle {#1} \rangle  }}
\newcommand{\bigE}[1]{{ \big\langle {#1} \big\rangle }}
\newcommand{\BigE}[1]{{ \Big\langle {#1} \Big\rangle }}

\jmlrheading{14}{2013}{2729-2770}{1/13; Revised 7/13}{9/13}{Manfred Opper, Ulrich Paquet and Ole Winther}

\ShortHeadings{Perturbative Corrections for Approximate Inference}{Opper, Paquet and Winther}
\firstpageno{1}

\begin{document}

\title{Perturbative Corrections for Approximate Inference in Gaussian Latent Variable Models}

\author{\name Manfred Opper \email opperm@cs.tu-berlin.de \\
       \addr Department of Computer Science \\
       Technische Universit\"{a}t Berlin \\
       D-10587 Berlin, Germany
       \AND
	   \name Ulrich Paquet \email ulripa@microsoft.com \\
	   \addr Microsoft Research Cambridge \\
	   Cambridge CB1 2FB, United Kingdom
       \AND
       \name Ole Winther \email owi@imm.dtu.dk \\
       \addr Informatics and Mathematical Modelling\\
       Technical University of Denmark \\
       DK-2800 Lyngby, Denmark
}

\editor{Neil Lawrence}

\maketitle

\begin{abstract}%
Expectation Propagation (EP) provides a framework for approximate inference.
When the model under consideration is over a latent Gaussian field, with the approximation being Gaussian, we show how these approximations can systematically be corrected.
A perturbative expansion is made of the exact but intractable correction, and can be applied to the model's partition function and other moments of interest. The correction is expressed over the higher-order cumulants which are neglected by EP's local matching of moments.
Through the expansion, we see that EP is correct to first order. By considering higher orders, corrections of increasing polynomial complexity can be applied to the approximation.
The second order provides a correction in quadratic time, which we apply to an array of Gaussian process and Ising models.
The corrections generalize to arbitrarily complex approximating families, which we illustrate on tree-structured Ising model approximations. Furthermore, they provide a polynomial-time assessment of the approximation error.
We also provide both theoretical and practical insights on the exactness of the EP solution.
\end{abstract}


\section{Introduction}

Expectation Propagation (EP) \citep{Opper2000,Min01,MinkaPhD}
is part of a rich family of variational methods, which approximate
the sums and integrals required for exact probabilistic inference by an optimization problem. Variational methods are perfectly amenable to
probabilistic graphical models,
as the nature of the optimization problem often allows it to be distributed
across a graph. By relying on local computations on a graph,
inference in very large probabilistic models becomes feasible.

Being an approximation, some error may invariably be introduced.
This paper is specifically concerned with the error that arises when a Gaussian approximating family is used,
and lays a systematic foundation for examining and correcting these errors.
It follows on earlier work by the authors \citep{opper08improving}.
The error that arises when the free energy (the negative logarithm of the partition function or normalizer of the distribution) is approximated, may for instance be written as a Taylor expansion \citep{opper08improving,paquet09perturbation}. A pleasing property of EP is that, at its stationary point, the first order term of such an expansion is zero.
Furthermore, the quality of the approximation can then be ascertained in polynomial time by including corrections beyond the first order, or beyond the standard EP solution. In general, the corrections improve the approximation when they are comparatively small, but can also leave a question mark on the quality of approximation when the lower-order terms are large.

The approach outlined here is by no means unique in correcting the approximation, as is evinced by cluster-based expansions \citep{paquet09perturbation}, marginal corrections for EP \citep{CsekeHeskesJMLR2010} and the Laplace approximation \citep{rue2009approximate}, and corrections to Loopy Belief Propagation \citep{chertkov06loop,sudderth08loop,welling12cluster}.

\subsection{Overview}

EP is introduced in a general way in Section \ref{sec:ep}, making it clear how various degrees of complexity can be included in its approximating structure.
The partition function will be used throughout the paper to explain the necessary machinery for correcting any moments of interest.
In the experiments, corrections to the marginal and predictive means and variances are also shown, although the technical details for correcting moments beyond the partition function are relegated to Appendix \ref{sec:marginals}.
The Ising model, which is cast as a Gaussian latent variable model in Section \ref{sec:glvm}, will furthermore be used as a running example throughout the paper.

The key to obtaining a correction lies in isolating the ``intractable quantity'' from the ``tractable part'' (or EP solution) in the true problem. 
This is done by considering the cumulants of both: as EP locally matches lower-order cumulants like means and variances, the ``intractable part'' exists as an expression over the higher-order cumulants which are neglected by EP.
This process is outlined in Section \ref{sec:corrections}, which concludes with two useful results: a shift of the ``intractable part'' to be an average over complex Gaussian variables with \emph{zero}
diagonal relation matrix, and Wick's theorem, which allows us to evaluate the expectations of polynomials under centered Gaussian measures. As a last stage, the ``intractable part'' is expanded in Sections \ref{sec:factorized} and \ref{sec:general-approximations} to obtain corrections to various orders.
In Section \ref{sec:convergence}, we provide a theoretical analysis of the radius of convergence of these expansions.

Experimental evidence is presented in Section \ref{sec:gpresults} on Gaussian process (GP) classification and (non-Gaussian) GP regression models. An insightful counterexample where EP diverges under increasing data, is also presented. Ising models are examined in Section \ref{sec:isingresults}.

Numerous additional examples, derivations, and material are provided in the appendices.
Details on different EP approximations can be found in Appendix \ref{app:gaussian-examples}, while corrections to tree-structured approximations are provided in Appendix \ref{app:tree}.
In Appendix \ref{sec:one-dim-example} we analytically show that the correction to a tractable example is zero. The main body of the paper deals with corrections to the partition function, while corrections to marginal moments are left to Appendix \ref{sec:marginals}. Finally, useful calculations of certain cumulants appear in Appendix \ref{app:cumulants}.

\section{Gaussian Latent Variable Models} \label{sec:glvm}

Let $\x = (x_1, \ldots, x_N)$
be an unobserved random variable with an intractable distribution $p(\x)$. In the Gaussian latent variable model (GLVM) considered in this paper,
terms $t_n(x_n)$ are combined
over a quadratic exponential $f_0(\x)$ to give
\begin{equation} \label{eq:mainquadratic}
p(\x) = \frac{1}{Z} \prod_{n=1}^N t_n (x_n) \, f_0(\x)
\end{equation}
with partition function (normalizer)
$$
Z = \int \prod_{n=1}^N t_n (x_n) \, f_0(\x) \, \mathrm{d}\x  \ .
$$
This model encapsulates many important methods used in statistical inference.
As an example, $f_0$ can encode the covariance matrix of a Gaussian process (GP) prior on latent function observations $x_n$. In the case of GP classification with a class label $y_n \in \{-1, +1\}$ on a latent function evaluation $x_n$,
the terms are typically probit link functions, for example
\begin{equation} \label{eq:gpcmodel}
p(\x) = \frac{1}{Z} \prod_{n=1}^N \Phi(y_n x_n) \, {\cal N}(\x \, ; \, \mathbf{0}, \, \K) \ .
\end{equation}
The probit function is the standard cumulative Gaussian density $\Phi(x) = \int_{-\infty}^{x} {\cal N}(z ; 0, 1) \, \mathrm{d} z$.
In this example, the partition function is not analytically tractable but for the one-dimensional case $N=1$.

An Ising model can be constructed by letting the terms $t_n$ restrict $x_n$ to $\pm 1$ (through Dirac delta functions).
By introducing the symmetric coupling matrix $\J$ and field $\btheta$
into $f_0$, an Ising model can be written as
\begin{equation} \label{eq:isingmodel}
p(\x) = \frac{1}{Z} \prod_{n=1}^N  \left[ \frac{1}{2} \delta(x_n + 1) + \frac{1}{2} \delta(x_n - 1) \right] \, \exp \left\{ \frac{1}{2} \x^T \J \x + \btheta^T \x \right\} \ .
\end{equation}
In the Ising model, the partition function $Z$ is intractable, as it sums $f_0(\x)$ over
$2^N$ binary values of $\x$.
In the variational approaches, the intractability is addressed by allowing approximations to $Z$ and other marginal distributions, decreasing the computational complexity from being exponential to polynomial in $N$, which is typically cubic for EP.

\section{Expectation Propagation} \label{sec:ep}

An approximation to $Z$ can be made by allowing
$p(\x)$ in Equation (\ref{eq:mainquadratic})
to factorize into a product of {\em factors} $f_a$.
This factorization is not unique, and the structure of the factorization of $p(\x)$ defines the complexity of the resulting approximation, resulting in different structures in the approximating distribution.
Where GLVMs are concerned, a natural and computationally convenient choice is to use Gaussian factors $g_a$, and as such, the approximating distribution $q(\x)$ in this paper will be Gaussian. Appendix \ref{app:gaussian-examples} summarizes a number of factorizations for Gaussian approximations.

The tractability of the resulting inference method imposes a pragmatic constraint on the choice of factorization; in the extreme case $p(\x)$ could be chosen as a single factor and inference would be exact.
For the model in Equation (\ref{eq:mainquadratic}),
a three-term product may be factorized as $(t_1) (t_2) (t_3)$, which gives the typical GP setup.
When a division is introduced and the term product factorizes as
$(t_1 t_2) (t_2 t_3) / (t_2)$, the resulting free energy will be that of the tree-structured EC approximation \citep{opper2005expectation}.
To therefore allow for regrouping, combining, splitting, and dividing terms,
a power $\facpower_a$ is associated with each $f_a$,
such that
\begin{equation} \label{eq:px}
p(\x) = \frac{1}{Z} \; \prod_a f_a(\x)^{\facpower_a}
\end{equation}
with intractable normalization (or partition function)
$Z = \int \prod_{a} f_{a}(\x)^{\facpower_a} \, \mathrm{d} \x$.\footnote{The 
factorization and EP energy function is expressed here in the form of Power EP \citep{minka04power}. 
}
Appendix \ref{app:gaussian-examples} shows how the
introduction of $\facpower_a$ lends itself to a clear definition of tree-structured and more complex approximations.

To define an approximation to $p$, terms $g_a$, which typically take an exponential family form, are chosen such that
\begin{equation} \label{eq:qx}
q(\x) = \frac{1}{Z_q} \; \prod_a g_a(\x)^{\facpower_a}
\end{equation}
has the same structure as $p$'s factorization.
Although not shown explicitly, $f_a$ and $g_a$ have a dependence on the \emph{same} subset of variables $\x_a$.
The optimal parameters of the $g_a$-term approximations
are found through a set of auxiliary
{\em tilted} distributions, defined by
\begin{equation} \label{eq:tilted}
q_a(\x) = \frac{1}{Z_a}\;  \left(\frac{q(\x) f_a(\x)}{g_a(\x)}\right) \ .
\end{equation}
Here a {\em single} approximating term $g_a$ is replaced by an original
term $f_a$. Assuming that this replacement leaves $q_a$ still tractable, the parameters in $g_a$ are determined by the condition
that $q(\x)$ and all $q_a(\x)$ should be made as similar as possible. This is usually achieved by requiring that these distributions share
a set of generalised moments which usually coincide with the sufficient statistics of the exponential family. For example with sufficient statistics $\phi(\x)$ we require that
\begin{equation} \label{eq:moments-match}
\< \phi(\x) \>_{q_a} = \< \phi(\x) \>_{q} \quad \textrm{for all} \ a \ .
\end{equation}
Note that those factors $f_a$ in $p(\x)$ which are already in the exponential family, such as the Gaussian terms in examples above, can trivially be solved for by setting $g_a = f_a$.
The partition function associated with this approximation is
\begin{equation} \label{eq:Z-EP}
Z_{\EP} = Z_q \; \prod_a Z_a^{\facpower_a} \ .
\end{equation}
Appendix \ref{sec:stationary-point} shows that the moment-matching conditions must hold at a
stationary point of $\log Z_{\EP}$.
The EP algorithm iteratively updates the $g_a$-terms by enforcing $q$ to share moments with each of the tilted distributions $q_a$; on reaching a fixed point all moments match according to Equation (\ref{eq:moments-match}) \citep{Min01,MinkaPhD}.
Although $Z_{\EP}$ is defined in the terminology of EP,
other algorithms may be required to solve for the fixed point, and $Z_{\EP}$, as a free energy, can be derived from the saddle point of a set of self-consistent (moment-matching) equations \citep{opper2005expectation,van2010efficient,seeger2010fast}.
We next make EP concrete by applying it to the Ising model, which will serve as a running example in the paper.
The section is finally concluded with a discussion of the interpretation of EP. 

\subsection{EP for Ising Models} \label{sec:EPIsing}

The Ising model in Equation (\ref{eq:isingmodel}) will be used as a running example throughout this paper. To make the technical developments more concrete, we will consider both the $N$-variate and bivariate cases. The bivariate case can be solved analytically, and thus allows for a direct comparison to be made between the exact and approximate solutions.

We use the factorized approximation as a running example, dividing $p(\x)$ in Equation (\ref{eq:isingmodel}) into $N+1$ factors with $f_0(\x)=\exp \{ \frac{1}{2} \x^T \J \x + \btheta^T \x \}$ and $f_n(x_n) = t_n(x_n)= \frac{1}{2} \delta(x_n + 1) + \frac{1}{2} \delta(x_n - 1) $, for $n=1,\ldots,N$ (see Appendix \ref{app:gaussian-examples} for generalizations). We consider the Gaussian exponential family such that $g_n(x_n)  = \exp\{ \lambda_{n1} x_n - \frac{1}{2} \lambda_{n2} x_n^2 \}$ and  $g_0(\x)=f_0(\x)$. The approximating distribution from Equation (\ref{eq:qx}), $q(\x) \propto f_0(\x) \prod_{n=1}^N g_n(x_n) $, is thus a \emph{full} multivariate Gaussian density, which we write as $q(\x) = {\cal N}(\x ; \bmu, \bSigma)$.

\subsubsection{Moment Matching}

The moment matching condition in Equation (\ref{eq:moments-match}) involves only the mean and variance if $q(\x)$ fully factorizes according to $p(\x)$'s terms. We therefore only need to match the mean and variances of marginals of $q(\x)$ and the tilted distribution $q_n(\x)$ in Equation (\ref{eq:tilted}).
The tilted distribution may be decomposed into a Gaussian and a discrete part as $q_n(\x)=q_n(\x_{\wo n}|x_n)q_n(x_n)$, where the vector $\x_{\wo n}$ consists of all variables apart from $x_n$. We may marginalize out $\x_{\wo n}$ and write $q_n(x_n)$ in terms of two factors:
\begin{equation}
q_n(x_n) \propto
\underbrace{ \frac{1}{2} \Big[ \delta(x_n + 1) + \delta(x_n - 1) \Big] }_{ f_n(\x) = t_n(x_n) }
\underbrace{ \exp \Big\{ \gamma x_n - \tfrac{1}{2} \Lambda x_n^2 \Big\} }_{\propto \ \int d \x_{\wo n} \, q(\x) / g_n(\x)} \ ,
\label{marginal:Ising}
\end{equation}
where we dropped the dependency of
$\gamma$ and  $\Lambda$ on $n$ for notational simplicity. Through some manipulation, the tilted distribution is equivalent to
\begin{equation} \label{eq:qtilt_n}
q_n(x_n) = \frac{1 + m_n}{2} \, \delta(x_n - 1) + \frac{1 - m_n}{2} \, \delta(x_n + 1) \ ,
\quad 
m_n = \tanh(\gamma) = \frac{\erm^\gamma - \erm^{-\gamma}}{\erm^\gamma + \erm^{-\gamma}} \ .
\end{equation}
This discrete distribution has mean $m_n$ and variance $1 - m_n^2$. By adapting the parameters of $g_n(x_n)$ using for example the EP algorithm, we aim to match the mean and variance of the marginal  $q(x_n)$ (of $q(\x)$) to the mean and variance of $q_n(x_n)$. The reader is referred to Section \ref{sec:isingresults} for benchmarked results for the Ising model.

\subsubsection{Analytic Bivariate Case}

Here we shall compare the exact result with EP and the correction for the simplest non-trivial model, the $N=2$ Ising model with no external field
$$
p(\x) = \frac{1}{4} \Big( \delta(x_1-1)+\delta(x_1+1)\Big) \Big( \delta(x_2-1)+\delta(x_2+1)\Big) \, \erm^{J x_1 x_2} \ .
$$
In order to solve the moment matching conditions we observe that the mean values must be zero because the distribution is symmetric around zero. Likewise the linear term in the approximating factors disappears and we can write
$g_n(x_n) = \exp\{-\lambda x_n^2/2\}$ and $q(\x) ={\cal N}\left(\x ; \mathbf{0}, \bSigma\right)$ with $\bSigma =  \left[ \begin{array}{cc}  \lambda & - J \\ -J & \lambda \end{array} \right]^{-1}$.
The moment matching condition for the variances, $1 = \Sigma_{nn}$, turns into a second order equation with solution $\lambda = \frac{1}{2} \left[ J^2 + \sqrt{J^4 + 4} \right]$. We can now insert this solution into the expression for the EP partition function in Equation (\ref{eq:Z-EP}).
By expanding the result to the second order in $J^2$, we find that
$$
\log Z_{\EP} = -\frac{1}{2} +  \frac{1}{2} \sqrt{1 + 4 J^2}
- \frac{1}{2} \log \left(\frac{1}{2}( 1+ \sqrt{1 + 4 J^2})\right) =
 \frac{J^2}{2} - \frac{J^4}{4} + \ldots \ .
$$
Comparing with the exact expression
$$
\log Z = \log \cosh(J) = \frac{J^2}{2} - \frac{J^4}{12} + \ldots
$$
we see that EP gives the correct $J^2$ coefficient, but the $J^4$ coefficient comes out wrong. In Section \ref{sec:corrections} we investigate how cumulant corrections can correct for this discrepancy.

\subsection{Two Explanations Why Gaussian EP is Often Very Accurate}

EP, as introduced above, is an algorithm. The justification for the algorithm put forward by Minka and adopted by others (see for example recent textbooks by \citealt{bishop06pattern}, \citealt{barberBRML2012} and \citealt{MurphyMLPP2012}) is useful for explaining the steps in the algorithm but may be misleading in order to
explain why EP often provides excellent accuracy in estimation of marginal moments and $Z$.

The general justification for EP  \citep{Min01,MinkaPhD} is based upon a minimization of Kullback-Leiber (KL) divergences. Ideally, one would determine the approximating distribution $q(\x)$ as the minimizer of  $\mathrm{KL}(p \|q)$ 
in an exponential family of (in our case, Gaussian) densities. Since this is not possible---it would require the computation 
of exact moments---we instead iteratively minimize ``local'' KL-divergences $\mathrm{KL}(q_a \| q)$, 
between the tilted distribution $q_a$ and $q$, with respect to $g_a$ (appearing in $q$). This leads to the moment matching conditions in Equation (\ref{eq:moments-match}). The argument for this procedure is essentially that this will ensure that 
the approximation $q$ will capture high density regions of the intractable posterior $p$. Obviously, this argument cannot be applied to Ising models because the exact and approximate distributions are very different,
with the former being discrete due to the Dirac $\delta$-functions that constrain $x_n = \pm 1$ to be binary variables. 
Even though the optimization still implies moment matching, this discrete-continuous discrepancy makes local KL-divergences $\mathrm{KL}(q_a \| q)$ {\em infinite}! 

In order to justify the usefulness of EP for Ising models we therefore need an alternative argument.  Our argument is entirely restricted to
{\em Gaussian} EP for our extended definition of GLVMs and do not extend to approximations 
with other exponential families.
In the following, we will discuss these assumptions in inference approximations 
that preceded the formulation of EP, in order to provide a possibly more relevant justification of the method. Although this justification is not strictly necessary for practically using EP nor corrections to EP, it nevertheless provides a good starting point for understanding both.  

The argument goes back to 
the  mathematical analysis of the Sherrington-Kirkpatrick (SK) model for a disordered magnet (a so-called spin glass) 
\citep{SKA}.  For this Ising model, the couplings $\J$ are drawn at random from a Gaussian distribution.
  An important contribution in the context of inference for this model
(the computations of partition functions and average magnetizations)  was the work of \citet{ThAnPa77} who derived  {\it 
self-consistency equations} 
which are assumed to be valid with a probability (with respect to the drawing of random couplings) approaching one  as 
the number of variables $x_n$ grows to infinity.
These so-called Thouless-Anderson-Palmer (TAP) equations are closely related to the EP moment matching conditions of Equation (\ref{eq:moments-match}), but they differ by partly relying on the specific assumption of the randomness of the couplings.
Self-consistency equations equivalent to the EP moment matching conditions which avoided such assumptions
on the statistics of the random couplings were first derived by \citet{Opper2000} by using a so-called cavity argument \citep{MePaVi87}. A new important contribution of \citet{Min01} was to provide an efficient algorithmic recipe for 
solving these equations.

We will now sketch the main idea of the cavity argument for the GLVM.
Let $\x_{\backslash n}$ (``$\x$ without $n$'') denote the complement to $x_n$, that is $\x = \x_{\backslash n} \cup x_{n}$. Without loss of generality we will take the quadratic exponential term to be written as $f_0(\x) \propto \exp( - \x^T \J \x / 2)$.
With similar definitions of $\J_{\backslash n}$, the 
exact marginal distribution of $x_n$ may be written as
\begin{align*}
p_n(x_n) & = \frac{1}{Z}  t_n(x_n) \int \exp \left\{ - \frac{1}{2} \x^T \J  \x \right\} \prod_{n'\neq n}  t_{n'}(x_{n'}) \, \mathrm{d}\x_{\backslash n} \\
& = \frac{ t_n(x_n)}{Z}  \, \mathrm{e}^{ - J_{nn}  \, x^2_n/2} \int \exp \left\{ -x_n \sum_{n'\neq n}  J_{n n'} x_{n'} - 
\frac{1}{2} \x_{\backslash n}^T \J_{\backslash n} \x_{\backslash n}  \right\} \prod_{n'\neq n}  t_{n'}(x_{n'}) \, \mathrm{d} \x_{\backslash n} \ .
\end{align*}
It is clear that $p_n(x_n)$ depends entirely on the statistics of the random variable $h_n \equiv \sum_{n'\neq n} J_{n n'} x_{n'}$. This 
is the total \emph{`field'} created by all other `magnetic moments' $x_{n'}$ in the `cavity' 
opened once $x_n$ has been removed from the system. 
In the context of densely connected models with weak couplings, we can appeal to the central limit 
theorem\footnote{In the context of sparsely connected models,
other cavity arguments lead to loopy belief propagation.}  to approximate $h_n$ by a Gaussian random variable with mean $\gamma_n$ and variance $V_n$. 
When looking at the influence  of the remaining variables $\x_{\backslash n}$ on $x_n$,
the non-Gaussian details of their distribution have been washed out in the marginalization. Integrating out the Gaussian
random variable $h_n$ gives the Gaussian cavity field approximation to the marginal distribution:
\begin{align*}
p_n(x_n) & \approx \mbox{const} \cdot t_n(x_n) \, \mathrm{e}^{ - J_{nn}  \, x^2_n/2}  \int \mathrm{e}^{- x_n h} \, {\cal N}(h \, ; \, \gamma_n, V_n)  \, \mathrm{d} h \\
& =  \mbox{const} \cdot t_n(x_n) \exp \left\{ -x_n \gamma_n - \oneh (J_{nn} - V_n) x_n^2\right\} \ .  
\end{align*}
This is precisely of the form of the marginal tilted distribution $q_n(x_n)$ of Equation (\ref{marginal:Ising}) 
as given by Gaussian EP. In 
the cavity formulation, $q(\x)$ is simply a \emph{placeholder} for the sufficient statistics of the individual Gaussian
cavity fields. 
So we may observe cases, with the Ising model or bounded support factors being the prime examples, where EP gives essentially correct results
for the marginal distributions of the $x_n$ and of the partition function $Z$, while 
$q(\x)$ gives a poor or even meaningless (in the sense of KL divergences) approximation to the 
multivariate posterior. 
Note however, that the entire {\em covariance matrix} of the $x_n$ can be computed simply from a derivative
of the free energy \citep{opper2005expectation} resulting in an approximation of this covariance by that of $q(\x)$. 
This may indicate that a good EP approximation of the free energy may also result in a good approximation
to the full covariance.
The near exactness of EP (as compared to exhaustive summation) in Section \ref{sec:isingresults} therefore shows the central limit theorem at work.  
Conversely, mediocre accuracy or even failure of Gaussian EP, as also observed in our simulations in Sections \ref{sec:gpbox} and \ref{sec:isingresults}, may be attributed to breakdown of the Gaussian cavity field assumption. Exact inference on the strongest couplings as considered for the Ising model in Section \ref{sec:isingresults} is one way to alleviate the shortcoming of the Gaussian cavity field assumption.

\section{Corrections to EP} \label{sec:corrections}

The $Z_{\EP}$ approximation can be corrected in a principled approach, which traces the following outline:
\begin{enumerate}
\item The exact  partition function $Z$ is re-written in terms of $Z_{\EP}$,  scaled by a correction factor $R=Z/Z_{\EP}$. This correction factor $R$ encapsulates the intractability in the model, and contains a ``local marginal'' contribution by each $f_a$ (see Section \ref{sec:exact-case}).

\item A ``handle'' on $R$ is obtained by writing it in terms of the
cumulants (to be defined in Section \ref{sec:cumulants}) of $q(\x)$ and $q_a(\x)$ from Equations (\ref{eq:qx}) and (\ref{eq:tilted}).
As $q_a(\x)$ and $q(\x)$ share their two first cumulants, the mean and covariance from the moment matching condition in Equation (\ref{eq:moments-match}),
a cumulant expansion of $R$ will be in terms of \emph{higher-order} cumulants (see Section \ref{sec:cumulants}).

\item $R$, defined in terms of cumulant differences, is written as a complex Gaussian average.
Each factor $f_a$ contributes a complex random variable $\k_a$ in this average
 (see Section \ref{sec:complex-expectation}).

\item Finally, the cumulant differences are used as ``small quantities'' in a Taylor series expansion of $R$, and the leading terms are kept (see Sections
\ref{sec:factorized} and \ref{sec:general-approximations}).

The series expansion is in terms of a  complex expectation with a \emph{zero} ``self-relation'' matrix,
and this has two important consequences.
Firstly, it causes all first order terms in the Taylor expansion to disappear, showing that $Z_{\EP}$ is correct to first order.
Secondly, due to Wick's theorem (introduced in Section \ref{sec:wick}),
these zeros will contract the expansion by making many other terms vanish.
\end{enumerate}
The strategy that is presented here can be re-used to correct other quantities of interest, like marginal distributions or the predictive density of new data when $p(\x)$ is a Bayesian probabilistic model. These corrections are outlined in Appendix \ref{sec:marginals}.

\subsection{Exact Expression for Correction} \label{sec:exact-case}

We define the (intractable) correction $R$ as $Z = R Z_{\EP}$. We can derive a useful expression for $R$ in a few steps as follows:
First we solve for $f_a$ in Equation (\ref{eq:tilted}), and substitute this into Equation (\ref{eq:px}) to obtain
\begin{equation} \label{eq:factor-product}
\prod_a f_a(\x)^{\facpower_a}
=
\prod_a \left(\frac{Z_a q_a(\x) g_a(\x)}{q(\x)}\right)^{\facpower_a}
= Z_{\EP} \; q(\x) \prod_a \left(\frac{q_a(\x) }{q(\x)}\right)^{\facpower_a}
 \ .
\end{equation}
We introduce $F(\x)$
$$
F(\x) \equiv \prod_a \left(\frac{q_a(\x) }{q(\x)}\right)^{\facpower_a}
$$
to derive the expression for the correction $R = Z/Z_{\EP}$ by integrating Equation (\ref{eq:factor-product}):
\begin{equation} \label{correc1}
R = \int  q(\x) F(\x) \, \mathrm{d} \x \ ,
\end{equation}
where we have used $Z = \int \prod_a f_a(\x)^{\facpower_a} \, \mathrm{d}\x $. Similarly we can write:
\begin{equation}\label{eq:exact}
p(\x) = \frac{1}{Z} \prod_a f_a(\x)^{\facpower_a} = \frac{Z_{\EP}}{Z} \; q(\x) F(\x) =  \frac{1}{R} \; q(\x) F(\x) \ .
\end{equation}
Corrections to the marginal and predictive densities of $p(\x)$ can be computed from this formulation.
This expression will become especially useful because the terms in $F(\x)$ turn out to be ``local'', that is, they only
depend on the marginals of the variables associated with factor $a$.
Let $f_a(\x)$ depend on the subset $\x_a$ of $\x$,
and let $\x_{\backslash a}$ (``$\x$ without $a$'') denote the remaining variables.
The distributions in Equations (\ref{eq:qx}) and (\ref{eq:tilted}) differ only with respect to their marginals on $\x_a$, $q_a(\x_a)$ and $q(\x_a)$, and therefore
\[
\frac{q_a(\x) }{q(\x)} =
\frac{  q(\x_{\backslash a}| \x_a) q_a(\x_a) }{ q(\x_{\backslash a}| \x_a) q( \x_a) } =
\frac{q_a(\x_a) }{q(\x_a)} \ .
\]
Now we can rewrite $F(\x)$ in terms of marginals:
\begin{equation} \label{eq:Fratio}
F(\x)  = \prod_a \left(\frac{q_a(\x_a) }{q(\x_a)}\right)^{\facpower_a} \ .
\end{equation}
The key quantity, then, is $F$, after which the key operation is to compute its expected value. The rest of this section is devoted to the task of obtaining a ``handle'' on $F$.

\subsection{Characteristic Functions and Cumulants} \label{sec:cumulants}

The distributions present in each of the ratios in $F(\x)$ in Equation (\ref{eq:Fratio}) share their first two cumulants, mean and covariance.  Cumulants and cumulant differences are formally defined in the next paragraph.
This simple observation has a crucial consequence:
As the $q(\x_a)$'s are Gaussian and do not contain any higher order cumulants (three and above), $F$ can be expressed in terms of the higher cumulants of the \emph{marginals} $q_a(\x_a)$.
When the term-product approximation is fully factorized,
these are simply cumulants of \emph{one}-dimensional distributions.

Let $N_a$ be the number of variables in subvector $\x_a$. In the examples presented in this work, $N_a$ is one or two.
Furthermore, let $\k_a$ be an $N_a$-dimensional vector
$\k_a = (k_1, \ldots, k_{N_a})_a$.
The characteristic function of $q_a$ is
\begin{equation} \label{eq:chiintermsofq}
\chi_a(\k_a) = \int \erm^{i \k_a^T \x_a} \, q_a(\x_a) \, \mathrm{d} \x_a = \bigE{ \erm^{i \k_a^T \x_a} }_{q_a} \ ,
\end{equation}
and is obtained through the Fourier transform of the density. Inversely,
\begin{equation} \label{eq:qintermsofchi}
q_a(\x_a) =  \frac{1}{(2\pi)^{N_a}} \int \erm^{-i \k_a^T \x_a} \chi_a(\k_a) \, \mathrm{d} \k_a \ .
\end{equation}
The cumulants  $c_{\balpha a}$ of $q_a$ are the coefficients that appear in the Taylor expansion of $\log \chi_a(\k_a)$ around the zero vector,
\[
c_{\balpha  a} = \left[ (-i)^l \left( \frac{\partial}{\partial \k_a} \right)^{\balpha} \log \chi_a(\k_a) \right]_{\k_a = \mathbf{0}}\ .
\]
By this definition of $c_{\balpha a}$, the Taylor expansion of $\log \chi_a(\k_a)$ is
\[
\log \chi_a(\k_a) = \sum_{l = 1}^{\infty} i^l \sum_{|\balpha| = l}
\frac{ c_{\balpha  a } }{\balpha !} \, \k_a^{\balpha} \ .
\]
Some notation was introduced in the above two equations to facilitate manipulating a multivariate series.
The vector $\balpha = (\alpha_1, \ldots, \alpha_{N_a})$, with $\alpha_j \in \mathbb{N}_0$, denotes a multi-index on the elements of $\k_a$.
Other notational conventions that employ $\balpha$ (writing $k_j$ instead of $k_{aj}$) are:
\[
|\balpha| = \sum_{j} \alpha_j \ , \qquad
\k_a^{\balpha} = \prod_{j} k_j^{\alpha_j} \ , \qquad
\balpha! = \prod_j \alpha_j! \ , \qquad
\left( \frac{\partial}{\partial \k_a} \right)^{\balpha} = \prod_j \frac{\partial^{\alpha_j}}{\partial k_j^{\alpha_j}} \ .
\]
For example, when $N_a = 2$, say for the edge-factors in a spanning tree, the set of multi-indices $\balpha$ where $|\balpha| = 3$ are $(3,0)$, $(2,1)$, $(1,2)$, and $(0,3)$.

There are two characteristic functions that come into play in $F(\x)$ and $R$ in Equation (\ref{eq:exact}). The first is that of the tilted distribution, $\log \chi_a(\k_a)$, and the other is the characteristic function of the EP marginal $q(\x_a)$, defined as $\chi(\k_a) = \E{ \erm^{i \k_a^T \x_a} }_{q}$. By virtue of matching the first two moments, and $q(\x_a)$ being Gaussian
with cumulants $c_{\balpha a}'$,
\begin{align} 
r_a(\k_a) = \log \chi_a(\k_a) - \log \chi(\k_a) & = \sum_{l \ge 1} i^l \sum_{|\balpha| = l}
\frac{ c_{\balpha a} -  c_{\balpha a}' }{\balpha !} \, \k_a^{\balpha} \nonumber \\
& = \sum_{l \ge 3} i^l \sum_{|\balpha| = l}
\frac{ c_{\balpha a} }{\balpha !} \, \k_a^{\balpha} \label{eq:ra}
\end{align}
contains the remaining higher-order cumulants where the tilted and approximate distributions \emph{differ}. All our subsequent derivations rest upon moment matching being attained. This especially means that one cannot use the derived corrections if EP has not converged.

\subsubsection{Ising Model Example} \label{sec:isingexample}

The cumulant expansion for the discrete distribution in Equation (\ref{eq:qtilt_n}) becomes
\begin{align*}
\log \chi_n (k_n) & = \log \int d x_n \, \erm^{i k_n x_n} q_n(x_n)
 = \log \left( \frac{1 + m}{2} \, \erm^{i k_n} + \frac{1 - m}{2} \, \erm^{-i k_n} \right) \\
 & = i m k_n - \frac{1}{2!}(1 - m^2) k_n^2
 - \frac{i}{3!} (- 2 m + 2m^3) k_n^3
 + \frac{1}{4!} (-2 + 8 m^2 - 6 m^4)k_n^4 + \cdots
\end{align*}
(we're compactly writing $m$ for $m_n$), from which the cumulants are obtained as
\begin{align*}
c_{1n} & = m  \ , & c_{4n} & = -2 + 8 m^2 - 6 m^4  \ , \\
c_{2n} & = 1 - m^2  \ ,  & c_{5n} & = 16m - 40m^3 + 24m^5  \ , \\
c_{3n} & = - 2 m + 2m^3  \ ,  & c_{6n} & = 16 - 136m^2 + 240m^4 - 120m^6 \ .
\end{align*}

\subsection{The Correction as a Complex Expectation} \label{sec:complex-expectation}

The expected value of $F$, which is required for the correction, has a dependence on a product of ratios of distributions
$q_a(\x_a) / q(\x_a)$. In the preceding section it was shown that the contributing distributions share lower-order statistics,
allowing a twofold simplification. Firstly, the ratio $q_a / q$ will be written as a single quantity that depends on
$r_a$, which was introduced above in Equation (\ref{eq:ra}).
Secondly, we will show that it is natural to shift integration variables into the complex plane,
and rely on complex Gaussian random variables
(meaning that both real and imaginary parts are jointly Gaussian).
These complex random variables that define the $r_a$'s have a peculiar property: they have a zero self-relation matrix! This property has important consequences in the resulting expansion.

\subsubsection{Complex Expectations} \label{sec:complexexpectations}

Assume that $q(\x_a) = \Ncal(\x_a \, ; \, \bmu_a, \bSigma_a)$ and $q_a(\x_a)$ share the same mean and covariance,
and substitute $\log \chi_a(\k_a) = r_a(\k_a) + \log \chi(\k_a)$ in the definition of $q_a$ in Equation (\ref{eq:qintermsofchi}) to give
\begin{equation} \label{eq:q_ratio_step1}
\frac{q_a(\x_a)}{q(\x_a)} = \frac{ \int \erm^{-i \k_a^T \x_a
+ r_a(\k_a) } \, \chi(\k_a) \, \mathrm{d} \k_a }{ \int \erm^{-i \k_a^T \x_a } \, \chi(\k_a) \, \mathrm{d} \k_a } \ .
\end{equation}
Although the $\k_a$ variables have not been introduced as random variables, we find it natural to \emph{interpret them
as such}, because the rules of expectations over Gaussian random variables will be extremely helpful in
developing  the subsequent expansions.
We will therefore write $q_a(\x_a)/q(\x_a)$ as an expectation of $\exp r_a(\k_a)$ over a density $p(\k_a|\x_a) \propto \erm^{-i \k_a^T \x_a } \chi(\k_a)$:
\begin{equation} \label{eq:qratio}
\frac{q_a(\x_a) }{q(\x_a)} = \BigE{ \exp r_a (\k_a) }_{\k_a | \x_a } \ .
\end{equation}
By substituting $\log \chi(\k_a)= i \bmu_a^T \k_a - \k_a^T \bSigma_a \k_a /2$ into Equation (\ref{eq:q_ratio_step1}),  we see
that $p(\k_a|\x_a)$ can be viewed as Gaussian, but not for real random variables!
We have to consider $\k_a$ as Gaussian random variables with a real and an imaginary part with
\[
\Re(\k_a) \sim {\cal N}\Big(\Re(\k_a) \, ; \, \0, \, \bSigma_a^{-1} \Big) \ , \qquad \Im(\k_a) = - \bSigma_a^{-1} (\x_a - \bmu_a) \ .
\]
For the purpose of computing the expectation in Equation (\ref{eq:qratio}), $\k_a|\x_a$ is a degenerate complex Gaussian that shifts the coefficients $\k_a$ into the complex plane.
The expectation of $\exp r_a(\k_a)$ is therefore taken over Gaussian random variables that have
$q(\x_a)$'s \emph{inverse} covariance matrix as their (real) covariance!
As shorthand, we write
\begin{equation} \label{eq:qafactorized}
p(\k_a|\x_a) = {\cal N} \Big(\k_a \, ; - i \bSigma_a^{-1} (\x_a - \bmu_a) \, , \,
\bSigma_a^{-1} \Big) \ .
\end{equation}
\begin{figure}[t]
\begin{center}
\epsfig{file=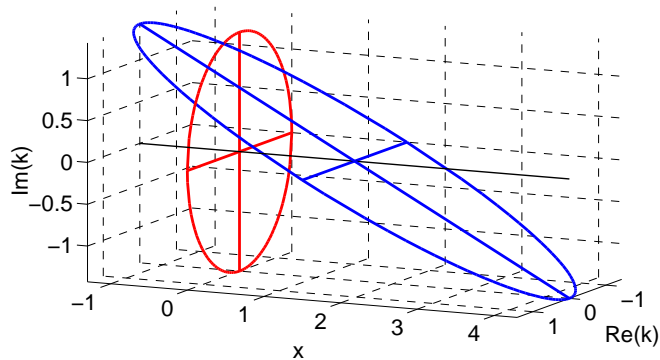, width=0.6\linewidth}
\caption{
Equation (\ref{eq:qafactorized}) shifts $\k_a$ to the complex plane. In the simplest case the joint density $p(k|x) \, q(x)$ is $x \sim {\cal N}(\mu, \sigma^2)$, $\Re(k) \sim {\cal N}(0, \sigma^{-2})$ and equality $\Im(k) = - \sigma^{-2}(x - \mu)$.
Notice that $\Re(k)$'s variance is the \emph{inverse} of that of $x$.
The joint density is a two-dimensional flat ellipsoidal pancake that lives in three dimensions: $x$ and the complex $k$ plane (tilted ellipsoid). Integrating over $x$ gives the
marginal over a complex $k$, which is \emph{still} a two-dimensional random variable (upright ellipsoid). The marginal has $\Im(k) \sim {\cal N}(0, \sigma^{-2})$, and hence $k$ has relation $\< (\Re(k) + i \Im(k))^2 \> = \sigma^{-2} - \sigma^{-2} = 0$ and variance $\< k \overline{k} \> = 2 \sigma^{-2}$.
}
\label{fig:kcomplex}
\end{center}
\end{figure}

\begin{figure}[t]
\begin{center}
\epsfig{file=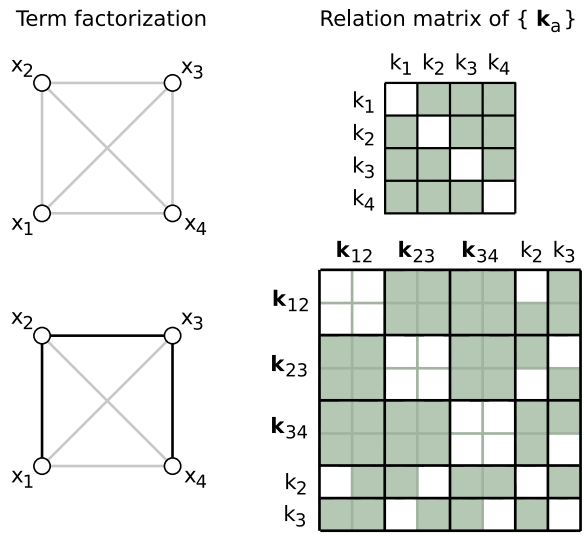, width=0.5\linewidth}
\caption{
The relation matrices between $\k_a$ for two factorizations of $\prod_{n=1}^{4} t_n(x_n)$: the top illustration is for $t_1 t_2 t_3 t_4$, while the bottom illustration is of a tree structure $(t_1 t_2)  (t_2 t_3)  (t_3 t_4) / t_2 / t_3$. The white squares indicate a zero relation matrix $\< \k_a \k_b^T\>$, with the \emph{diagonal} being zero.
From the properties of Equation (\ref{eq:kcov}) there are additional zeros
in the tree structure's relation matrix, where edge and node factors share variables.
The factor $f_0 = g_0$ is shadowed in grey in the left-hand figures, and can make $q(\x)$ densely connected.}
\label{fig:kcov}
\end{center}
\end{figure}

Figure \ref{fig:kcomplex} illustrates a simple density $p(\k_a|\x_a)$, showing that the imaginary component is a deterministic function of $\x_a$. Once $\x_a$ is averaged out of the joint density $p(\k_a|\x_a) \, q(\x_a)$, a \emph{circularly symmetric complex Gaussian} distribution over $\k_a$ remains. It is circularly symmetric as $\< \k_a \> = \0$, relation matrix $\< \k_a \k_a^T \> = \0$, and covariance matrix $\big\langle \k_a \overline{\k_a}^T \big\rangle = 2 \bSigma_a^{-1}$ (notation $\overline{k}$ indicates the complex conjugate of $k$). For the purpose of computing the expected values with Wick's theorem (following in Section \ref{sec:wick} below), we \emph{only} need the relations $\< \k_a \k_b^T \>$ for pairs of factors $a$ and $b$. All of these will be derived next:

According to Equation (\ref{correc1}), a further expectation over $q(\x)$ is needed, after integrating over $\k_a$, to determine $R$.
These variables will be \emph{combined} into complex random variables to make the averages in the expectation easier to derive.
By substituting Equation (\ref{eq:qratio}) into Equation (\ref{correc1}), $R$ is equal to
\begin{equation} \label{eq:Fexpectation}
R = \bigE{ F(\x) }_{ \x \sim q(\x) } = \< \prod_a \BigE{ \exp r_a(\k_a) }_{ \k_a | \x_a }^{\facpower_a} \>_{ \x } \ .
\end{equation}
When $\x$ is given, the $\k_a$-variables are independent. However, when they are averaged over $q(\x)$, the
$\k_a$-variables become coupled. They are zero-mean complex Gaussians
\[
\< \k_a \> = \< \< \k_a \>_{\k_a | \x_a } \>_{\x} = \BigE{ - i \bSigma_a^{-1} (\x_a - \bmu_a) }_{\x } = \mathbf{0}
\]
and are coupled with a zero self-relation matrix!
In other words, if $\bSigma_{ab} = \cov(\x_a, \x_b)$, the expected values
$\<\k_a \k_b^T \>$ between the variables in the set $\{ \k_a \}$ are
\begin{align}
\< \k_a \k_b^T \>
& = \< \< \k_a \k_b^T \>_{\k_{a,b} | \x} \>_{ \x }
+ i^2 \bSigma_a^{-1} \BigE{ (\x_a - \bmu_a) (\x_b - \bmu_b)^T }_{\x} \bSigma_b^{-1} \nonumber \\
& = \left\{ \begin{array}{ll}
\mathbf{0} & \textrm{if $a = b$} \\
-\bSigma_a^{-1} \bSigma_{ab} \bSigma_b^{-1}& \textrm{if $a \neq b$}
\end{array} \right. \ . \label{eq:kcov}
\end{align}
Complex Gaussian random variables are additionally characterized by $\big\langle \k_a \overline{\k_b}^T \big\rangle$. However, these expectations are not required for computing and simplifying the expansion of $\log R$ in Section \ref{sec:factorized}, and are not needed for the remainder of this paper.
Figure \ref{fig:kcov} illustrates the structure of the resulting relation matrix $\< \k_a \k_b^T \>$ for two different factorizations of the same distribution. Each factor $f_a$ contributes a $\k_a$ variable, such that the tree-structured approximation's relation matrix will be larger than that of the fully factorized one.

Section \ref{sec:factorized} shows that when $\facpower_a = 1$, the above expectation can be written directly over $\{ \k_a \}$ and expanded.
In the general case, discussed in Section \ref{sec:general-approximations}, the inner expectation is first expanded (to treat the $\facpower_a$ powers) before computing an expectation over $\{ \k_a \}$.
In both cases the expectation will involve polynomials in $k$-variables.
The expected values of Gaussian polynomials can be evaluated with
Wick's theorem.

\subsection{Wick's Theorem} \label{sec:wick}

Wick's theorem provides a useful formula for mixed central moments of Gaussian variables. Let $k_{n_1}, \ldots, k_{n_\ell}$  be real or complex
centered jointly Gaussian variables, noting that they do not have to be different. Then
\begin{equation} \label{eq:wick}
\< k_{n_1} \cdots k_{n_\ell} \> = \sum \prod_{\eta} \< k_{i_\eta} k_{j_\eta} \> \ ,
\end{equation}
where the sum is over all partitions of $\{ n_1, \ldots, n_\ell \}$ into disjoint pairs $\{ i_\eta , j_\eta\}$. If $\ell = 2 m$ is even, then there are $(2m)! / ( 2^m m!) = (2m-1)!!$ such partitions.\footnote{The double factorial is $(2m-1)!! = (2m-1) \times (2m-3) \times (2m-5) \times \cdots 1$.} If $\ell$ is odd, then there are none, and the expectation in Equation (\ref{eq:wick}) is zero.

Consider the one-dimensional variable $k \sim {\cal N}(k ; 0, \sigma^2)$. Wick's theorem states that $\E{ k^\ell } = (\ell - 1)!! \, \sigma^\ell$ if $\ell$ is even, and $\E{ k^\ell } = 0$ if $\ell$ is odd. In other words,
$\E{ k^3 } = 0$, $\E{ k^4 } = 3 (\sigma^2)^2$,  $\E{ k^6 } = 15 (\sigma^2)^3$, and so forth.

\section{Factorized Approximations} \label{sec:factorized}

In the fully factorized approximation, with $f_n(x_n) = t_n(x_n)$, the exact distribution  in Equation (\ref{eq:exact}) depends on the \emph{single node marginals} $F(\x) = \prod_n q_n(x_n) / q(x_n)$.
Following Equation (\ref{eq:Fexpectation}),
the correction to the free energy
\begin{equation} \label{eq:factorizedR}
R = \< \prod_n \BigE{ \exp r_n (k_n) }_{k_n | x_n} \>_\x =
\< \exp \left[ \sum_n r_n (k_n) \right] \>_{\k}
\end{equation}
is taken directly over the centered complex-valued Gaussian random variables $\k = (k_1, \ldots, k_N)$, which have a
relations
\begin{equation} \label{eq:kcov1d}
\< k_m k_n \>
= \left\{ \begin{array}{ll}
0 & \textrm{if $m = n$} \\
- \Sigma_{mn} / (\Sigma_{mm} \Sigma_{nn}) & \textrm{if $m \neq n$}
\end{array} \right. \ .
\end{equation}
In the section to follow, all expectations shall be with respect to $\k$, which will be dropped where it is clear from the context.

Thus far, $R$ is re-expressed in terms of site contributions. The  expression in Equation (\ref{eq:factorizedR}) is exact, albeit still intractable, and will be treated through a power series expansion. Other quantities of interest, like marginal distributions or moments, can similarly be expressed exactly, and then expanded (see Appendix \ref{sec:marginals}).

\subsection{Second Order Correction to $\log R$}

Assuming that the $r_n$'s are small on average with respect to $\k$, Equation (\ref{eq:factorizedR}) is expanded and the lower order terms kept:
\begin{align}
\log R = \log \< \exp\left[ \sum_n r_n (k_n) \right]
\>
& = \sum_n \< r_n \> + \frac{1}{2}
\< \left(\sum_n r_n\right)^2 \> - \frac{1}{2} \left(\sum_n \< r_n \> \right)^2 + \cdots \nonumber \\
& = \frac{1}{2}
\sum_{m\neq n} \< r_m r_n \> + \cdots \label{eq:factorized2ndOrder}
\end{align}
The simplification in the second line is a result of the variance terms being zero from Equation (\ref{eq:kcov1d}).
The single marginal terms also vanish (and hence EP is correct to first order) because both
$\< k_n \> = 0$ and $\< k_n^2 \> = 0$.

This result can give us a hint in which situations the corrections are expected to be small:
\begin{itemize}
\item Firstly, the $r_n$ could be small for values of $k_n$ where the density of $\k$
is not small. For example, under a zero noise Gaussian process classification model, $q_n(x_n)$ equals a step function $t_n(x_n)$
times a Gaussian, where the latter often has small variance compared to
the mean. Hence, $q_n(x_n)$ should be very close to a Gaussian.
\item Secondly, for systems with weakly (posterior) dependent variables $x_n$ we might expect that the
log partition function $\log Z$ would scale approximately linearly with $N$, the number of variables.
Since terms with $m=n$ vanish in the computation of $\ln R$, there are no corrections that are \emph{proportional to $N$} when $\Sigma_{mn}$ is sufficiently small as $N\to\infty$. Hence, the dominant contributions to $\log Z$
should already be included in the EP approximation.
However, Section \ref{sec:gpbox} illustrates an example where this need not be the case.
\end{itemize}
The expectation $\< r_m r_n \>$, as it appears in Equation (\ref{eq:factorized2ndOrder}), is treated by
substituting $r_n$ with its cumulant expansion $r_n(k_n) = \sum_{l\geq 3}  i^l  c_{l n} k_n^l / l!$ from Equation (\ref{eq:ra}).
Wick's theorem now plays a pivotal role in evaluating the expectations that appear in the expansion:
\begin{align}
\< r_m(k_m) r_n(k_n) \> & =
\sum_{l,s \geq 3}  i^{l + s} \frac{c_{l n} \, c_{s m} }{l! s!}
 \E{k_m^s k_n^l } \nonumber \\
& = \sum_{l \geq 3}  i^{2 l} l! \frac{c_{l n} \, c_{s m} }{ (l!)^2}
 \E{k_m k_n}^l \nonumber \\
& = \sum_{l \geq 3}   \frac{c_{l m} \, c_{l n} }{l!}
\left(\frac{\Sigma_{mn}}{\Sigma_{mm}\Sigma_{nn}} \right)^l \ . \label{eq:r-cumulants}
\end{align}
The second line above follows from contractions in Wick's theorem.
All the \emph{self-pairing terms}, when for example one of the $l$ $k_n$'s is paired with another $k_n$ in Equation (\ref{eq:wick}), are zero because $\< k_n^2 \> = 0$.
To therefore get a non-zero result for $\< k_m^s k_n^l \>$, using Equation (\ref{eq:wick}),
\emph{each} factor $k_n$ has to be paired with some factor $k_m$, and this is possible
only when $l = s$. Wick's theorem sums over all pairings, and there are $l!$ ways of pairing a $k_n$ with a $k_m$, giving the result in Equation (\ref{eq:r-cumulants}).
Finally, plugging Equation (\ref{eq:r-cumulants}) into Equation (\ref{eq:factorized2ndOrder}) gives the second order correction
\begin{equation} \label{eq:logR-factorized-corr}
\log R = \frac{1}{2} \sum_{m\neq n}  \sum_{l \geq 3}   \frac{c_{l m} \, c_{l n} }{l!}
\left(\frac{\Sigma_{mn}}{\Sigma_{mm}\Sigma_{nn}} \right)^l + \cdots \ .
\end{equation}

\subsubsection{Ising Example Continued}

We can now compute the second order $\log R$ correction for the $N=2$ Ising model example of Section \ref{sec:EPIsing}. The covariance matrix has $\Sigma_{nn} = 1$ from moment matching and $\Sigma_{12}= J / (\lambda^2 - J^2)$ with
$\lambda = \frac{1}{2} \left[ J^2 + \sqrt{J^4 + 4} \right]$. The uneven terms in the cumulant expansion derived in Section \ref{sec:isingexample} disappear because $m=0$. The first nontrivial term is therefore $l=4$ which gives a contribution of $\frac{1}{2} \times 2 \times  \frac{c_{4}^2}{4!} \Sigma_{12}^4= \frac{(-2)^2}{4!} \Sigma_{12}^4= \frac{1}{6}\Sigma_{12}^4$. In Section \ref{sec:EPIsing}, we saw that $\log Z - \log Z_{\rm EP} = \frac{J^4}{6}$ plus terms of order $J^6$ and higher. To lowest order in $J$ we have $\Sigma_{12}=J$ and thus $\log R = \frac{J^4}{6}$ which exactly cancels the lowest order error of EP.

\subsection{Corrections to Other Quantities}

The schema given here is applicable to any other quantity of interest, be it marginal or predictive distributions, or the marginal moments of $p(\x)$.
The cumulant corrections for the marginal moments are derived in Appendix \ref{sec:marginals};
for example, the correction to the marginal mean $\mu_i$ of an approximation $q(\x) = {\cal N}(\x ; \bmu, \bSigma)$ is
\begin{equation} \label{eq:marginal-mean}
\< x_i \>_{p(\x)} - \mu_i =
\sum_{l\geq 3} \sum_{j\neq n}
\frac{\Sigma_{ij}}{\Sigma_{jj}} \frac{c_{l+1,j} c_{ln}} {l!}
\left(\frac{\Sigma_{j n}}{\Sigma_{jj}\Sigma_{nn}} \right)^l + \cdots \ ,
\end{equation}
while the correction to the marginal covariance is
\begin{align}
\< (x_i - \mu_i)(x_{i'} - \mu_{i'}) \>_{p(\x)} - \Sigma_{i i'} & = 
\sum_{l\geq 3} \sum_{j\neq n} \frac{\Sigma_{ij}\Sigma_{i'j}}{\Sigma_{jj}^2} \frac{c_{l+2,j} c_{ln}} {l!}
\left(\frac{\Sigma_{jn}}{\Sigma_{jj}\Sigma_{nn}} \right)^l \nonumber \\
& \quad\quad +
\sum_{l\geq 3} \sum_{j\neq n}
\frac{\Sigma_{ij}}{\Sigma_{jj}}\frac{\Sigma_{i'n}}{\Sigma_{nn}} \frac{c_{lj} c_{ln}} {l!}
\left(\frac{\Sigma_{jn}}{\Sigma_{jj}\Sigma_{nn}} \right)^{l-1} + \cdots \ .
\label{eq:marginal-covariance}
\end{align}

\subsection{Edgeworth-Type Expansions}

To simplify the expansion of Equation (\ref{eq:factorizedR}),
we integrated (combined) degenerate complex Gaussians $k_n | x_n$ over $q(\x)$ to obtain fully complex Gaussian random variables $\{ k_n \}$. We've then relied on $\< k_n^2\> = 0$ to simplify the expansion of $\log R$.

The expectations $\< k_n^2\> = 0$ are closely related to the orthogonality of Hermite polynomials,
and this can be employed in an alternative derivation.
In particular, one can \emph{first} make a Taylor expansion of $\exp r_n(k_n)$ around zero, giving complex-valued polynomials in $\{ k_n \}$.
When the inner average in Equation (\ref{eq:factorizedR}) is then taken over $k_n | x_n$, a real-valued series of Hermite polynomials in $\{ x_n \}$ arises. These polynomials are orthogonal under $q(\x)$. The series that describes the tilted distribution $q_n(x_n)$ is equal to
the product of $q(x_n)$ and an expansion of polynomials for the higher-cumulant \emph{deviation} from a Gaussian density. This line of derivation gives an Edgeworth expansion for\emph{each} factor's tilted distribution.

As a second step, Equation (\ref{eq:factorizedR}) couples the product of separate Edgeworth expansions (one for each factor) together by requiring an outer average over $q(\x)$. The orthogonality of Hermite polynomials under $q(\x)$ now come into play: it allows products of orthogonal polynomials under $q(\x)$ to integrate to zero. This is similar to contractions in Wick's theorem, where $\< k_n^2\> = 0$ allows us to simplify Equation (\ref{eq:r-cumulants}).
Although it is not the focus of this work, an example of such a derivation appears in Appendix \ref{sec:one-dim-analytic}. 

\section{Radius of Convergence} \label{sec:convergence}

We may hope that in practice the low order terms in the cumulant expansions
will account already for the dominant contributions. But will such an expansion actually converge when
extended to arbitrary orders? While we will leave a more general answer to future research, we can at least give 
a partial result for the example of the Ising model.
Let $\D = \mathrm{diag}(\bSigma)$, the diagonal of the covariance matrix of the EP approximation $q(\x)$.
We prove here that a cumulant expansion for $R$ will converge when the eigenvalues of $\D^{-1/2} \bSigma \D^{-1/2}$---which has diagonal values of one---are bounded between zero and two.

In practice we've found that even if the largest of these eigenvalues grows with $N$, the second-order correction gives a remarkable improvement.
This, with the results in Figure \ref{fig:gpbox}, lead us to believe that the power series expansion is often divergent.
It may well be that our expansions are only of an asymptotic type \citep{boyd99asymptotic} for which the summation
of only a certain number of terms might give an improvement whereas further terms would lead to
worse results.
It leads to a paradoxical situation, which seems common when interesting functions are computed:
On the one hand we may have a series which does not converge, but in many ways is more practical;
on the other hand one might obtain an expansion that converges, but only impractically. 
Quoting George F.~Carrier's rule from
\cite{boyd99asymptotic}:
\begin{quote}
\emph{Divergent series converge faster than convergent series because they don't have to converge.}
\end{quote}
For this, we do not yet have a clear-cut answer.

\subsection{A Formal Expression for the Cumulant Expansion to All Orders}

To discuss the question when our expansion will converge when
extended to arbitrary orders, we introduce a single extra parameter $\lambda$ into $R$,
which controls the strength of the contribution of cumulants. Expanded into a series in powers of
$\lambda$, contributions of cumulants
of {\em total} order $l$ are multiplied by a factor $\lambda^l$, for example
$\lambda^l c_{nl}$ or $\lambda^{k+l} c_{n k} c_{nl}$. Of course, at the end of the calculation, we set $\lambda =1$.
This approach is obviously achieved by replacing
$$
r_n(k_n) \rightarrow r_n(\lambda k_n)
$$
in Equation (\ref{eq:factorizedR}). Hence, we define
\[
R(\lambda) = \< \exp \left[ \sum_n r_n (\lambda k_n) \right] \>_{\k} =
\< \exp \left[ \sum_n r_n ( k_n) \right] \>_{\k'}
\]
where
$$
\< k'_m k'_n \>
= \left\{ \begin{array}{ll}
0 & \textrm{if $m = n$} \\
- \lambda^2 \Sigma_{mn} / (\Sigma_{mm} \Sigma_{nn}) & \textrm{if $m \neq n$}
\end{array} \right. \ .
$$
By working backwards, and expressing everything by the original densities over $x_n$, the correction can be written as
\begin{equation} \label{eq:formalexpressionR}
R(\lambda) = \< \prod_n \frac{q_n(x_n)}{q(x_n)} \>_{q_\lambda(\x)} \ ,
\end{equation}
where the density $q_\lambda(\x)$ is a multivariate Gaussian with mean 
$\bmu$ and covariance given by
$$
\bSigma_\lambda = \D + z (\bSigma -  \D)\ ,
$$
where $\D = \mbox{diag}(\bSigma)$ and $z = \lambda^2$.
Hence, we see that the expansion in powers of $\lambda$ is actually equivalent to an
expansion in products of nondiagonal elements of $\bSigma$.

Noticing that as $R(\lambda)$ depends on $\lambda$ through the density
$
q_\lambda(\x) \propto |\bSigma_\lambda|^{-1/2} \erm^{-\oneh \x^\top \bSigma^{-1}_\lambda \x}$, we can see 
by expressing $\bSigma^{-1}_\lambda$ in terms of eigenvalues and eigenvectors
that
for any {\em fixed} $\x$,  $q_\lambda(\x)$ is an analytic function of the {\em complex variable}  $z$ as long as
$\bSigma_\lambda$ is positive definite. Since
$$
\bSigma_\lambda = \D^{1/2} \left \{\I + z\left(\D^{-1/2}\bSigma \D^{-1/2}- \I\right)\right\}\D^{1/2}
$$
this is equivalent to the condition that
the matrix
$\I + z (\D^{-1/2}\bSigma \D^{-1/2}- \I)$ is positive definite.
Introducing $\gamma_i$, the eigenvalues  of  $\D^{-1/2}\bSigma \D^{-1/2}$, positive definiteness 
fails when for the first time $1 + z (\gamma_i -1) =0$.
Thus the series for $q_\lambda(\x)$ is convergent for
$$
|z| < \min_i \frac{1}{|1 - \gamma_i|} \ .
$$
Setting $z =1$, this is equivalent to the condition
$$
1 < \min_i \frac{1}{|1 - \gamma_i|} \ .
$$
This means that the eigenvalues have to fulfil $0< \gamma_i < 2$.
Unfortunately, we can not conclude from this condition that pointwise convergence of $q_\lambda(\x)$
for each $\x$ leads to convergence of $R(\lambda)$ (which is an integral of $q_\lambda(\x)$ over all $\x$!).
However, in cases where the integral eventually becomes a finite sum, such as the Ising model, pointwise
convergence in $\x$ leads to convergence of $R(\lambda)$.

\subsubsection{Ising Model Example}

From Section \ref{sec:isingexample} the tilted distribution for the running example Ising model is $q_n(x_n) = \frac{1}{2} [ \delta(x_n + 1) + \delta(x_n - 1)]$,
and hence $q(x_n) = \frac{1}{(2\pi)^{1/2}} \erm^{- x_n^2 / 2}$. As each $q(x_n)$ is a unit-variance Gaussian, $\D=\mbox{diag}(\bSigma) = \I$. Hence $\D^{-1/2}\bSigma \D^{-1/2}=\bSigma$ and
$$
R(\lambda) =
\frac{1}{\sqrt{|(1- \lambda^2) \I + \lambda^2 \bSigma |}}
\frac{\erm^{N/2}}{2^N}
\sum_{\x \in \{-1,1\}^N} \exp\left[ - \oneh \x^T \left((1- \lambda^2) \I + \lambda^2 \bSigma\right)^{-1} \x\right]
$$
follows from Equation (\ref{eq:formalexpressionR}).
The arguments of the previous section show that the {\em radius of convergence} of $R(\lambda)$ is determined by the condition that the matrix  $\I + \lambda^2
(\bSigma - \I)$ is positive definite or the eigenvalues $l_i$ of $\bSigma$ fulfil $|l_i -1| \leq 1/\lambda^2$.

In the $N=2$ case,  $\bSigma = \left(\begin{array}{cc}
1  & c\\
c & 1 \\
\end{array}\right)$ with $c=c(J) \in ]-1,1[$ which has eigenvalues $1-c$ and $1+c$,
meaning that cumulant expansion for
$R(\lambda)$ is convergent for the $N=2$ Ising model. For $N>2$, it is easy to show that this is not necessarily true. Consider the `isotropic' Ising model with $J_{ij}=J$ and zero external field, then $\Sigma_{ii}=1$ and $\Sigma_{ij}=c$ for $i \neq j$ with $c=c(J) \in ]-1/(N-1),1[$. The eigenvalues are now $1+(N-1)c$ and $1-c$ (the latter with degeneracy $N-1$). For finite $c$, the largest eigenvalue will scale with $N$ and thus be larger than the upper value of two that would be required for convergence.
Scaling with $N$ for the largest eigenvalue of $\D^{-1/2}\bSigma \D^{-1/2}$ is also observed in the Ising model simulations Section \ref{sec:isingresults}.

We conjecture that convergence of the cumulant series for $R(\lambda)$ also implies convergence of 
the series for $\log R(\lambda)$ but leave an investigation of this point to future research.  
We only illustrate this point for the $N=2$ Ising model case, where we have the explicit
formula
$$
\log R(\lambda) =  1 -\oneh \log \left(1 - \lambda^4 c^2\right)
- \frac{1}{1 - \lambda^4 c^2} + \log \cosh\left(
\frac{\lambda^2 c}{1 - \lambda^4 c^2}
\right) \ .
$$
As can be easily seen, an expansion in $\lambda$ converges for $c^2 \lambda^4 < 1$ which gives the
same radius of convergence $|c| < 1$ as for the expansion of $R$.

\section{General Approximations} \label{sec:general-approximations}

The general approximations differ from the factorized approximation in that an expansion in terms of expectations under $\{ \k_a \}$ doesn't immediately arise.
Consider $R$ in Equation (\ref{eq:Fexpectation}): Its inner expectations are over $\k_a | \x$, and outer expectations are over $\x$.
First take the binomial expansion of the \emph{inner} expectation, and keep it
to second order in $r_a$:
\begin{align*}
\< \erm^{r_a(\k_a)} \>_{\k_a | \x}^{\facpower_a}
& =
\left( 1 + \<r_a\> + \frac{1}{2} \<r_a^2\> + \cdots \right)^{\facpower_a} \\
& = 1 + \facpower_a \left[ \<r_a\> + \frac{1}{2} \<r_a^2\> + \cdots \right]
+ \frac{ \facpower_a (\facpower_a - 1)}{2} \left[ \<r_a\> + \frac{1}{2} \<r_a^2\> + \cdots \right] ^2 + \cdots \\
& = 1 +  \facpower_a \<r_a\> + \frac{\facpower_a}{2} \<r_a^2\>
+  \frac{ \facpower_a (\facpower_a - 1)}{2} \<r_a\>^2 + \cdots \ .
\end{align*}
Notice that $r_a(\k_a)$ can be complex, but $\< r_a(\k_a) \>_{\k_a | \x}$, as it appears in the above expansion, is real-valued.
Using this result, again expand $\E{ \prod_a \E{ \erm^{r_a} }_{\k_a | \x}^{\facpower_a} }_{\x}$. The correction to $\log R$, up to second order, is
\begin{align}
\log R & = \frac{1}{2} \sum_{a \neq b} \facpower_a \facpower_b \< \<r_a(\k_a) \>_{\k_a | \x}
\<r_b(\k_b) \>_{\k_b | \x} \>_{\x} \nonumber \\
& \qquad \qquad + \frac{1}{2} \sum_{a} \facpower_a (\facpower_a - 1) \< \<r_a(\k_a) \>_{\k_a | \x}^2 \>_{\x}
+ \cdots \ . \label{eq:generalLogR}
\end{align}
In the above relation the first-order terms all disappeared as $\E{ \E{r_{a} (\k_a)} } = 0$.
Terms involving $\E{ \E{r_{a}(\k_a)^2} } = 0$ similarly disappear, as every polynomial in the expansion $r_{a}(\k_a)^2$ averages to zero. This is a general case of Equation (\ref{eq:factorized2ndOrder}), in which $\facpower_n = 1$ for all factors. In Appendix \ref{app:tree} we show how to use the general result for the case where the factorization is a tree and our factors are edges (pairs) and nodes (single variables).

\section{Gaussian Process Results} \label{sec:gpresults}

One of the most important applications of EP is to statistical models with Gaussian process (GP) priors, where $\x$ is a latent variable with Gaussian prior distribution with a kernel matrix $\K$ as covariance $\mathbb{E}[\x \x^T] = \K$.

It is well known that for many models, like GP classification, inference with EP is on par with MCMC ground truth \citep{kuss06assessing}. Section \ref{sec:gpc-results} underlines this case, and shows
corrections to the partition function on the USPS data set over a range of kernel hyperparameter settings.

A common inference task is to predict the output for previously unseen data. Under a GP regression model, a key quantity is the predictive mean function. The predictive mean is analytically tractable when the latent function is corrupted with Gaussian noise to produce observations $y_n$. This need not be the case; in Section \ref{sec:uniform-noise} we examine the problem of quantized regression, where the noise model is non-Gaussian with sharp discontinuities.
We show practically how the corrections transfer to other moments, like the predictive mean.
Through it, we arrive at a hypothetical rule of thumb: if the data isn't ``sensible'' under the (probabilistic) model of interest, there is no guarantee for EP giving satisfactory inference.

Armed with the rule of thumb, Section \ref{sec:gpbox} constructs an insightful counterexample where the EP estimate diverges or is far from ground truth with more data. Divergence in the partition function
is manifested in the initial correction terms, giving a test for the approximation accuracy that doesn't rely on any Monte Carlo ground truth.

\subsection{Gaussian Process Classification} \label{sec:gpc-results}

The GP classification model arises when we observe $N$ data points $\s_{n}$ with class labels $y_{n} \in \{-1, 1 \}$, and model $y$ through a latent function $x$ with a GP prior. The likelihood terms for $y_{n}$ are assumed to be $t_{n}(x_{n}) = \Phi(y_{n} x_{n})$, where $\Phi(\cdot)$ denotes the cumulative Normal density.

\begin{figure}[t]
\centering
\epsfig{file=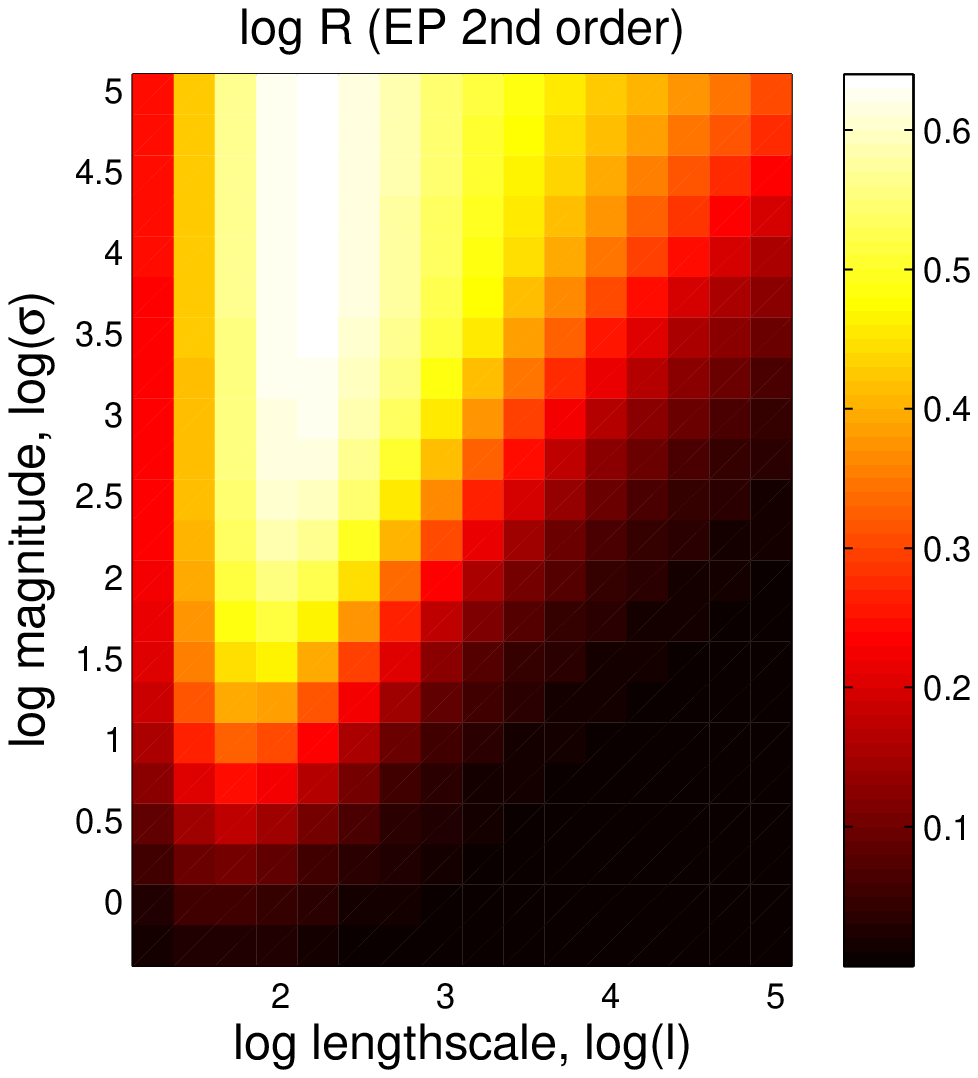, width=0.45\linewidth}
$\phantom{\quad}$
\epsfig{file=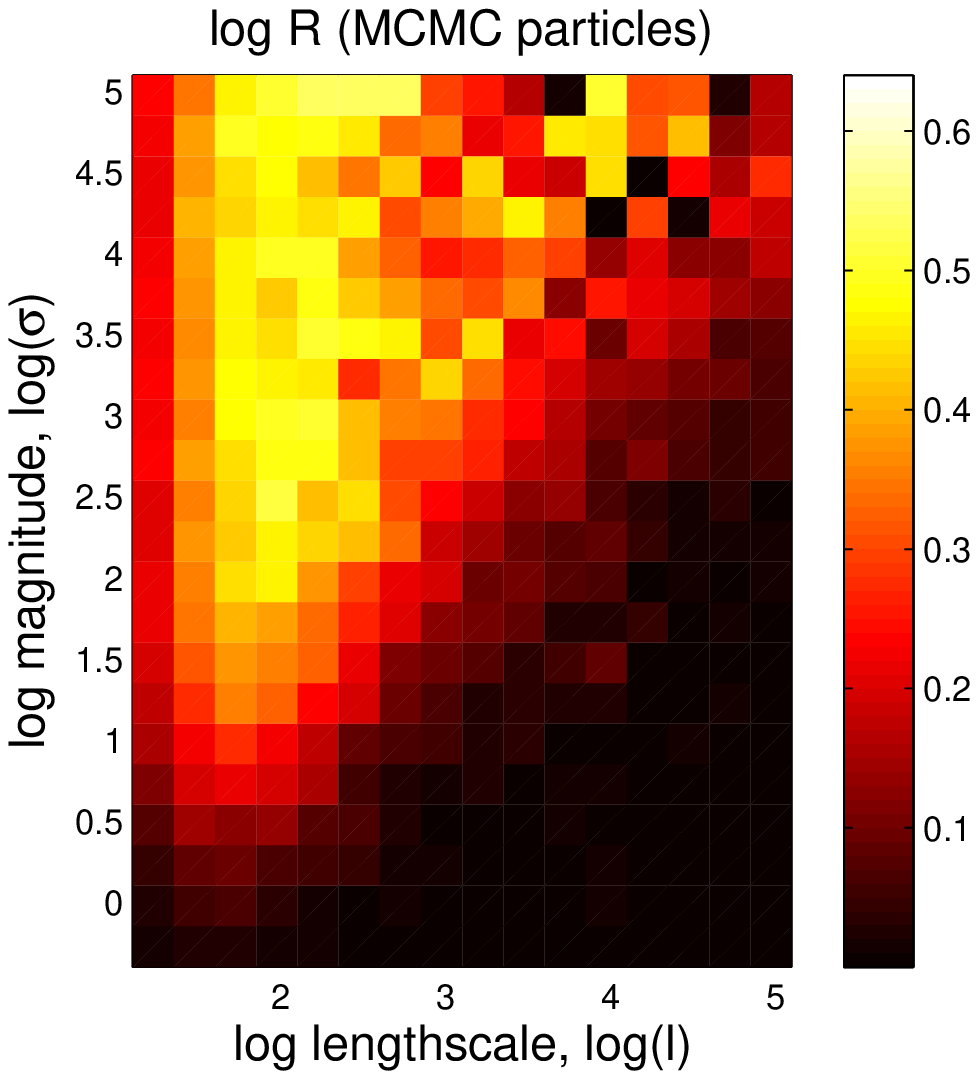, width=0.45\linewidth}
\caption{
A comparison of $\log R$ using a perturbation expansion of Equation (\ref{eq:logR-factorized-corr}) against Monte Carlo estimates of $\log R$, using the USPS data set from \citet{kuss06assessing}.
The second order correction to $\log R$, with $l = 3, 4$, is used on the \emph{left}; the \emph{right} plot uses a Monte Carlo estimate of $\log R$.
} \label{fig:USPS}
\end{figure}

An extensive MCMC evaluation of EP for GP classification
on various data sets was given by \cite{kuss06assessing}, showing that the log marginal likelihood of the data can be approximated remarkably well.
As shown by \cite{opper08improving}, an even more accurate estimation of the approximation error is given by considering the second order correction in Equation (\ref{eq:logR-factorized-corr}).
For GPC we generally found that the $l = 3$ term dominates $l = 4$, and we do not include any higher cumulants here.

Figure \ref{fig:USPS} illustrates
the correction to $\log R$, with $l = 3, 4$,
on the binary subproblem of the USPS 3's vs.~5's digits data set, with $N = 767$.
This is the same set-up of \cite{kuss06assessing} and \cite{opper08improving}, using the kernel
$k(\s, \s') = \sigma^{2} \exp ( -\frac{1}{2} \| \s - \s'\|^{2} / \ell^{2})$, and we refer the reader to both papers for additional and complimentary figures and results. We evaluated Equation (\ref{eq:logR-factorized-corr}) on a similar grid of $\log \ell$ and $\log \sigma$ values.
For the same grid values we obtained Monte Carlo estimates of $\log Z$, and hence $\log R$.
The correction, compared to the magnitude of the $\log Z$ grids by \cite{kuss06assessing},
is remarkably small, and underlines
their findings on the accuracy of EP for GPC.

The correction from Equation (\ref{eq:logR-factorized-corr}), as computed here, is $\mathcal{O}(N^{2})$, and compares favorably to $\mathcal{O}(N^{3})$ complexity of EP for GPC.

\subsection{Uniform Noise Regression} \label{sec:uniform-noise}

We turn our attention to a regression problem, that of learning a latent function $x(\s)$ from inputs $\{ \s_n \}$ and matching real-valued observations $\{ y_n \}$.
A frequent nonparametric treatment assumes that $x(\s)$ is \emph{a priori} drawn from a GP prior with covariance function $k(\s,\s')$, from which a corrupted version $y$ is observed.
Analytically tractable inference is no longer possible in this model when the observation noise is non-Gaussian.
Some scenarios include that of quantized regression, where $y_n$ is formed by rounding $x(\s_n)$ to, say, the nearest integer, or where $x(\s)$ indicates a robot's path in a control problem, with conditions to stay within certain ``wall'' bounds. In these scenarios the latent function $x(\s_n)$ can be reconstructed from $y_n$ by adding sharply discontinuous uniformly random ${\cal U}[-a,a]$ noise,
\[
p(\x) = \frac{1}{Z} \prod_{n} \mathbb{I} \Big[ |x_n - y_n| < a \Big] \, {\cal N}( \x \, ; \, \mathbf{0}, \, \K) \ .
\]

We now assume an EP approximation $q(\x) = {\cal N}(\x \, ; \bmu, \, \bSigma)$, which can be obtained by using
the moment calculations in Appendix \ref{sec:uniform-cumulants}.
To simplify the exposition of the predictive marginal, we follow the notation of \citet[Chapter 3]{rasmussen05gaussian} and let $\blambda_n = (\tau_n, \nu_n)$, so that the final EP approximation multiplies $g_n$ terms $\prod_{n} \exp \{ - \frac{1}{2} \tau_{n} x_{n}^2 + \nu_{n} x_{n} \}$ into a joint Gaussian ${\cal N}(\x \, ; \, \mathbf{0}, \K)$.

\subsubsection{Making Predictions for New Data}

The latent function $x(\s_*)$ at any new input $\s_*$ is obtained by the predictive marginal $q(x_{*})$ of $q(\x, x_{*})$. The marginal $q(x_{*})$---given below in Equation (\ref{eq:regressionmarginal})---is directly obtained from the EP approximation $q(\x) = {\cal N}(\x \, ; \bmu, \, \bSigma)$. However, the correction to its mean, as was given in Equation (\ref{eq:marginal-mean}), requires covariances $\Sigma_{*n}$, which are derived here.

Let $\kappa_{*} = k(\s_{*}, \s_{*})$, and $\k_{*}$ be a vector containing the covariance function evaluations $k(\s_{*}, \s_{n})$.
Again following \cite{rasmussen05gaussian}'s notation, let $\tilde{\bSigma}$ be the diagonal matrix containing $1 / \tau_{n}$ along its diagonal. The EP covariance, on the inclusion of $x_{*}$, is
\begin{align}
\bSigma_{*} & = \left( \left[ \begin{array}{cc} \K & \k_{*} \\ \k_{*}^T & \kappa_{*} \end{array} \right]^{-1}  +
\left[ \begin{array}{cc} \tilde{\bSigma}^{-1} & \mathbf{0} \\ \mathbf{0}^T & 0 \end{array} \right]
\right)^{-1} \nonumber \\
& =
\left[ \begin{array}{cc} \bSigma
& \k_{*} - \K  (\K + \tilde{\bSigma})^{-1} \k_{*} \\
\k_{*}^T - \k_{*}^T (\K + \tilde{\bSigma})^{-1} \K
& \kappa_{*} - \k_{*}^T (\K + \tilde{\bSigma})^{-1} \k_{*} \end{array} \right]
\ , \label{eq:sigmastar}
\end{align}
with $\bSigma = \K - \K (\K + \tilde{\bSigma})^{-1} \K$.
There is no observation associated with $\s_{*}$, hence $\tau_{*} = 0$ in the first line above, and its inclusion has $c_{l*} = 0$ for $l \ge 3$.
The second line follows by computing matrix partitioned inverses twice on $\bSigma_{*}$. The joint EP approximation for any new input point $\s_{*}$ is directly obtained as
\[
q(\x, x_{*}) = {\cal N} \left(
\left[ \begin{array}{c} \x \\ x_{*} \end{array} \right]
\, ; \, \left[ \begin{array}{c} \bmu \\ \k_{*}^T \K^{-1} \bmu \end{array} \right] , \, \bSigma_{*}
\right) \ ,
\]
with the marginal $q(x_{*})$ being
\begin{equation} \label{eq:regressionmarginal}
q(x_{*}) = {\cal N}( x_{*} \, ; \, \k_{*}^T \K^{-1} \bmu, \, \kappa_{*} - \k_{*}^T (\K + \tilde{\bSigma})^{-1} \k_{*}) =
{\cal N}( x_{*} \, ; \, \mu_*, \, \sigma_*^2) \ .
\end{equation}
According to Equation (\ref{eq:marginal-mean}), one needs the covariances $\Sigma_{*j}$ to correct the marginal's mean; they appear in the last column of $\bSigma_{*}$ in Equation (\ref{eq:sigmastar}). The correction is
\[
\< x_* \>_{p(\x, x_*)} - \mu_* =
\sum_{l\geq 3} \sum_{j\neq n}
\frac{\Sigma_{*j}}{\Sigma_{jj}} \frac{c_{l+1,j} c_{ln}} {l!}
\left(\frac{\Sigma_{j n}}{\Sigma_{jj}\Sigma_{nn}} \right)^l + \cdots \ .
\]
The sum over pairs $j\neq n$ include the added dimension $*$, and thus pairs $(j, *)$ and $(*, n)$.
The cumulants for this problem, used both for EP and correcting it, are derived in Appendix \ref{sec:uniform-cumulants}.

\subsubsection{Predictive Corrections}

\begin{figure}[t]
\centering
\epsfig{file=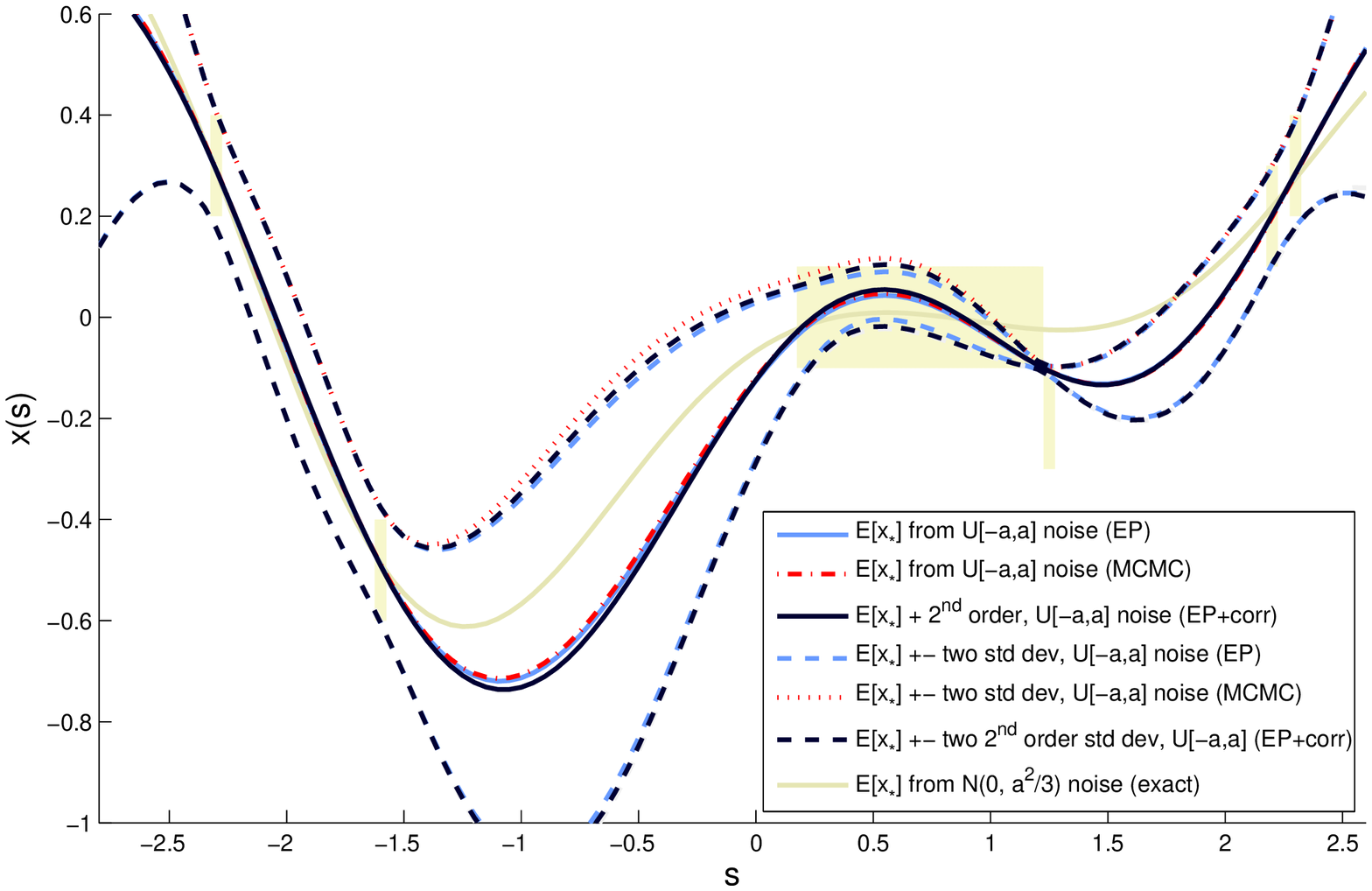, width=0.75\linewidth} \\
\epsfig{file=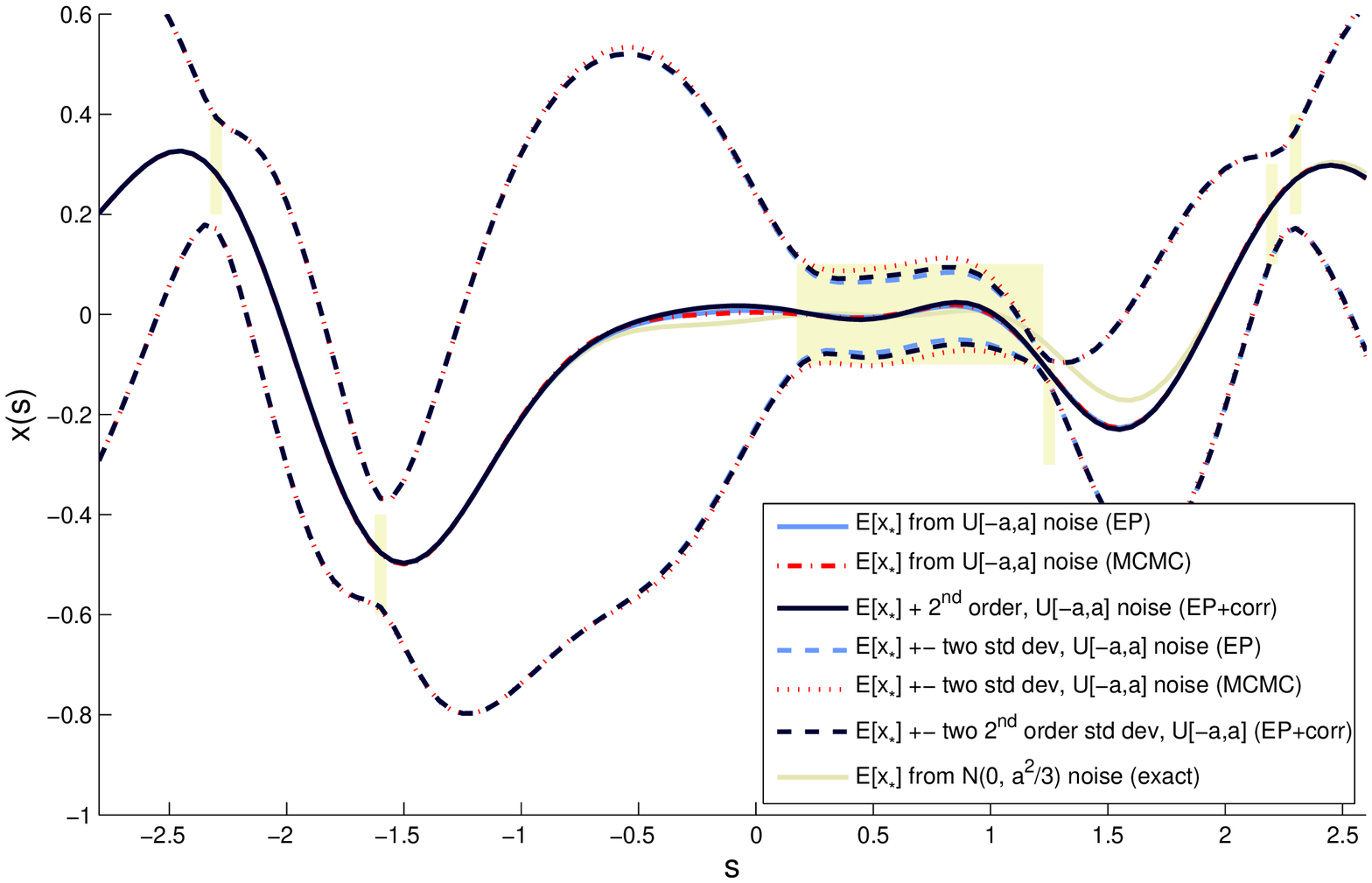, width=0.75\linewidth}
\caption{Predicting $x(s_{*})$ with a GP.
The ``boxed'' bars indicate the permissible $x(s_n)$ values; they are linked to observations $y_n$ through the uniform likelihood $\mathbb{I}[|x_n - y_n| < a]$.
Due to the ${\cal U}[-a,a]$ noise model, $q(x_{*})$ is ambivalent to where in the ``box'' $x(s_{*})$ is placed. A second order correction to the mean of $q(x_{*})$ is shown in a dotted line.
The lightly shaded function plots $p(x_{*})$, \emph{if} the likelihood was also Gaussian with variance matching that of the ``box''.
In the \emph{top} figure both the prior amplitude $\theta$ and lengthscale $\ell$ are overestimated.
In the \emph{bottom} figure, $\theta$ and $\ell$ were chosen by maximizing $\log Z_{\EP}$ with respect to their values. Notice the smaller EP approximation error.
} \label{fig:uniformlikelihood}
\end{figure}

In Figure \ref{fig:uniformlikelihood} we investigate the predictive mean correction for two cases, one where the data cannot realistically be expected to appear under the prior, and the other where the prior is reasonable.
For $s \in \mathbb{R}$, the values of $x(s_{*})$ are predicted using a GP with squared exponential covariance function
$k(s,s') = \theta \exp( - \frac{1}{2} (s - s')^2 / \ell)$.

In the first instance, the prior amplitude $\theta$ and lengthscale $\ell$
are deliberately set to values that are too big; in other words, a typical sample from the prior would not match the observed data.
We illustrate the posterior marginal $q(x_{*})$, and using Equations (\ref{eq:marginal-mean}) and (\ref{eq:marginal-covariance}), show visible corrections to its mean and variance.\footnote{In the correction for the mean in Equation (\ref{eq:marginal-mean}), we used $l = 3$ and $l = 4$ in the second order correction.
For the correction to the variance in Equation (\ref{eq:marginal-covariance}), we used $l = 3$ in the first sum, and $l = 3$ and $l = 4$ in the second sum.}
For comparison, Figure \ref{fig:uniformlikelihood} additionally shows what the predictive mean would have been were $\{ y_n \}$ observed under Gaussian noise with the same mean and variance as ${\cal U}[-a, a]$: it is substantially different.

In the second instance, $\log Z_{\EP}$ is maximized with respect to the covariance function hyperparameters $\theta$ and $\ell$ to get a kernel function that more reasonably describes the data. The correction to the mean of $q(s_{*})$ is much smaller, and furthermore, generally follows the ``Gaussian noise'' posterior mean.
When the observed data is not typical under the prior, the correction to $\< x_{*} \>$ is substantially bigger than when the prior is representative of the data.

\subsubsection{Underestimating the Truth}

\begin{figure}[t]
\centering
\epsfig{file=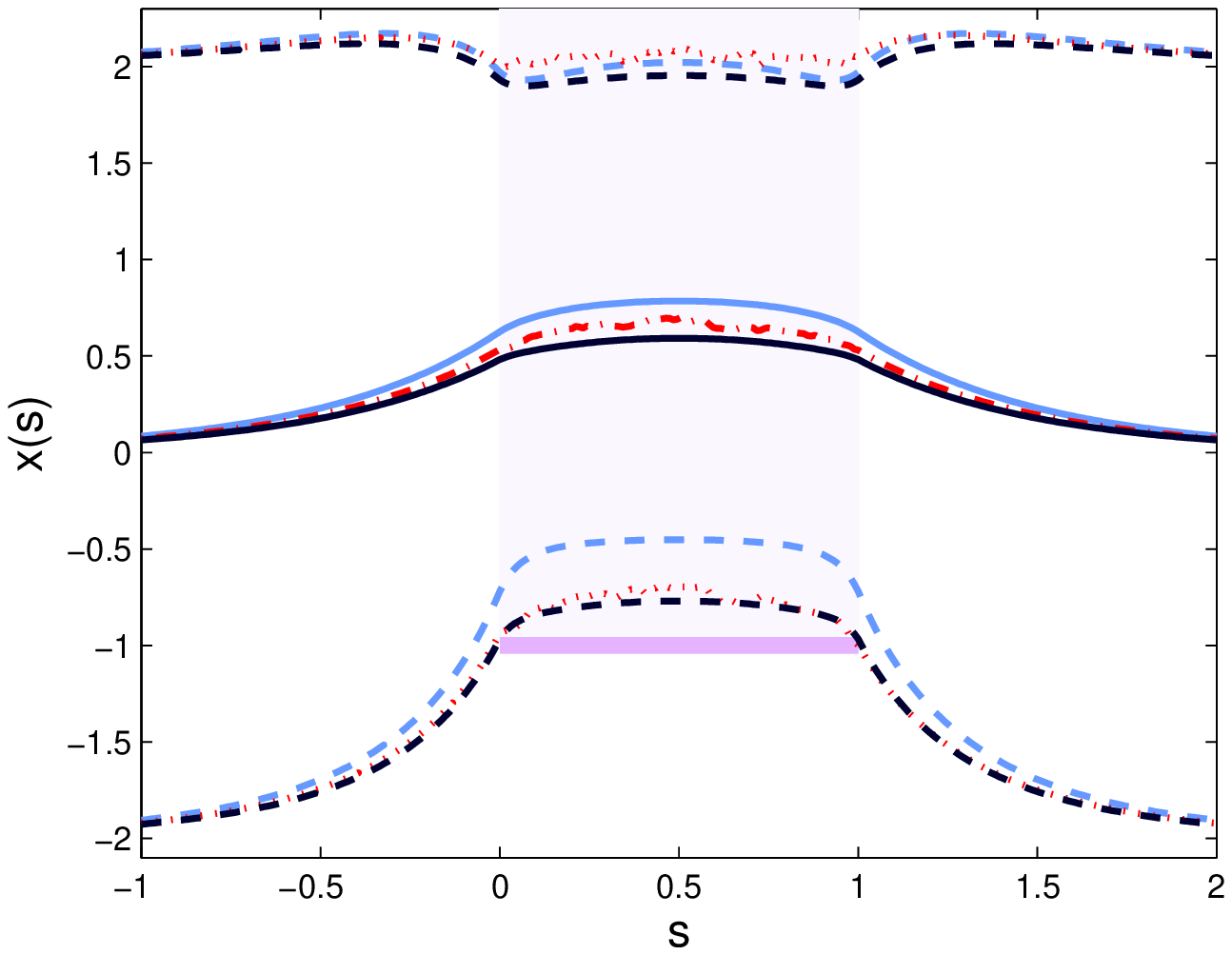, width=0.49\linewidth}
\epsfig{file=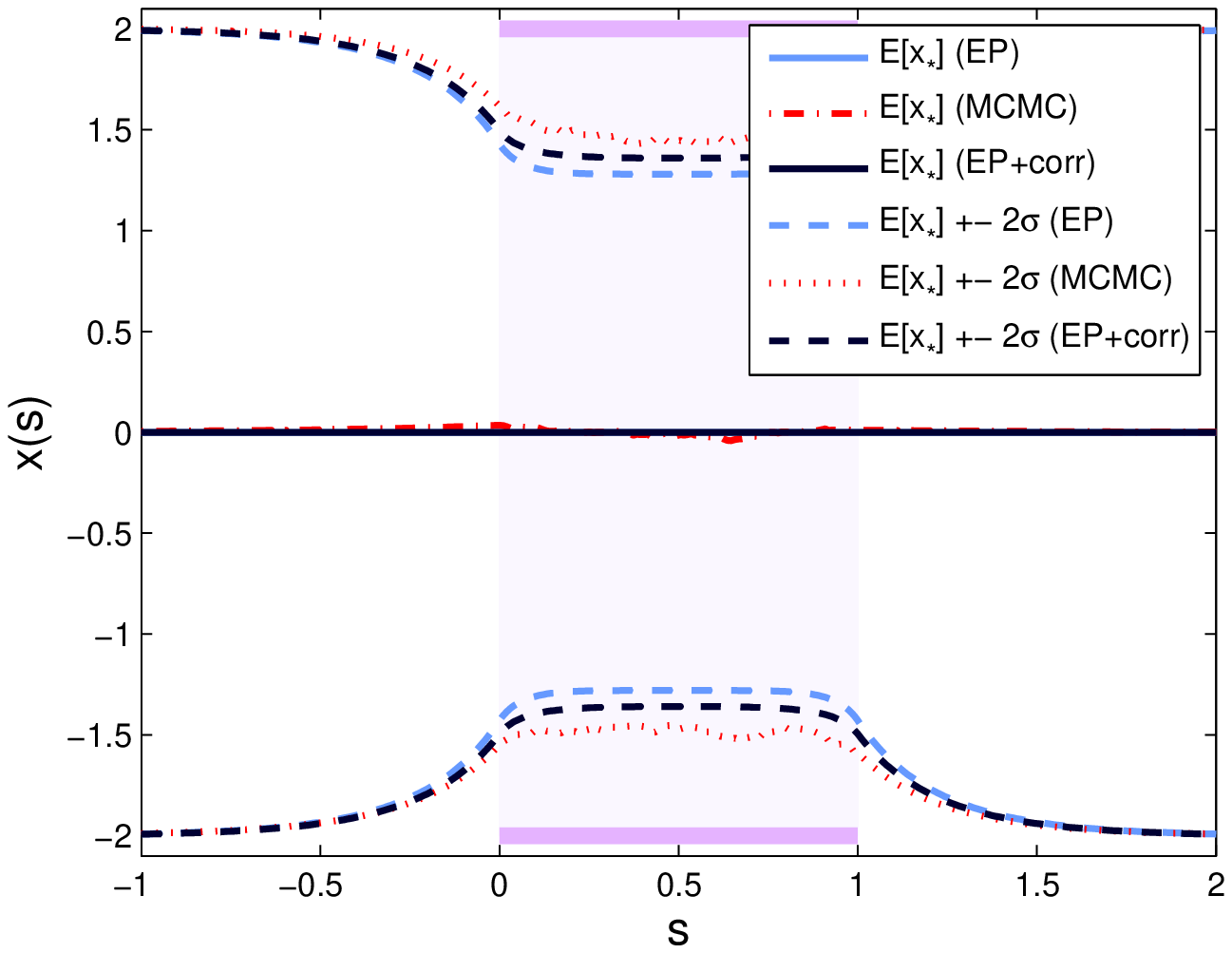, width=0.49\linewidth}
\caption{Predicting $x(s_{*})$ with a GP with $k(s,s') = \exp \{ - | s - s'| / 2 \ell\}$ and $\ell = 1$. In the \emph{left} figure $\log R_{\MCMC} = 0.41$, while the second order correction estimates it as $\log R \approx 0.64$.
On the \emph{right}, the correction to the variance is not as accurate as that on the \emph{left}. The \emph{right} correction is $\log R_{\MCMC} = 0.28$, and its discrepancy with $\log R \approx 0.45$ (EP+corr) is much bigger.
} \label{fig:boxes}
\end{figure}

Under closer inspection, the variance in Figure \ref{fig:uniformlikelihood} is slightly underestimated in regions where there are many close box constraints $|x_n - y_n| < a$. However, under sparser constraints relative to the kernel width, EP accurately estimates the predictive mean and variance. In Figure \ref{fig:boxes} this is taken further: for $N = 100$ uniformly spaced inputs $s \in [0,1]$, it is clear that $q(\x)$ becomes too narrow. The second order correction, on the other hand, provides a much closer estimate to the ground truth.

One might inquire about the behavior of the EP estimate as $N \to \infty$ in Figure \ref{fig:boxes}. In the next section, this will be used as a basis for illustrating a special case where $\log Z_{\EP}$ \emph{diverges}.

\subsection{Gaussian Process in a Box} \label{sec:gpbox}

\begin{figure}[t]
\centering
\epsfig{file=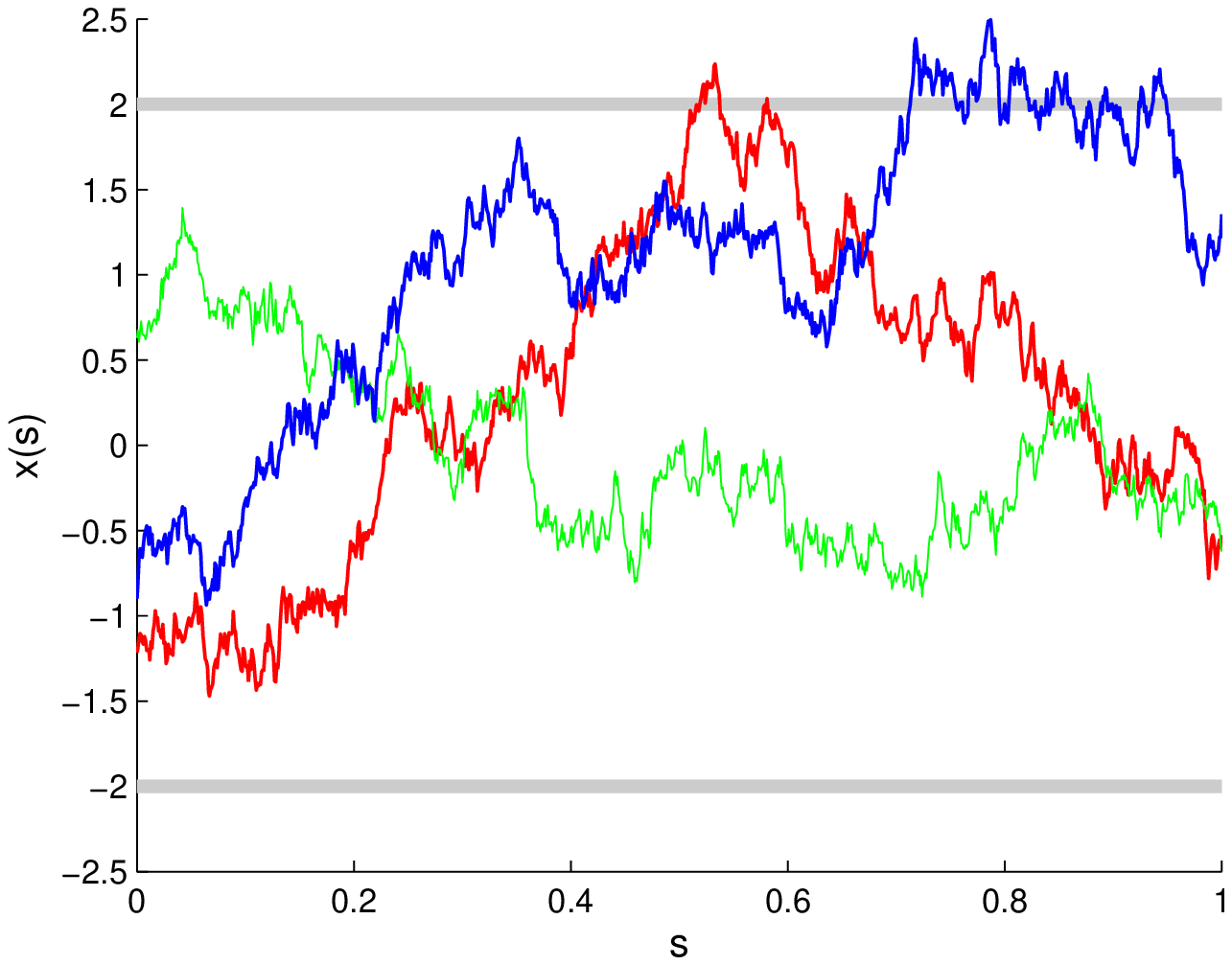, width=0.48\linewidth}
\epsfig{file=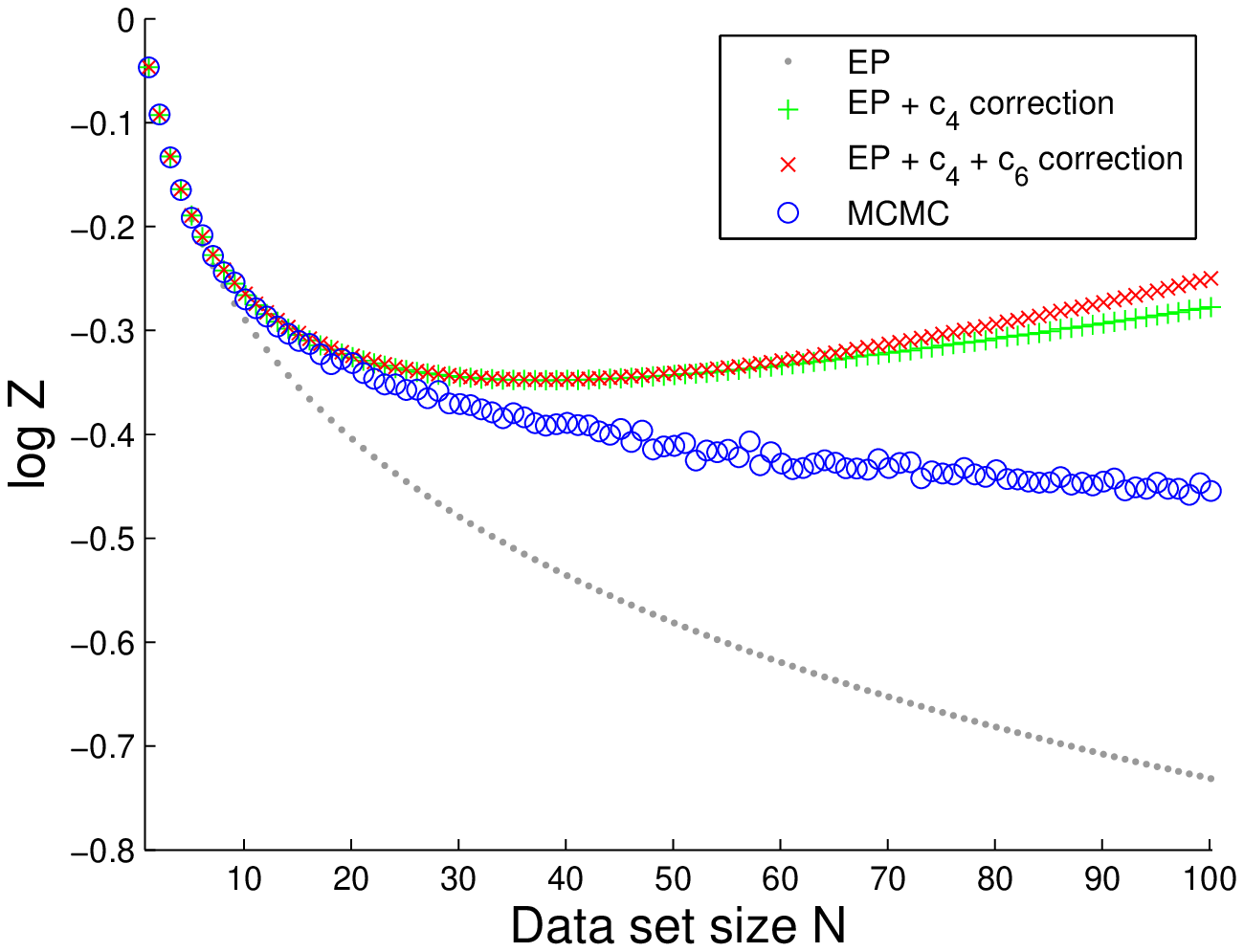, width=0.48\linewidth} \\
\epsfig{file=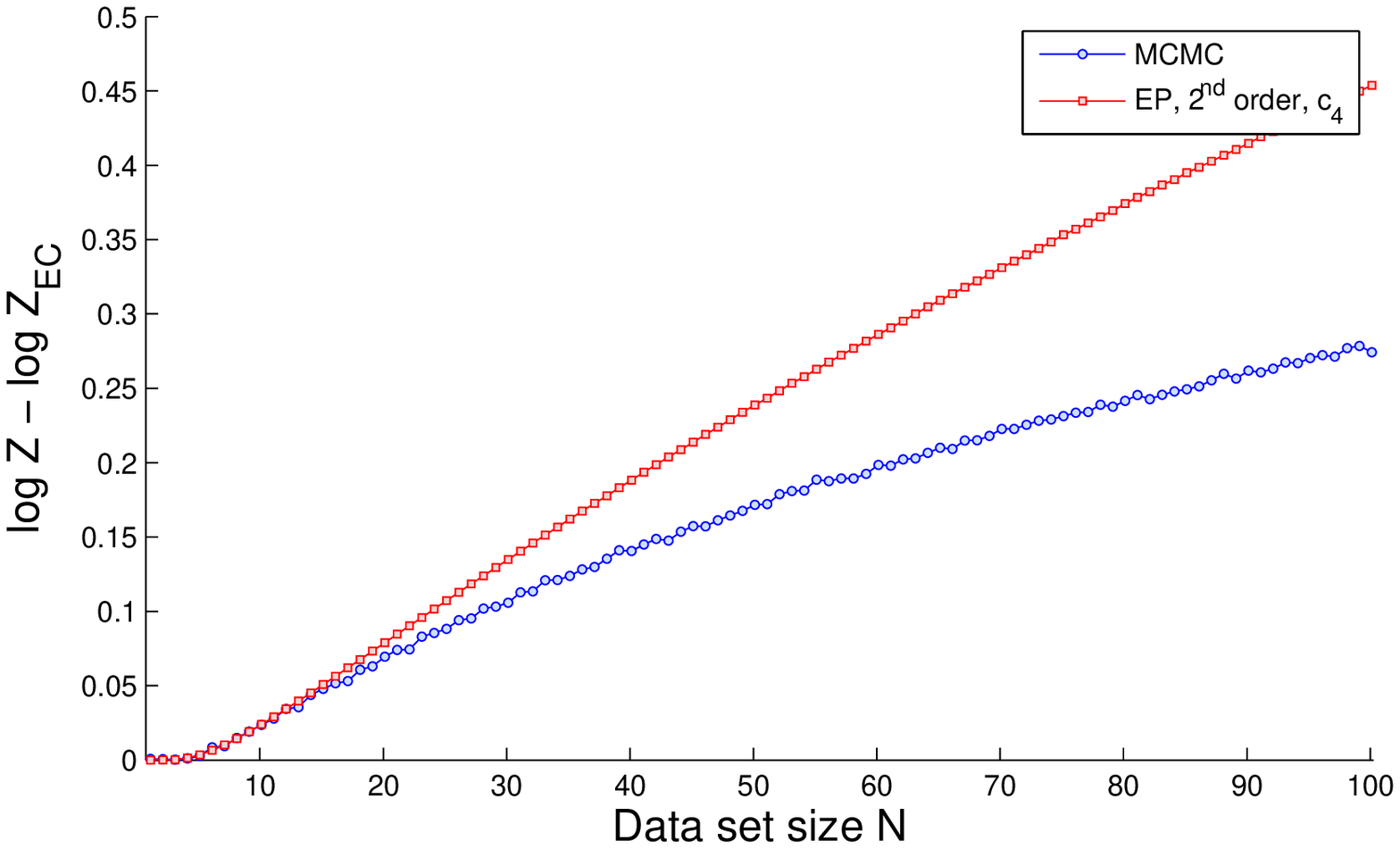, width=0.55\linewidth}
\epsfig{file=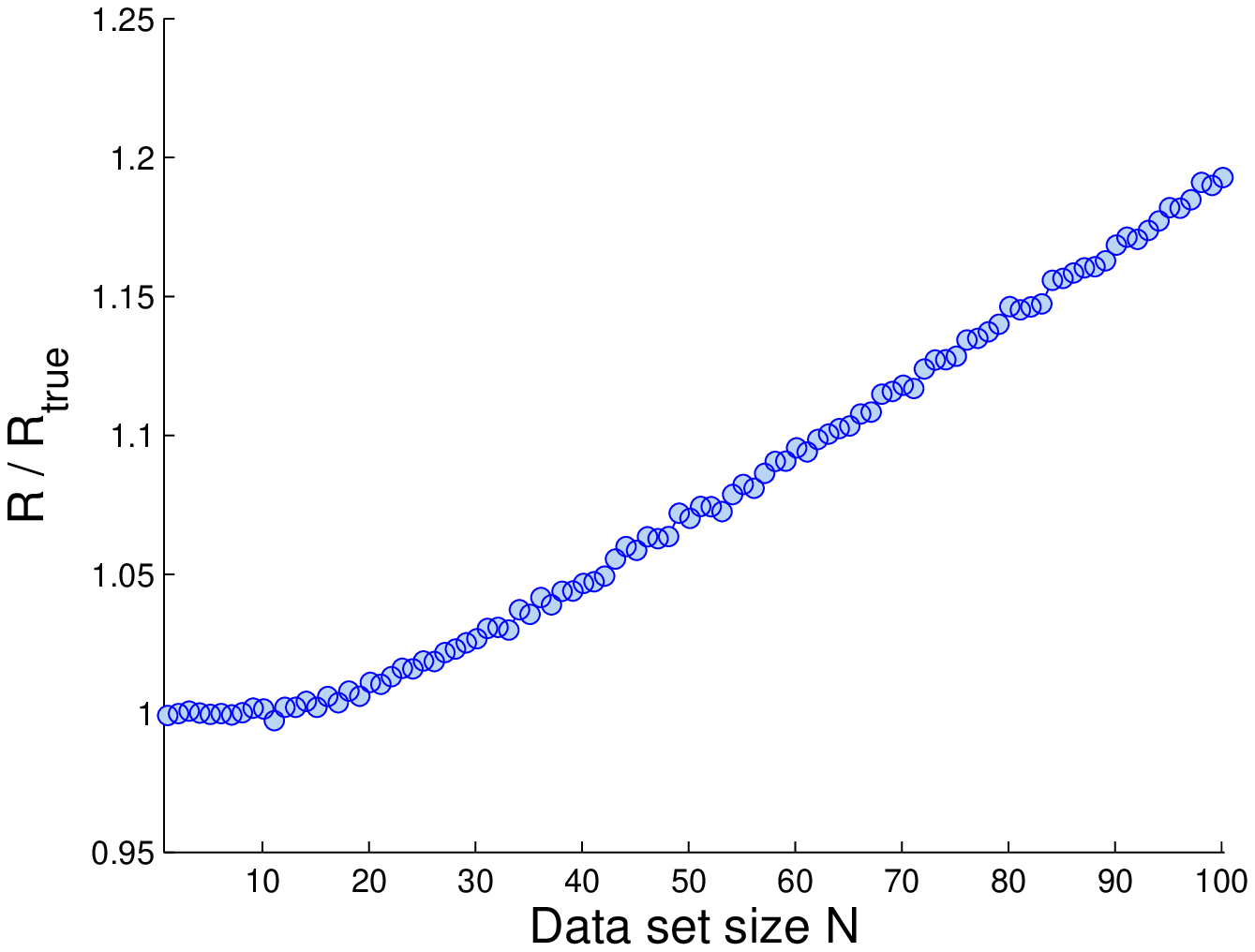, width=0.43\linewidth}
\caption{Samples from a GP prior with kernel
$k(s,s') = \exp \{ - | s - s'| / 2 \ell\}$ with $\ell = 1$, two of which are not contained in the $[-a,a]$ interval, are shown \emph{top left}. As
$N$ increases in Equation (\ref{eq:gpbox}), with $s_n \in [0,1]$, $\log Z_{\EP}$ diverges, while $\log Z$ converges to a constant. This is shown \emph{top right}. The $+$'s and $\times$'s  indicate the inclusion of the fourth ($+$) and fourth and sixth ($\times$) cumulants from the $2^{\mathrm{nd}}$ order in Equation (\ref{eq:logR-factorized-corr}) (an arrangement by total order would include $3^{\mathrm{rd}}$ order $c_4$--$c_4$--$c_4$ in $\times$).
\emph{Bottom left} and \emph{right} show the growth for $2^{\mathrm{nd}}$ order $c_4$ correction relative to the exact correction.
} \label{fig:gpbox}
\end{figure}

\begin{figure}[t]
\centering
\epsfig{file=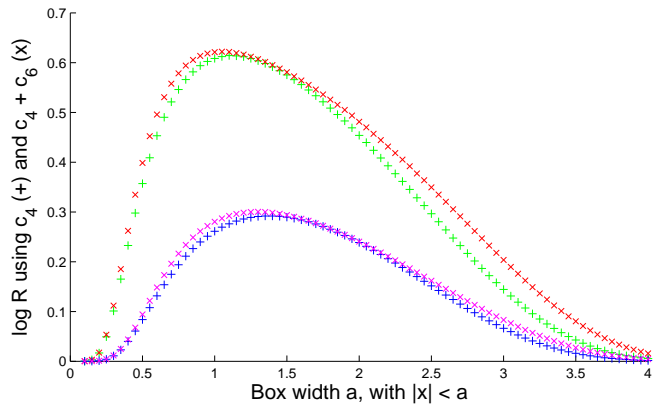, width=0.65\linewidth}
\caption{The accurateness of $\log Z_{\EP}$ depends on the size of the $[-a, a]$ box relative to $\ell$, with the estimation being exact as $a \to 0$ and $a \to \infty$. The second order correction for Figure \ref{fig:gpbox}'s kernel is illustrated here over varying $a$'s.
The $+$'s and $\times$'s  indicate the inclusion of the $4^{\mathrm{th}}$ ($+$) and $4^{\mathrm{th}}$ and $6^{\mathrm{th}}$ ($\times$) cumulants in  Equation (\ref{eq:logR-factorized-corr}). Of these, the top pair of lines are for $N=100$, and the bottom pair for $N=50$.} \label{fig:gpgrowth}
\end{figure}

In the following insightful example---a special case of uniform noise regression---$\log Z_{\EP}$ diverges from the ground truth with more data.
Consider the ratio of functions $x(s)$ over $[0,1]$, drawn from a GP prior with kernel $k(s, s')$, such that $x(s)$ lies within the $[-a, a]$ box.
Figure \ref{fig:gpbox} illustrates three random draws from a GP prior, two of which are not contained in the $[-a, a]$ interval.
The ratio of functions contained in the interval is equal to the normalizing constant of
\begin{equation} \label{eq:gpbox}
p(\x) = \frac{1}{Z} \prod_{n} \mathbb{I} \Big[ |x_n| < a \Big] \, {\cal N}( \x \, ; \, \mathbf{0}, \, \K) \ .
\end{equation}
The fraction of samples from the GP prior that lie inside $[-a,a]$ shouldn't change as the GP is sampled at increasing granularity of inputs $s$. As Figure \ref{fig:gpbox} illustrates, the MCMC estimate of $\log Z$ converges to a constant as $N \to \infty$. The EP estimate $\log Z_{\EP}$, on the other hand, diverges to $- \infty$. (The cumulants that are required for the correction in Equation (\ref{eq:logR-factorized-corr}), and recipes for deriving them, are given in Appendix \ref{sec:gpbox-cumulants}.) Of course the correction also depends on the value $a$ chosen. Figure
\ref{fig:gpgrowth} shows that for both $a \to 0$ and $a \to \infty$ the correction is zero for large $N$.

An intuitive explanation, due to Philipp Hennig, takes a one-dimensional model $p(x) = \mathbb{I}[ |x| < a]^N \, {\cal N}( x \, ; \, 0, \, 1)$. A fully-factorized approximation therefore has $N-1$ \emph{redundant} factors, as removing them doesn't change $p(x)$. However, each additional $\mathbb{I}[ |x| < a]$ truncates the estimate, forcing EP to further reduce the variance of $q(x)$.
The EP estimate using $N$ factors $\mathbb{I}[ |x| < a]^{1/N}$ is correct (see Appendix \ref{sec:one-dim-example} for a similar example and analysis), even though the original problem remains unchanged.
Even though this immediate solution cannot be applied to Equation (\ref{eq:gpbox}), the \emph{redundancy} across factors could be addressed by a principled junction tree-like factorization, where tuples of ``neighboring'' factors can be co-treated. Although beyond the scope of this paper, Appendix \ref{app:gaussian-examples} gives a guideline on how to structure such an approximation.

\section{Ising Model Results} \label{sec:isingresults}

This section discusses various aspects of corrections to EP as applied to the Ising model---a Bayesian network with binary variables and pairwise potentials---in Equation (\ref{eq:isingmodel}).

We consider the set-up proposed by \citet{wainwright2006log} in which $N=16$ nodes are either fully connected or connected to their nearest neighbors in a 4-by-4 grid.
The external field (observation) strengths $\theta_i$ are drawn from a
{\it uniform} distribution $\theta_i \sim {\cal U}[-d_{\rm
obs},d_{\rm obs}]$ with $d_{\rm obs}=0.25$. Three types of
coupling strength statistics are considered: repulsive
(anti-ferromagnetic) $J_{ij} \sim {\cal U}[-2d_{\rm coup},0]$,
mixed  $J_{ij} \sim {\cal U}[-d_{\rm coup},+d_{\rm coup}]$, and
attractive (ferromagnetic) $J_{ij} \sim {\cal U}[0,+2d_{\rm
coup}]$.

Previously we have shown \citep{opper2005expectation} that EP/EC gives very competitive results compared to several standard methods. In Section \ref{WJ} we are interested in investigating whether a further improvement is obtained with the cumulant expansion.
In Section \ref{sec:varepsilon}, we revisit the correction approach proposed in \citet{paquet09perturbation} and make and empirical comparison with the cumulant approach.

\subsection{Cumulant Expansion} \label{WJ}

For the factorized approximation we use Equations
(\ref{eq:factorized2ndOrder})
and (\ref{eq:marginal-mean}) for the $\log Z$ and marginal corrections, respectively. The expression for the cumulants of the Ising model is given in Section \ref{sec:isingexample}. The derivation of the corresponding tree expressions may be found in Appendices \ref{app:tree} and \ref{sec:two-variable-ising}.

Table \ref{table:wainwright_setup-marginal} gives the average absolute deviation (AAD) of marginals
$$
{\rm AAD} = \frac{1}{N}\sum_i \Big| p(x_i = 1) - p(x_i = 1|{\rm method}) \Big| = \frac{1}{2N} \sum_{i}
\big| m_{i}-m^{\rm est}_{i} \big| \ ,
$$
while Table \ref{table:wainwright_setup-logZ}
gives the absolute deviation of $\log Z$  averaged of 100 repetitions. In two cases (Grid, $d_{\rm coup}=2$ Repulsive and Attractive coupling) we observed some numerical problems with the EC tree solver. 
It might be some cases that a solution does not exist but we ascribe numerical instabilities in our implementation as the main cause for these problems. It is currently out of the scope of this work to come up with a better solver. We choose to report the average performance for those runs that could attain a high degree of expectation consistency: $\sum_{i=1}^N ( \< x_i \>_{q_i} - \< x_i \>_{q} ) ^2 \leq 10^{-20}$. This was 69 out of 100 in the mentioned cases and 100 of 100 in the remaining.
\begin{table}[t]
\begin{center}
\begin{small}
\begin{tabular}{|c|c|c|c|c|c|c|c|} \hline
\multicolumn{3}{|c|}{Problem type} & \multicolumn{5}{c|}{AAD marginals}  \\
\hline Graph & Coupling & $d_{\rm coup}$ & LBP & LD & EC & EC c & EC t \\
\hline \hline
\multirow{6}{*}{Full}
 & \multirow{2}{*}{Repulsive}
   & 0.25 & .037 & .020 & .003 & {\bf .0006} & .0017  \\ \cline{3-8}
 & & 0.50 & .071 & .018 & .031 & .0157 & {\bf .0143}  \\ \cline{2-8}
 & \multirow{2}{*}{Mixed}
   & 0.25 & .004 & .020 & .002 & {\bf .0004} & .0013  \\ \cline{3-8}
 & & 0.50 & .055 & .021 & .022 & .0159 & {\bf .0151}  \\ \cline{2-8}
 & \multirow{2}{*}{Attractive}
   & 0.06 & .024 & .027 & .004 & {\bf .0023} & .0025  \\ \cline{3-8}
 & & 0.12 & .435 & .033 & .117 & .1066 & {\bf .0211}  \\ \hline \hline
\multirow{6}{*}{Grid}
 & \multirow{2}{*}{Repulsive}
   & 1.0  & .294 & .047 & .153 & {\it .1693} & {\bf .0031} \\ \cline{3-8}
 & & 2.0  & .342 & .041 & .198 & {\it .4244} & {\bf .0021} \\ \cline{2-8}
 & \multirow{2}{*}{Mixed}
   & 1.0  & .014 & .016 & .011 & {\it .0122} & {\bf .0018} \\ \cline{3-8}
 & & 2.0  & .095 & .038 & .082 & {\it .0984} & {\bf .0068} \\ \cline{2-8}
 & \multirow{2}{*}{Attractive}
   & 1.0  & .440 & .047 & .125 & {\it .1759} & {\bf .0028} \\ \cline{3-8}
 & & 2.0  & .520 & .042 & .177 & {\it .4730} & {\bf .0002} \\ \hline
\end{tabular}
\end{small}
\end{center}
\caption{Average absolute deviation (AAD) of marginals in a Wainwright-Jordan set-up, comparing loopy
belief propagation (LBP), log-determinant relaxation (LD), EC, EC with
$l=4$ second order correction (EC c), and an EC tree (EC t). Results in bold face highlight best results, while italics indicate where the cumulant expression is less accurate than the original approximation.}
\label{table:wainwright_setup-marginal}
\end{table}

\begin{table}[t]
\begin{center}
\begin{small}
\begin{tabular}{|c|c|c|c|c|c|c|c|} \hline
\multicolumn{3}{|c|}{Problem type} & \multicolumn{5}{c|}{Absolute deviation $\log Z$} \\
\hline Graph & Coupling & $d_{\rm coup}$ & EC & EC c & EC $\varepsilon$c & EC t & EC tc \\
\hline \hline
\multirow{6}{*}{Full}
 & \multirow{2}{*}{Repulsive}
   & 0.25 & .0310  &  .0018 & .0061 & .0104 & {\bf .0010} \\ \cline{3-8}
 & & 0.50 & .3358  &  .0639 & .0697 & .1412 & {\bf .0440} \\ \cline{2-8}
 & \multirow{2}{*}{Mixed}
   & 0.25 & .0235  &  .0013 & .0046 & .0129 & {\bf .0009} \\ \cline{3-8}
 & & 0.50 & .3362  &  .0655 & .0671 & .1798 & {\bf .0620} \\ \cline{2-8}
 & \multirow{2}{*}{Attractive}
   & 0.06 & .0236  &  .0028 & .0048 & .0166 & {\bf .0006} \\ \cline{3-8}
 & & 0.12 & .8297  &  {\bf .1882} & .2281 & .2672 & .2094 \\ \hline \hline
\multirow{6}{*}{Grid}
 & \multirow{2}{*}{Repulsive}
   & 1.0  & 1.7776 &  .8461 & .8124 & .0279 & {\bf .0115} \\ \cline{3-8}
 & & 2.0  & 4.3555 & 2.9239 & 3.4741 & .0086 & {\bf .0077} \\ \cline{2-8}
 & \multirow{2}{*}{Mixed}
   & 1.0  & .3539  &  .1443 & .0321 & .0133 & {\bf .0039} \\ \cline{3-8}
 & & 2.0  & 1.2960 &  .7057 & .4460 & .0566 & {\bf .0179} \\ \cline{2-8}
 & \multirow{2}{*}{Attractive}
   & 1.0  & 1.6114 &  .7916 & .7546 & .0282 & {\bf .0111} \\ \cline{3-8}
 & & 2.0  & 4.2861 & 2.9350 & 3.4638 & .0441 & {\bf .0433} \\ \hline
\end{tabular}
\end{small}
\end{center}
\caption{Absolute deviation log partition function in a Wainwright-Jordan set-up, comparing EC, EC with
$l=4$ second order correction (EC c),
EC with a full second order $\varepsilon$ expansion (EC $\varepsilon$c),
EC tree (EC t) and EC tree with $l=4$ second order correction (EC tc). Results in bold face highlight best results. The cumulant expression is consistently more accurate than the original approximation.}
\label{table:wainwright_setup-logZ}
\end{table}

\begin{figure}[t]
\centering
\epsfig{file=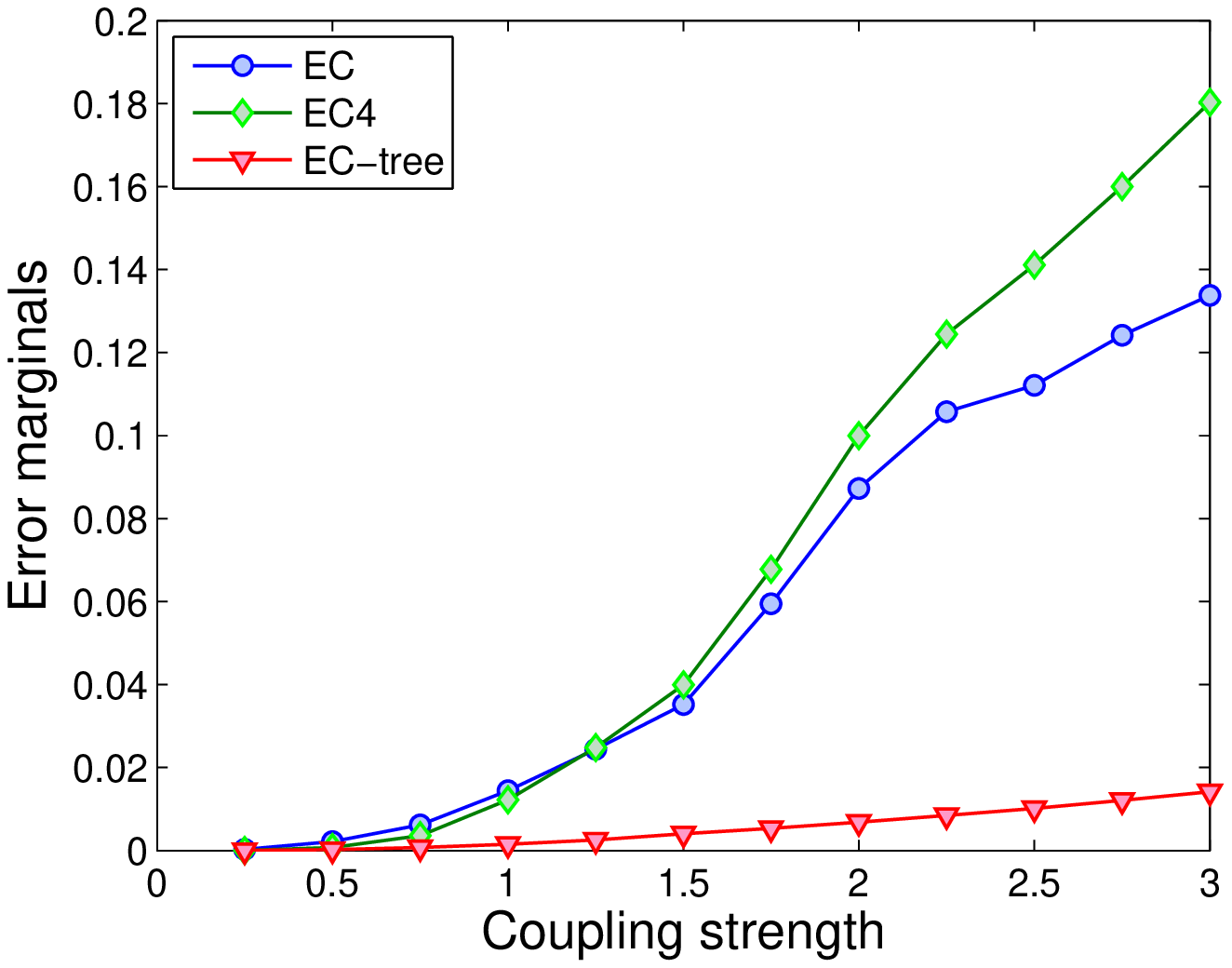, width=0.48\linewidth}
\epsfig{file=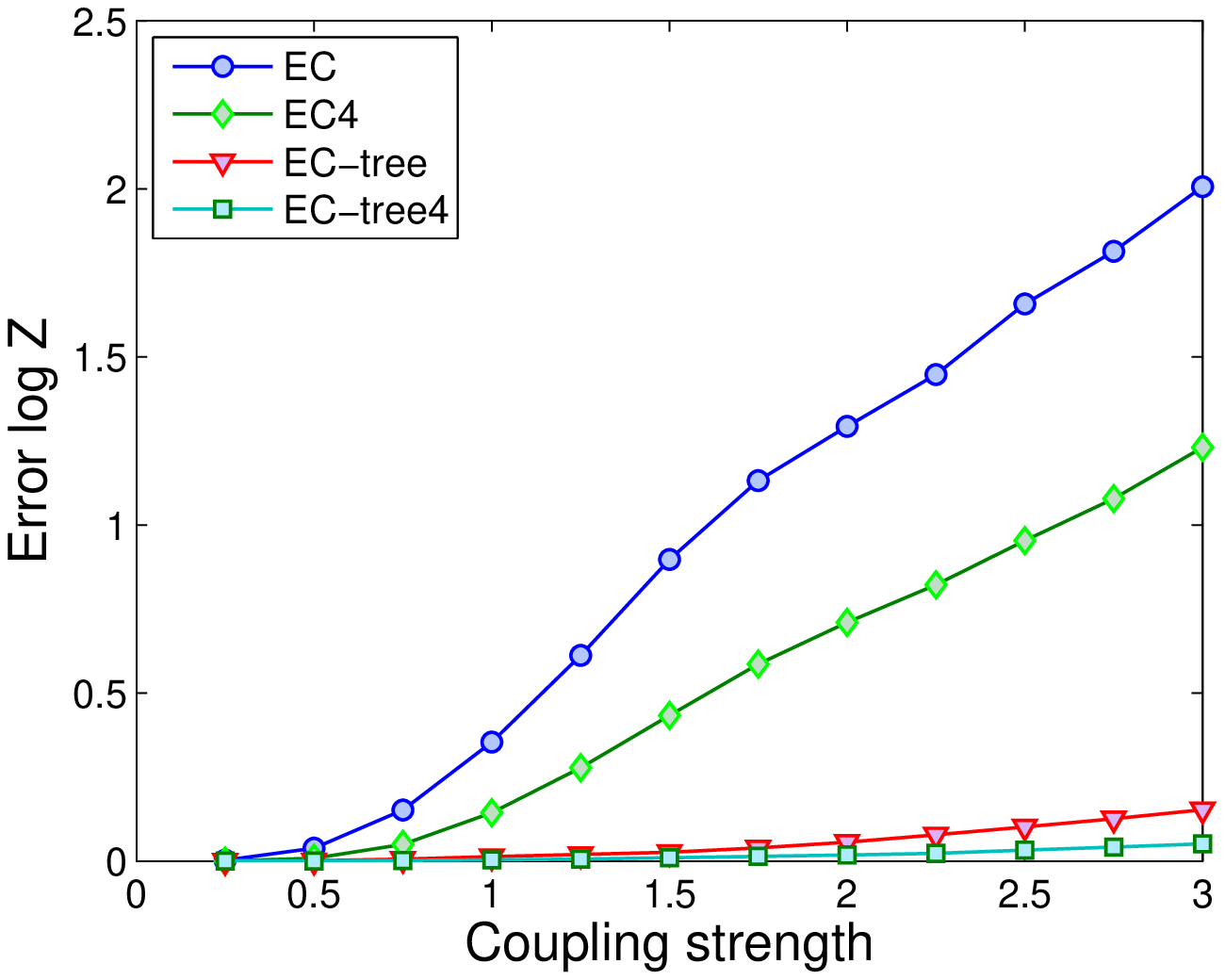, width=0.48\linewidth}
\caption{Error on marginal \emph{(left)} and $\log Z$ \emph{(right)} for grid and mixed couplings as a function of coupling strength.} \label{fig:ising_dcoup}
\end{figure}

We observe that for the Grid simulations, the corrected marginals in factorized approximation are less accurate than the original approximation. In Figure \ref{fig:ising_dcoup} we vary the coupling strength for a specific set-up (Grid Mixed) and observe a cross-over between the correction and original for the error on marginals as the coupling strength increases. We conjecture that when the error of the original solution is high then the number of terms needed in the cumulant correction increases. The estimation of the marginal seems more sensitive to this than the $\log Z$ estimate. The tree approximation is very precise for the whole coupling strength interval considered and the fourth order cumulant in the second order expansion is therefore sufficient to get often quite large improvements over the original tree approximation.

\subsection{The $\varepsilon$-Expansion} \label{sec:varepsilon}

In \citet{paquet09perturbation} we introduced an alternative expansion for $R$ and applied it to Gaussian processes and mixture models. It is obtained from Equation (\ref{correc1}) using a finite series expansion, where the normalized deviation
$$\varepsilon_n(x_n) = \frac{q_n(x_n)}{q(x_n)} -1$$
is treated as the small quantity instead of higher order cumulants. $R$ has an exact representation with $2^N$ terms that we may truncate at lowest non-trivial order:
\begin{align*}
& R = \< \prod_n\left(1 + \varepsilon_n(x_n)\right) \>_{q(\x)}
\approx  1 + \sum_{m < n} \< \varepsilon_m(x_m)
\varepsilon_n(x_n) \> + {\cal O} (\varepsilon^3) \ .
\end{align*}
The linear terms are all equal to one because $ \< \frac{q_n(x_n)}{q(x_n)} \>_{q}=\int q(x_n) \frac{q_n(x_n)}{q(x_n)} \, \mathrm{d} x_n =1$ and since $q_n(x_n)$ is a binary distribution the quadratic term becomes a weighted sum of ratios of Normal distributions:
$$
\< \frac{q_m(x_m)}{q(x_m)} \>_{q(\x)} =  \sum_{x_n,x_m = \pm 1} \frac{1+x_m m_m}{2} \frac{1+x_n m_n}{2} \frac{q(x_m,x_n)}{q(x_m)q(x_n)} \ .
$$
The final expression for the lowest order approximation to $R$ is then
\[
R \approx 1 + \sum_{m < n} \sum_{x_n,x_m = \pm 1} \frac{1+x_m m_m}{2} \frac{1+x_n m_n}{2} \frac{q(x_m,x_n)}{q(x_m)q(x_n)} - \frac{N(N-1)}{2} \ .
\]
From Table \ref{table:wainwright_setup-logZ} we observe an improvement over the original factorized approximation and results similar to the cumulant correction to the factorized approximation for all settings. The $\varepsilon$-expansion may also used to calculate marginals and applied to generalized factorizations. These topics will be studied elsewhere.

\section{Future Directions} \label{sec:conclusion}

Corrections to Gaussian EP approximations were examined in this paper. The Gaussian measure allowed for a convenient set of mathematical tools to be employed, mostly because it admits orthogonality of a set of polynomials, the Hermite polynomials, which allowed a clean simplification of many expressions. So far we have restricted ourselves to expansions to low
orders in cumulants. Our results indicate that these first corrections to EP can already provide useful 
information about the quality of the EP solution. Small corrections typically show that EP is fairly accurate and the corrections
improve on that. On the other hand, large corrections indicate that the EP approximation performs poorly.
The low order corrections can yield a step in the right direction but in general their result may not be trusted
and alternatives to the Gaussian EP approximation should be considered. 
It will be interesting to develop similar expansions to EP approximations with other exponential families besides the Gaussian one. 

Can we expect that higher order terms in the cumulant expansion
will give more reliable approximations? Before such a question could be attacked one first would need to decide 
in which order the terms of the expansion should be evaluated in order to obtain
the most dominant contributions. For example, we might think of trying to first compute all terms in the second order expansion
of the exponential in Equation
(\ref{eq:factorized2ndOrder}), and then move on to higher orders. An alternative is to 
sort the expansion by the total sum of the orders of cumulants involved. This is in fact possible by introducing a
suitable expansion 
parameter (which is later set equal to one) such that the formal Taylor series with respect to this parameter yields
the desired expansion. However, it is not clear yet if and when such a power series expansion would actually converge.
It may well be that our expansions are only of an asymptotic type \citep{boyd99asymptotic} for which the summation
of only a certain number of terms might give an improvement whereas further terms would lead to
worse results.
 
 We expect that such questions could at least be answered
for toy models such as the {\em Gaussian process in a box} model of Section \ref{sec:gpbox}. Our results for the  latter example 
(together with the related {\em uniform noise regression} case) indicates that EP may not be understood as 
an off the shelf method for approximately calculating arbitrary high dimensional sums or integrals.   
One may conjecture that its quality strongly depends on the fact that such sums or integrals 
may or may not have an interpretation in terms of a proper statistical inference model which contain data that are 
highly probable with respect to the model. It would be interesting to see if one can develop a theory for the average case performance of EP under such statistical assumptions of the data.

\appendix

\section{Factorizations: Gaussian Examples} \label{app:gaussian-examples}

As $p(\x)$ is a latent Gaussian model, the $g$-terms in Equation (\ref{eq:qx}) are chosen in this paper to
give a Gaussian approximation
\[
q(\x) = \frac{1}{Z_q} \exp \{ \blambda^{T} \phi(\x) \} = \Ncal(\x \, ; \, \bmu, \bSigma) \ .
\]
The sufficient statistics $\phi(\x)$ and natural parameters $\blambda$ of the Gaussian are defined as
\[
\phi(\x) = (\x, -\tfrac{1}{2} \x \x^T) \quad \textrm{and} \quad
\blambda = (\bgamma, \bLambda) \ ,
\]
where $\blambda^{T} \phi(\x) = \bgamma^T \x - \frac{1}{2} \tr[\bLambda \x \x^T] = \bgamma^T \x - \frac{1}{2} \x^{T} \bLambda \x$.
There exists a bijection between the canonical parameters $\bmu$ and $\bSigma$ and natural parameters,
such that the mean and covariance can be determined with
$\bSigma = \bLambda^{-1}$ and $\bmu = \bSigma \bgamma$.

In Equation (\ref{eq:mainquadratic}) we can define $g_0(\x) = \exp \{ \blambda_0^{T} \phi(\x) \}$, where $\blambda_0 = (\bgamma^{(0)}, \bLambda^{(0)})$, such that it is essentially a rescaling of factor $f_0$.
In the Ising model in Equation (\ref{eq:isingmodel}), this means that
$\bLambda^{(0)} = - \J$ and $\bgamma^{(0)} = \btheta$.
In the Gaussian process classification model in Equation (\ref{eq:gpcmodel}), this implies that
$\bLambda^{(0)} = \K^{-1}$ and $\bgamma^{(0)} = \mathbf{0}$.

\subsection{Term-Wise Factorizations}

It remains to define a suitable factorization for the term-product $\prod_n t_n(x_n)$. This factorization can be fully factorized, factorized over disjoint sets of variables, factorized as a tree, or follow more arbitrary factorizations (see the simple example in Appendix \ref{sec:one-dim-example}).
A few such factorizations are given below in increasing orders of complexity. \emph{In each case we do not include the $f_0$ factor for clarity.} Furthermore, even though the term factorization may be chosen to fully factorize, $q(\x)$ may be fully connected through the inclusion of $f_0$.

\subsubsection{Fully Factorized} 
A common factorization of $\prod_n t_n(x_n)$ is to set
$f_n(\x) = t_n (x_n)$.
The natural parameters of
$g_n(\x) = \exp \{ \blambda_n^T \phi(\x) \} $ are chosen to be $\blambda_n = ( \gamma_n^{(n)}, \Lambda_{nn}^{(n)})$, corresponding to $\phi_n(x_n) = (x_n, - \frac{1}{2} x_n^2)$.
For clarity the other $\gamma$ and $\Lambda$ parameters in $\blambda_n$ are not shown, as they are clamped at zero. This gives an approximation
$q(\x)$ that is defined by $\blambda = \blambda_0 + \sum_n \blambda_n$.

\subsubsection{Factorization into Disjoint Pairs}
As a second step the $N$ variables can be subdivided into \emph{disjoint}
pairs $\x_{\bpi} = (x_m, x_n)$.
The factorization over terms couples pairs of variables through
\[
\prod_n t_n(x_n) = \prod_{\bpi = (m,n)} [ \, t_m(x_m) t_n(x_n) \,] = \prod_{\bpi} f_{\bpi}(\x) \ .
\]
In this case each factor will have a contribution
$g_{\bpi}(\x) = \exp \{ \blambda_{\bpi}^T \phi(\x) \}$
to the overall approximation, and, as $g_{\bpi}$ is a function of two variables, it is
parameterized by the ``correlated Gaussian form''
$\blambda_{\bpi} = ( \gamma_m^{(\bpi)}, \gamma_n^{(\bpi)}, \Lambda_{mm}^{(\bpi)}, \Lambda_{nn}^{(\bpi)}, \Lambda_{mn}^{(\bpi)})$.
By symmetry $\Lambda_{nm}^{(\bpi)} = \Lambda_{mn}^{(\bpi)}$.
The resulting
$q(\x)$ is defined in terms of these disjoint sets with
$\blambda = \blambda_0 + \sum_{\bpi} \blambda_{\bpi}$.

\subsubsection{Tree-structured Factorization} \label{sec:treestructured-basics}
A tree structure factorization can be defined by extending the above ``disjoint pairs'' case to allow for overlaps between terms.
Let ${\cal G}$ define a spanning tree structure over all $\x$, and let $\btau = (m,n) \in {\cal G}$ define the edges in the tree. Let $d_n$ be the number of edges emanating from node $x_n$ in the graph.
Through a clever regrouping of terms into a ``junction tree'' form with
\[
\prod_n t_n(x_n) = \frac{ \prod_{\btau = (m,n)} [ \, t_m(x_m) t_n(x_n) \, ] }{ \prod_{n} t_{n}(x_n)^{d_n - 1} }
=
\frac{ \prod_{\btau} f_{\btau}(\x) }{ \prod_{n} f_{n}(\x)^{d_n - 1} } \ ,
\] 
the term-approximation will be tree-structured.
In this example the
$\facpower_a$ powers are 1 for edge factors $f_{\btau}$ and $(1 - d_n)$
for node factors $f_n$.
Let $g_{\btau}(\x)$ and $g_n(\x)$ be parameterized by $\blambda_{\btau}$
and $\blambda_n$, as was done in the two examples above. Using
\[
\frac{ \prod_{\btau} g_{\btau}(\x) }{ \prod_{n} g_{n}(\x)^{d_n - 1} }
=
\frac{ \prod_{\btau} \exp \{ \blambda_{\btau}^T \phi(\x) \} }{
\prod_{n} \exp \{ \blambda_{n}^T \phi(\x) \}^{d_n - 1}
} \ ,
\]
the resulting $q(\x)$ has parameter vector
$\blambda = \blambda_0 + \sum_{\btau} \blambda_{\btau} - \sum_n (d_n - 1) \blambda_n$.

It is useful to note that the form of the tree-structured approximation given here
is that used by \cite{opper2005expectation}; it approximates the ``junction tree'' form using a Power EP
factorization \citep{minka04power}.
The factorization and stationary condition is \emph{different} from that of Tree EP \citep{MinQi04_correct}.

\subsection{Stationary Point} \label{sec:stationary-point}

The EP moment matching conditions from Equation (\ref{eq:moments-match}) are uniquely met at the stationary point of $\log Z_{\EP}$ in Equation (\ref{eq:Z-EP}), and are shown here.
Consider the logarithm of the normalizer,
\begin{equation} \label{eq:log-Z-EP}
\log Z_{\EP} = \log Z_q + \sum_a \facpower_a \log Z_a \ .
\end{equation}
Using the sufficient statistics and natural parameters defined above, the two normalizers that constitute Equation (\ref{eq:log-Z-EP}) are
\begin{align*}
Z_q & = \int \mathrm{e}^{\sum_a \facpower_a \blambda_a^T \phi(\x)} \, \mathrm{d} \x \ , \\
Z_a & = \frac{1}{Z_q} \int \mathrm{e}^{\sum_b \facpower_b \blambda_b^T \phi(\x)
- \blambda_a^T \phi(\x)} \, f_a(\x) \, \mathrm{d} \x \ .
\end{align*}
Using these definitions, the derivatives of the terms in Equation (\ref{eq:log-Z-EP}) with respect to some EP factor $c$'s parameters $\blambda_c$ are
\begin{align*}
\frac{\partial \log Z_q}{ \partial \blambda_c} & = \facpower_c \< \phi(\x) \>_q \ , \\
\frac{\partial \log Z_a}{ \partial \blambda_c} & = 
\left\{ \begin{array}{l l}
     \facpower_c \< \phi(\x) \>_{q_a} - \facpower_c \< \phi(\x) \>_q & \quad \text{if $c \neq a$} \\
      (\facpower_c - 1) \< \phi(\x) \>_{q_c} - \facpower_c \< \phi(\x) \>_q & \quad \text{if $c = a$ \ .}
   \end{array} \right.
\end{align*}
When $\partial \log Z_{\EP} / \partial \blambda_c = \0$ for any $c$, the following therefore holds:
\[
\0 = (\facpower_c - 1) ( \< \phi(\x) \>_{q_c} - \< \phi(\x) \>_{q}) + \sum_{a \neq c} \facpower_a ( \< \phi(\x) \>_{q_a} - \< \phi(\x) \>_{q}) \ .
\]
Let $\D$ be a square matrix where the values in column $a$ are $\facpower_a$; all the rows in $\D$ are equal and it is singular. Furthermore, let $\bm{\psi}_a = \< \phi(\x) \>_{q_a} - \< \phi(\x) \>_{q}$. By stacking all the $\bm{\psi}_a$'s into a column vector $\bm{\psi}$, the above set of equalities lead to a system of equations
\[
\0 = ((\D - \I) \otimes \I_{\mathrm{dim}}) \, \bm{\psi} \ .
\]
(The Kronecker product is only required as the sufficient statistics' differences $\bm{\psi}_a$ have dimensionality ``dim'', usually larger than one.)
As $\D - \I$ is nonsingular, it is solved by $\bm{\psi} = \0$, and hence $\< \phi(\x) \>_{q_a} = \< \phi(\x) \>_{q}$ for all $a$.

The choice of parameterization of $\blambda_a$ might give an overcomplete representation, and the exact moment-matching
conditions $\< \phi(\x) \>_{q_a} = \< \phi(\x) \>_{q}$ might have more than one unique solution. However, this does not invalidate that at the stationary point of Equation (\ref{eq:log-Z-EP}), all moment-matching conditions must hold.

\section{Tree-Structured Approximation} \label{app:tree}

Let the factorization of the term-product $\prod_n t_n(x_n)$ take the form of a tree $\cal G$ with edges $\btau = (m,n) \in \cal G$, as is described in Appendix \ref{sec:treestructured-basics}.
The number connections to a node or vertex $n$ shall be denoted by $d_n$.
From Equation (\ref{eq:generalLogR}) the second order expansion is
\begin{align}
\log R & =
\frac{1}{2} \sum_{\btau \neq \btau'} \< \< r_{\btau} \> \< r_{\btau'} \>\>
+ \frac{1}{2} \sum_{m \neq n} (1 - d_{m}) (1 - d_{n}) \< \< r_{m} \> \< r_{n} \> \>
\nonumber \\
& \quad
+ \sum_{\btau, n} (1 - d_n) \< \< r_{\btau} \> \< r_{n} \> \>
+ \frac{1}{2} \sum_{n} (1 - d_{n})(- d_{n}) \< \< r_{n} \>^2 \> + \cdots \ , \label{eq:treesecondorder}
\end{align}
where the inner expectations are over $\k_{\btau} | \x$ and $k_n | \x$, while the outer expectations are over $\x$.\footnote{Some readers might wonder why there is no $\frac{1}{2}$ associated with the sum over $(\btau, n)$ in Equation (\ref{eq:treesecondorder}). In the other quadratic sums, for example over $m \neq n$, each $(m,n)$ pair appears twice, as $r_m r_n$ and as $r_n r_m$. Each edge-node pair makes only one appearance in the sum; if the sum double-counted by including node-edge pairs, a division by two would have been necessary.}
The edge-edge, edge-node, and node-node expectations that are needed in Equation (\ref{eq:treesecondorder}) are given in the following three sections.

\subsection{Edge-Edge Expectations} \label{sec:edge-edge}

The edge-edge expectation provides a beautiful illustration of the combinatorics that may be involved in Wick's theorem. For $\btau \neq \btau'$, the following expectation needs to be evaluated:
\begin{align} 
& \< \< r_{\btau} (\k_{\btau}) \> \< r_{\btau'} (\k_{\btau'}) \> \> \nonumber \\
& \qquad = \<
\sum_{l \ge 3} \sum_{s \ge 3} i^{l + s}
\left\{ \sum_{|\balpha| = l}
\, \frac{ c_{\balpha \btau} }{ \balpha ! } \,
\< \k_{\btau}^{\balpha} \>_{ \k_{\btau} | \x } \right\}
\left\{ \sum_{|\balpha'| = s}
\, \frac{ c_{\balpha' \btau'} }{ \balpha' ! } \,
\< \k_{\btau'}^{\balpha'} \>_{\k_{\btau'} | \x } \right\}
\>_\x \ . \label{eq:rr}
\end{align}
The vectors $\balpha$ that are summed over to get $|\balpha| = l$ are $\balpha = (0,l), (1, l-1), \ldots, (l, 0)$; let $\balpha = (\alpha_1, l - \alpha_1)$ when $|\balpha| = l$.
From the independence of $\k_{\btau} | \x$ and $\k_{\btau'} | \x$,
\begin{equation} \label{eq:localk}
\<
\E{ \k_{\btau}^{\balpha} }_{\k_{\btau} | \x} \,
\E{ \k_{\btau'}^{\balpha'} }_{\k_{\btau'} | \x}
\>_\x
=
\<
\E{ \k_{\btau}^{\balpha} \, \k_{\btau'}^{\balpha'} }_{\k_{\btau}, \k_{\btau'} | \x}
\>_\x
=
\< k_{\tau_{1}}^{\alpha_1} \, k_{\tau_{2}}^{l - \alpha_1} \, k_{\tau_{1}'}^{\alpha_1'} \, k_{\tau_{2}'}^{s - \alpha_1'} \>_{\k_{\btau}, \k_{\btau'}} \ ,
\end{equation}
and therefore $\E{ \E{r_{\btau}} \E{r_{\btau'}}  } = \E{ \E{ r_{\btau} \, r_{\btau'} } }$ whenever $\btau \neq \btau'$.

Wick's theorem is again instrumental in computing $\langle \k_{\btau}^{\balpha} \k_{\btau'}^{\balpha'} \rangle$,
as all possible pairings of the random variables $\k_{\btau} = (k_{\tau_{1}}, k_{\tau_{2}})$ and $\k_{\btau'} = (k_{\tau_{1}'}, k_{\tau_{2}'})$
need to be included.
As $\E{ k_{\tau_{1}}^{2} } = 0$, $\E{ k_{\tau_{1}} k_{\tau_{2}} } = 0$, $\E{ k_{\tau_{1}'}^{2} } = 0$, and $\E{ k_{\tau_{1}'} k_{\tau_{2}'} } = 0$, the only non-zero expectations in the Wick expansion of Equation (\ref{eq:localk}) occur when \emph{all} the variables in $\k_{\btau}$ and $\k_{\btau'}$ are paired.
This immediately means that $\E{ k_{\tau_{1}}^{\alpha_1} \, k_{\tau_{2}}^{l - \alpha_1} \, k_{\tau_{1}'}^{\alpha_1'} \, k_{\tau_{2}'}^{s - \alpha_1'} } = 0$ whenever $l \neq s$, as there will be some remaining variables in $\k_{\btau}$ (or $\k_{\btau'}$) that can't be paired and have to be self-paired with zero expectation.

Given $l = s$, evaluate the expectation in Equation (\ref{eq:localk}).
We introduce the ``pairing count'' vector $\bbeta$ with elements $\beta_{j} \in \mathbb{N}_{0}$ and constraint $\sum_{j=1}^{4} \beta_{j} = l$.
Let $\beta_1$ count the number of pairings of $k_{\tau_1}$ with $k_{\tau_1'}$, and $\beta_2$ count the number of pairings of $k_{\tau_1}$ with $k_{\tau_2'}$. As there are $\alpha_1$ $k_{\tau_1}$ terms,
the sum of its outgoing pairings should equal $\alpha_1$ with
\[
\beta_{1} + \beta_{2} = \alpha_1 \ .
\]
A furthermore requirement is that
\[
\beta_{1} + \beta_{3} = \alpha_1' \ , \quad
\beta_{3} + \beta_{4} = \alpha_2 \ , \quad
\beta_{2} + \beta_{4} = \alpha_2' \ ,
\]
where $\alpha_2 = l - \alpha_1$ and $\alpha_2' = l - \alpha_1'$, and $\beta_3$ and $\beta_4$ be as in the Wick expansion below.
Define $\mathcal{B}$ to be the set of all such $\bbeta$'s, and let $\mathcal{C}(\bbeta)$ count the number of permuted configurations for a
given pairing $\bbeta$. From Wick's theorem the expected value is equal to the sum over all possible pairings $\bbeta$:
\[
\< k_{\tau_{1}}^{\alpha_1} \, k_{\tau_{2}}^{\alpha_2} \, k_{\tau_{1}'}^{\alpha_1'} \, k_{\tau_{2}'}^{\alpha_2'} \>_{\k_{\btau}, \k_{\btau'}}
= \sum_{\bbeta \in \mathcal{B}} \mathcal{C}(\bbeta) \,
\E{ k_{\tau_{1}} \, k_{\tau_{1}'} }^{\beta_{1}}
\E{ k_{\tau_{1}} \, k_{\tau_{2}'} }^{\beta_{2}}
\E{ k_{\tau_{2}} \, k_{\tau_{1}'} }^{\beta_{3}}
\E{ k_{\tau_{2}} \, k_{\tau_{2}'} }^{\beta_{4}} \ .
\]
A simple scheme to enumerate all $\bbeta \in \mathcal{B}$ is to let
\[
\bbeta = \Big[ \, \beta_1, \ \alpha_1 - \beta_1, \ \alpha_1' - \beta_1, \ ( l + \beta_1) - (\alpha_1 + \alpha_1') \, \Big] ,
\]
so that $\bbeta \in \mathcal{B}$ for each $\beta_1 \in \{ \max(0, (\alpha_1 + \alpha_1') - l), \ldots, \min(\alpha_1,\alpha_1') \}$. The remaining components of $\bbeta$ are uniquely determined from $\beta_1$.

\subsubsection{Counting Pairings}

How many permuted pairings $\mathcal{C}(\bbeta)$ are there?
\begin{enumerate}
\item There are $\binom{ \alpha_1 }{ \beta_1 }$ ways of choosing $\beta_1$ $k_{\tau_{1}}$'s, and then $\frac{\alpha_1'!}{(\alpha_1' - \beta_1)!}$ ways of choosing $k_{\tau_{1}'}$ to pair with.
\item This leaves a remaining $(\alpha_1 - \beta_1)$ $k_{\tau_{1}}$'s, that need to be paired with $(l - \alpha_1')$ $k_{\tau_{2}'}$'s.
There are $\frac{(l - \alpha_1')!}{((l - \alpha_1') - (\alpha_1 - \beta_1))!}$ such pairings.
\item There are also $\alpha_1' - \beta_1$ remaining $k_{\tau_{1}'}$'s, that need to be paired with $k_{\tau_{2}}$ variables.
There are $\binom{l - \alpha_1}{\alpha_1' - \beta_1}$ ways of picking a $k_{\tau_{2}}$, and a further $(\alpha_1' -\beta_1)!$ ways of arranging the remaining $k_{\tau_{1}'}$.
\item Finally, the $(l - \alpha_1') - (\alpha_1 - \beta_1)$ remaining $k_{\tau_{2}}'s$ need to be coupled with the remaining $k_{\tau_{2}'}$'s, and there are $((l - \alpha_1') - (\alpha_1 - \beta_1))!$ such arrangements.
\end{enumerate}
Multiplying the possible pairings from the four steps above gives
\begin{align*}
\mathcal{C}(\bbeta)
& = \binom{ \alpha_1 }{\beta_1} \ \frac{\alpha_1'!}{(\alpha_1' - \beta_1)!} \
\frac{(l - \alpha_1')!}{((l + \beta_1) - (\alpha_1 + \alpha_1'))!} \cdots \\
& \quad\quad \cdots \times
\binom{l - \alpha_1}{ \alpha_1' -\beta_1} \ (\alpha_1' - \beta_1)! \
((l + \beta_1) - (\alpha_1 + \alpha_1'))! \\
& = \binom{ \alpha_1 }{\beta_1} \ \alpha_1'! \
(l - \alpha_1')! \
\binom{l - \alpha_1}{\alpha_1' - \beta_1} \ ,
\end{align*}
which adds up to the total number of possible pairings $\sum_{\bbeta \in \mathcal{B}} \mathcal{C}(\bbeta) = l!$.
A further useful simplification is $\mathcal{C}(\bbeta) / \balpha! \balpha'! = 1 / \bbeta!$ when $|\balpha| = |\balpha'| = l$, and is used below.

\subsubsection{Edge-edge Expectation}

The absence of any self-interacting loops from Wick's theorem
lets the $\sum_{s \ge 3}$ drop away in Equation (\ref{eq:rr}), as all terms are zero except for when $l = s$. Substituting $\E{ \k_{\btau}^{\balpha} \k_{\btau'}^{\balpha'} }$ and $\mathcal{C}(\bbeta)$ into Equation (\ref{eq:rr}) gives the final result,
\begin{align*} 
& \< \< r_{\btau} (\k_{\btau}) \> \< r_{\btau'} (\k_{\btau'}) \> \>
\\
& \qquad =
\sum_{l \ge 3} (-1)^{l}
\sum_{|\balpha| = l}
\sum_{|\balpha'| = l}
c_{\balpha \btau } \, c_{\balpha' \btau'} \left\{
\sum_{\bbeta \in \mathcal{B}}
\frac{1}{\bbeta!}
\E{ k_{\tau_{1}} \, k_{\tau_{1}'} }^{\beta_1}
\E{ k_{\tau_{1}} \, k_{\tau_{2}'} }^{\beta_2}
\E{ k_{\tau_{2}} \, k_{\tau_{1}'} }^{\beta_3}
\E{ k_{\tau_{2}} \, k_{\tau_{2}'} }^{\beta_4}
\right\} \ . 
\end{align*}

\subsection{Edge-Node Expectations}

The derivation for the edge-node expectations is similar to that of the edge-edge case,
\begin{align*}
\< \< r_{\btau} (\k_{\btau}) \> \< r_n (k_n) \> \>
& = \< \sum_{l \ge 3} \sum_{s \ge 3} i^{l + s}
\sum_{|\balpha| = l} \, \frac{ c_{\balpha \btau} \, c_{s n} }{ \balpha ! \, \, s! } \,
\< \k_{\btau}^{\balpha} \>_{ \k_{\btau} | \x }
\< k_n^{s} \>_{k_n | \x } \>_\x \\
& = \sum_{l \ge 3} (-1)^l
\sum_{|\balpha| = l} \, \frac{ c_{\balpha \btau} \, c_{l n} }{ \balpha ! } \, \< k_{\tau_1} k_n \>^{\alpha_1} \< k_{\tau_2} k_n \>^{l - \alpha_1} \ ,
\end{align*}
where the expectations in the last line are again over $\{ \k_{\btau}, k_n \}$.
When $\E{ \k_{\btau}^{\balpha} \, k_n^s }$ is evaluated with Wick's theorem,
there are $\alpha_1$ copies of $k_{\tau_1}$, $l - \alpha_1$ copies of $k_{\tau_2}$,
and $s$ copies of $k_n$.
The zero relation of $\k_{\btau}$ and $k_n$ ensures that the only non-zero terms in the Wick sum are those where all the $k_\tau$'s are paired with $k_n$'s; in other words, when $l = s$. There are $l!$ possible pairings, which cancels $l!$ in the denominator.

The above edge-node expectation is for any edge and node in the tree, but
notice that it simplifies greatly when the edge $\btau$ is a connection to node $n$.
Say $\tau_1$ is the edge variable corresponding to $x_n$.
In this case  the covariance with respect to the \emph{opposite} pair is zero, with
$\E{ k_{\tau_2}, k_n } = 0$ (see Figure \ref{fig:kcov})
and only \emph{one} of the $\balpha$'s  will have a non-zero contribution to the sum, namely when $\balpha = (l,0)$.

\subsection{Node-Node Expectations}

The node-node expectation is given in Equation (\ref{eq:r-cumulants}), and is also used for $\E{ \E{ r_{n} }^2 }$.\footnote{Due to the square in
$\E{ \E{ r_{n} }_{k_n | \x}^2 }_\x$,
the inner average $\E{ r_{n} }_{k_n | \x}$ should first be computed to
give an expansion over Hermite polynomials in $x_n - \mu_n$.
An example of such a result is given Appendix \ref{sec:one-dim-example}.
The orthogonality of these polynomials over $q(\x - \bmu)$ allows $\E{ \E{ r_{n} }_{k_n | \x}^2 }_\x$ to also reduce to Equation (\ref{eq:r-cumulants}).}

\section{A Tractable, One-Dimensional Example} \label{sec:one-dim-example}

The following example illustrates a tractable one-dimensional model with two factors. It is shown analytically that the correction to $\log Z_{\EP}$ must be zero, and that
the result is reflected in the higher-order terms in Equation (\ref{eq:generalLogR}), which are also zero.

Consider the factorization of a probit term with a Gaussian prior into
\[
p(x) = \frac{1}{Z} \, \Phi(x) \, {\cal N}(x  ; 0, 1) =
\frac{1}{Z} \, f_a(x)^{1/2} f_b(x)^{1/2} {\cal N}(x  ; 0, 1) \ ,
\]
where $\Phi(x)$ is the cumulative Gaussian density function, and
$f_a(x) = f_b(x) = \Phi(x)$. $Z$ can be computed exactly, but for the sake of example $p(x)$ will be approximated with
\[
q(x) = \frac{1}{Z_q} \, g_a(x)^{1/2} g_b(x)^{1/2} {\cal N}(x  ; 0, 1) =
{\cal N}(x ; \mu, \sigma^{2}) \ .
\]
Choose $g_a(x) = \exp \{ \phi(x)^T \blambda_a \}$, and $g_b(x) = \exp \{ \phi(x)^T \blambda_b \}$.
The $q$ approximation has parameter vector $\blambda = \blambda_0 + \frac{1}{2} \blambda_a + \frac{1}{2} \blambda_b$.
The EP fixed point is defined by
$\blambda_a = \blambda_b$ and $Z_a = Z_b$.
(For example, subtracting $\blambda_a$ at the fixed point will leave
$\blambda_{\wo a} = \blambda_0 + \mathbf{0}$, which is equal to a scaled version of the prior $f_0(x)$.
The factor $f_a(x) = \Phi(x)$ is hence incorporated into the prior, giving $Z_a$. By a symmetric argument, $Z_a = Z_b$.)
Although it is trivial to show that
$Z_{\EP} = Z_q Z_a^{1/2} Z_b^{1/2}$ will be equal to the true partition function $Z$, we shall prove it by showing that the correction term is
$\log R = 0$.

\subsection{Analytic Correction} \label{sec:one-dim-analytic}

In this section a transformation of variables from $x$ to $y \sim {\cal N}(y;0,1)$, with $y = (x - \mu) / \sigma$, will be used to make the derivation slightly simpler, and therefore
\[
k_a | y \sim {\cal N} \left( k_a \, ; \, - \frac{i y}{\sigma} , \, \sigma^{-2} \right) \ ,
\qquad
k_b | y \sim {\cal N} \left( k_b \, ; \, - \frac{i y}{\sigma} , \, \sigma^{-2} \right) \ .
\]
Below we analytically show that the correction $\log R$ is zero, and hence that
\begin{equation} \label{eq:simple-R}
R = \< \< \erm^{r_a(k_a)} \>_{k_a | y}^{1/2} \< \erm^{r_b(k_b)} \>_{k_b | y}^{1/2} \>_y =
\< \sqrt{ {\cal F}_a(y) } \sqrt{ {\cal F}_b(y) } \>_y =
1 \ ,
\end{equation}
where ${\cal F}_a(y)$ is a shorthand for $\E{ \erm^{r_a(k_a)} }_{k_a | y}$
and
\[
r_a(k_a) = \sum_{l \ge 3} i^l \frac{c_{al}}{l!} k_a^l \ , \qquad 
r_b(k_b) = \sum_{l \ge 3} i^l \frac{c_{bl}}{l!} k_b^l \ .
\]
Because $f_a = f_b$, the cumulants will be the same for all $l$, hence $c_{al} = c_{bl}$.
Furthermore, $k_a | y$ and $k_b | y$ are both distributed according to the \emph{same} density.
Now define, using $\erm^{r_a} = 1 + r_a + \frac{1}{2}r_a^2 + \cdots$,
\begin{align}
{\cal F}_a(y)
& =
\<
1 + \sum_{l \ge 3} i^l \frac{c_{al}}{l!} k_a^l
+ \frac{1}{2} \sum_{l,s \ge 3} i^{l+s} \frac{c_{al} c_{as} }{l! s!} k_a^{l+s} + \cdots
\>_{k_a | y} \nonumber \\
& =
\<
1 + \sum_{l \ge 3} \frac{c_{al}}{l!} \left( \frac{1}{\sigma}\right)^l (y + i u)^l
+ \frac{1}{2} \sum_{l,s \ge 3} \frac{c_{al} c_{as} }{l! s!}
\left( \frac{1}{\sigma}\right)^{l+s} (y + i u)^{l+s} + \cdots
\>_{u} \nonumber \\
& =
1 + \sum_{l \ge 3} \frac{c_{al}}{l!} \left( \frac{1}{\sigma}\right)^l
H_{l} (y)
+ \frac{1}{2} \sum_{l,s \ge 3} \frac{c_{al} c_{as} }{l! s!}
\left( \frac{1}{\sigma} \right)^{l+s} H_{l+s} (y) + \cdots \label{eq:hermites}
\end{align}
In the second line above a transformation of variables was made in the integral, with $u = \sigma k_a + i y$, such that $k_a = (u - i y) / \sigma$. The Jacobian $1 / \sigma$ ensures proper normalization so that the average is over $u \sim {\cal N}(u; 0, 1)$.
In the last line
$H_l(y)$ is the \emph{Hermite polynomial} of degree $l$,
\begin{align*}
H_0(y) & = 1  \ , & H_1(y) & = y  \ , & H_2(y) & = y^2 - 1  \ , \\
H_3(y) & = y^3 - 3y  \ , & H_4(x) & = y^4 - 6 y^2 + 3  \ , & H_5(y) & = y^5 - 10 y^3 + 15 y \quad \cdots
\end{align*}
which can be obtained for any real $y$ and integer $l = 0, 1, 2, \ldots$ from the average $H_l(y) = \< (y + i u)^l\>_u$ over $u \sim {\cal N}(u; 0, 1)$.\footnote{When ${\cal F}(y)$ in Equation (\ref{eq:hermites}) is rearranged as a power series in $\sigma^l$, we obtain an Edgeworth expansion to arbitrary order $l$. The deviation from the Gaussian $q(y)$ is thereby factorized out of tilted distribution with $q_a(y) = q(y) {\cal F}(y)$. The interested reader is pointed to \cite{blinnikov98expansions}.
}

The remarkable property $ \< H_l(y) \>_y = 0$ for all $l$, ensures that
$\< {\cal F}_a(y) \>_y = 1$ in Equation (\ref{eq:hermites}).
Furthermore, ${\cal F}_a(y) = {\cal F}_b(y)$ follows from the equivalence in cumulants $c_{al} = c_{bl}$;
the roots in Equation (\ref{eq:simple-R}) disappear to give $\< {\cal F}_a(y) \>_y$, proving that $R = 1$ in Equation (\ref{eq:simple-R}).

\subsection{Second Order Correction}

The second order expansion in Equation (\ref{eq:generalLogR}) in Section \ref{sec:general-approximations} evaluates to zero, as the matching cumulants
$c_{al} = c_{bl}$ and equal distributions of $k_a | x$ and $k_b | x$ ensure that $\< r_a(k_a) \>_{k_a | x} =  \< r_b(k_b) \>_{k_b | x}$:
\begin{align*}
\log R & = \frac{1}{4} \< \< r_a(k_a) \>_{k_a | x} \< r_b(k_b) \>_{k_b | x} \>_x - \frac{1}{8} \left( \< \< r_a(k_a) \>_{k_a | x}^2 \>_x
+ \< \< r_b(k_b) \>_{k_b | x}^2 \>_x \right) + \cdots \\
& = \frac{1}{4} \< \< r_a(k_a) \>_{k_a | x}^{2} \>_x - \frac{1}{8} \left( 2  \< \< r_a(k_a) \>_{k_a | x}^2 \>_x \right) + \cdots
\\
& = 0 + \cdots \ .
\end{align*}

\section{Corrections to Marginals Distributions} \label{sec:marginals}

Corrections to the marginal distributions follow from a similar derivation to that of the normalizing constant.
As a simplification, let the Gaussian approximation be centred with $\y = \x - \bmu$, so that $q(\y) = {\cal N}(\y \, ; \, \mathbf{0}, \bSigma)$,
and assume that $q(\x)$ is arises from the fully factorized approximation in Section \ref{sec:factorized}.
In this appendix corrections will be computed for the mean $\< x_i - \mu_i\>_{p(\x)} = \< y_i \>_{p(\y)}$,
and variance $\< (x_i - \mu_i)(x_j - \mu_j) - \Sigma_{ij} \>_{p(\x)} = \< y_i y_j \>_{p(\y)} - \Sigma_{ij}$.

A further simplification that will be employed in the following section is a change of variables
$\eta_n = k_n + i \Sigma_{nn}^{-1} y_n$, so that $\eta_n \sim {\cal N} (\eta_n \, ; \, 0, \, \Sigma_{nn}^{-1})$.
Let
\[
z_n = \eta_n - i \Sigma_{nn}^{-1} y_n \ ,
\]
which is zero-mean complex Gaussian random variable with a relation $\< z_n^2 \> = 0$
and $\< z_m z_n \> = - \Sigma_{mn} / (\Sigma_{mm} \Sigma_{nn})$ when $m \neq n$.
Following Equation (\ref{eq:factorizedR}), the correction reads
\[
R = \< \prod_n \BigE{ r_n (k_n) }_{k_n | y_n} \>_\y =
\< \prod_n \< r_n \Big( \eta_n - i \Sigma_{nn}^{-1} y_n \Big)  \>_{\eta_n} \>_\y =
\< \exp \left[ \sum_n r_n (z_n) \right] \>_{\z} \ .
\]

\subsection{The Marginal Mean} \label{sec:marginalmeancorrection}

The lowest order correction to the EP marginal's mean follows from the result in Equation (\ref{eq:exact}):
\begin{align*}
\< y_i \>_{p(\y)} & = \frac{1}{R} \< y_i\; \erm^{\sum_n r_n(z_n)} \>_{\z}  \\
& = \frac{1}{R} \sum_j \Sigma_{ij} \< \frac{\partial}{\partial y_j}\;  \erm^{\sum_n r_n(z_n)} \> \\
& = \frac{1}{R} \sum_j \Sigma_{ij} \< \frac{\partial}{\partial y_j}\left( 1 + \sum_n r_n(z_n) + \frac{1}{2} \sum_{m,n} r_m(z_m) r_n(z_n) + \cdots \right) \> \\
& = \frac{1}{R} \sum_j \Sigma_{ij} \< \frac{\partial r_j(z_j)}{\partial y_j} + \sum_n r_n(z_n) \frac{\partial r_j(z_j)}{\partial y_j} + \cdots \> \ .
\end{align*}
In the above expansion the first order term is
$\frac{\partial r_j(z_j)}{\partial y_j}  = \frac{\partial r_j(z_j)}{\partial z_j} \frac{\partial z_j}{\partial y_j} = -i \Sigma_{jj}^{-1} \frac{\partial r_j(z_j)}{\partial z_j}$,
and disappears as $ \< \frac{\partial r_j(z_j)}{\partial z_j} \> = 0$.
The $j=n$ second order term also disappears as $ \< r_j(z_j) \frac{\partial r_j(z_j)}{\partial z_j} \> = 0$.
These equivalences can be seen by taking $r_j(z_j)$ (and also its derivative) as a expansion over powers of $z_j$;
as $\E{ z_j^2 } = 0$, Wick's theorem states that every expectation of powers of $z_j$ should be zero. Hence
\begin{equation} \label{eq:marginalcorrection-almosthere}
\< y_i \>_{p(\y)} = - \frac{i}{R} \sum_{j \neq n} \frac{\Sigma_{ij}}{\Sigma_{jj}} \< r_{n}(z_n) \frac{\partial r_j(z_j)}{\partial z_j} \>_{\z} + \cdots \ .
\end{equation}
The derivative of the characteristic function, as required in Equation (\ref{eq:marginalcorrection-almosthere}), is
\[
\frac{\partial r_j(z_j)}{\partial z_j}
= \frac{\partial}{\partial z_j} \left[ \sum_{l \ge 3} i^l \frac{c_{lj}}{l!} z_j^l \right]
= i \sum_{l \ge 3} i^{l-1} \frac{c_{lj}}{(l - 1)!} z_j^{l-1}
= i \sum_{l \ge 2} i^l \frac{c_{l+1,j}}{l!} z_j^{l} \ .
\]
The expectations for $j \neq n$ in Equation (\ref{eq:marginalcorrection-almosthere}) evaluate to
\begin{align}
 \< r_{n}(z_n) \frac{\partial r_j(z_j)}{\partial z_j} \>_{\z} & =
 i \sum_{s, l \ge 3} i^{s+l} \frac{c_{l+1,j}, c_{sn} }{l!s!} \<z_j^l z_n^s \>
+ i \sum_{s \ge 3, l = 2} i^{s + l} \frac{c_{l+1,j}}{2! s!} \< z_j^{l} z_n^s \> \nonumber \\
& = i  \sum_{l\geq 3} i^{2l} \frac{c_{l+1,j} c_{ln}} {(l!)^2} \< z_j^l z_n^l \> \ , \label{eq:expectationOfZ}
\end{align}
with the second term disappearing as $s > l = 2$ ensures that some $z_n$ is always self-paired in Wick's theorem.
Finally, by substituting Equation (\ref{eq:expectationOfZ}) into (\ref{eq:marginalcorrection-almosthere}),
 the correction to the mean is
\[
\< y_i \>_{p(\y)} = \sum_{l\geq 3} \sum_{j\neq n}
\frac{\Sigma_{ij}}{\Sigma_{jj}} \frac{c_{l+1,j} c_{ln}} {l!} \left(\frac{\Sigma_{j n}}{\Sigma_{jj}\Sigma_{nn}} \right)^l \pm \cdots \ .
\] 

\subsection{The Marginal Covariance}

The correction to the second moments follow the same recipe as that of the marginal mean in Appendix \ref{sec:marginalmeancorrection}.
We proceed by first treating $y_i$ with
\begin{align*}
\< y_i y_j \>_{p(\y)} & = \frac{1}{R} \< y_i \left\{ y_j \; \erm^{\sum_n r_n(z_n)} \right\} \>_{\z}  \\
& = \frac{1}{R} \sum_k \Sigma_{ik} \< \frac{\partial}{\partial y_k} \left\{ y_j \;  \erm^{\sum_n r_n(z_n)} \right\} \> \\
& = \frac{1}{R} \sum_k \Sigma_{ik} \< \delta_{jk} \; \erm^{\sum_n r_n(z_n)} + y_j \frac{\partial}{\partial y_k} \erm^{\sum_n r_n(z_n)} \> \\
& = \Sigma_{ij} + \frac{1}{R} \sum_k \Sigma_{ik} \< y_j \frac{\partial}{\partial y_k} \erm^{\sum_n r_n(z_n)} \> \ .
\end{align*}
Reapplying the recipe gives the correction to the covariance:
\begin{align*}
\< y_i y_j \>_{p(\y)} - \Sigma_{ij} & = \frac{1}{R} \sum_{kl} \Sigma_{il} \Sigma_{jk} \< \frac{\partial^2}{\partial y_k \partial y_l}\;  \erm^{\sum_n r_n(z_n)} \>_{\z} \\
& = - i \sum_{kl} \frac{\Sigma_{il}}{\Sigma_{ll}} \Sigma_{jk}
\< \frac{\partial}{\partial y_k }
\frac{\partial r_l(z_l)}{\partial z_l } \erm^{\sum_n r_n(z_n)} \> + \cdots
\\
& = - \sum_{kl} \frac{\Sigma_{il}}{\Sigma_{ll}} \frac{\Sigma_{jk}}{\Sigma_{kk}}
\< \left[ \delta_{kl} \frac{\partial^2 r_l(z_l)}{\partial z_l^2 } +
\frac{\partial r_k(z_k)}{\partial z_k } \frac{\partial r_l(z_l)}{\partial z_l }
 \right] \erm^{\sum_n r_n(z_n)} \>
\\
& = \sum_{s\geq 3} \sum_{k\neq l} \frac{\Sigma_{il}\Sigma_{jl}}{\Sigma_{ll}^2} \frac{c_{sk} c_{s+2,l}} {s!}
\left(\frac{\Sigma_{kl}}{\Sigma_{kk}\Sigma_{ll}} \right)^s \\
& \qquad \qquad +
\sum_{s\geq 3} \sum_{k\neq l}
\frac{\Sigma_{il}}{\Sigma_{ll}}\frac{\Sigma_{jk}}{\Sigma_{kk}} \frac{c_{sk} c_{sl}} {s!}
\left(\frac{\Sigma_{kl}}{\Sigma_{kk}\Sigma_{ll}} \right)^{s-1} + \cdots \ .
\end{align*}

\section{Higher Order Cumulants} \label{app:cumulants}

Much of this paper hinges on cumulants beyond the second order. These are frequently more cumbersome to obtain than the initial moments that are required by EP. This appendix provides details of the cumulants used in this paper.

The cumulants of a distribution $q_n(x)$ can be obtained from its moments through
\begin{align*}
c_3 & = \< x^3 \> - 3 \< x^2 \> \< x \> + 2 \< x \>^3 \ , \\
c_4 & = \< x^4 \> - 4 \< x^3 \> \< x \> - 3 \< x^2 \>^2 + 12 \< x^2 \> \< x \>^2 - 6 \< x \>^4 \ , \\
c_5 & = \< x^5\> -  5 \< x^4 \> \< x \> -  10 \< x^3 \> \< x^2 \> + 20 \< x^3 \> \< x \>^2 + 30 \< x^2 \>^2 \< x \> - 60 \< x^2 \> \< x \>^3 + 24 \< x\>^5  \, ;
\end{align*}
they are derived for doubly-truncated Gaussian distributions in Appendices \ref{sec:gpbox-cumulants} and \ref{sec:uniform-cumulants}. One might also directly take derivatives of the cumulant generating function, and the cumulants of a Probit-times-Gaussian distribution, common to GP classification models, are derived this way in Appendix \ref{sec:gpc-cumulants}.

The tree-structured approximation in Sections \ref{sec:general-approximations} and \ref{WJ},
and Appendices \ref{sec:treestructured-basics} and \ref{app:tree}, require cumulants over two variables. They are presented in Appendix \ref{sec:two-variable-ising} for the Ising model.

\subsection{Doubly Truncated Centered Gaussian} \label{sec:gpbox-cumulants}

Consider the centered distribution $q_n(x_n) \propto \mathbb{I}[|x_n| < a] \, {\cal N}(x_n \, ; \, 0, \, \lambda_n^{-1})$.
The odd moments of this tilted distributions are, by symmetry, $\< x_n \> = \< x_n^3\> = \< x_n^5\> = 0$. Let
\[
Z_n = 2 \sqrt{ \frac{\lambda}{2 \pi}} \int_0^a \erm^{-\frac{1}{2} \lambda x^2} \, \mathrm{d} x = 2 \Phi ( z ) - 1 \ , \qquad z = \sqrt{\lambda} a \ ,
\]
with the Probit function being  $\Phi(x) = \int_{-\infty}^{x} {\cal N} ( z  ;  0, 1) \, \mathrm{d}z$.
Subscripts $n$ are dropped where they are clearly implied by their context.
To get the even moments, consider
\begin{align*}
A_1 & = \partial_{\lambda} \log Z_n = \partial_{\lambda} \log \left( \sqrt{\lambda} \int_{-a}^{a} d x \, \erm^{-\frac{1}{2} \lambda x^2} \right) = \frac{1}{2 \lambda} - \frac{1}{2} \< x^2 \> \ ,
\\
A_2 & = \partial_{\lambda}^2 \log Z_n = - \frac{1}{2 \lambda^2} + \frac{1}{4} \Big( \< x^4\> - \< x^2\>^2 \Big) \ .
\end{align*}
Using the partition function, we get
\begin{align*}
A_1 & = \partial_{\lambda} \log \left( 2 \Phi (z) - 1 \right) = \frac{a}{\sqrt{\lambda}}
\left( \frac{ {\cal N} (z) }{ 2 \Phi (z) - 1} \right) \ ,
\\
A_2 & = \frac{a^2}{2 \lambda}
\left( \frac{ z \, {\cal N} (z) }{ 2 \Phi (z) - 1} \right)
- \frac{a}{2 \lambda^{3/2}} \left( \frac{ {\cal N} (z) }{ 2 \Phi (z) - 1} \right)
- \frac{a^2}{\lambda} \left( \frac{ {\cal N} (z) }{ 2 \Phi (z) - 1} \right)^2 \ ,
\end{align*}
and thus
\begin{align*}
\< x^2 \> & = \frac{1}{\lambda} - 2 A_1 \ , \\
\< x^4 \> & = \frac{2}{\lambda^2} + \< x^2 \>^2 + 4 A_2 \ .
\end{align*}
We can further determine $A_3 = \partial_{\lambda}^3 \log Z_n$ using the partition function, giving
\begin{align*}
A_3 & =
\frac{3 a}{4 \lambda^{5/2}} \left( \frac{ {\cal N}(z) }{2 \Phi(z) - 1} \right)
+  \frac{3 a^2}{4 \lambda^2} \left( \frac{ z {\cal N}(z)}{2 \Phi(z) - 1} \right)
+ \frac{3 a^2}{ 2 \lambda^2} \left( \frac{ {\cal N}(z) }{2 \Phi(z) - 1} \right)^2
+ \frac{2 a^3}{\lambda^{3/2}}
\left( \frac{ {\cal N}(z) }{2 \Phi(z) - 1} \right)^3
\\
& \qquad \qquad +
\frac{3 a^3}{2 \lambda^{3/2}} \left( \frac{ {\cal N}(z) }{2 \Phi(z) - 1} \right) \left( \frac{ z {\cal N}(z) }{2 \Phi(z) - 1} \right)
+
\frac{a^3}{4 \lambda^{3/2}} \left( \frac{(z^2 - 1) {\cal N}(z)}{2 \Phi(z) - 1} \right) \ .
\end{align*}
Therefore
\[
\< x^6 \>
= \frac{8}{\lambda^3}
+  \< x^2 \> \< x^4 \>
- 2 \< x^2 \>^3 + 2 \< x^4 \> \< x^2 \> - 8 A_3 \ .
\]

\subsection{Doubly Truncated Non-Centered Gaussian} \label{sec:uniform-cumulants}

\begin{figure}[t]
\centering
\epsfig{file=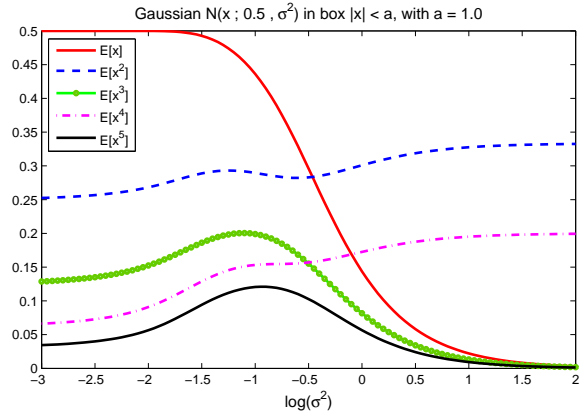, width=0.6\linewidth}
\caption{The moments of $q_n(x) \propto \mathbb{I}[|x| < a] \, {\cal N}(x \, ; \, \mu, \, \sigma^2)$, as a function of $\sigma^2$. As the Gaussian variance $\sigma^2 \to \infty$, the moments converge to that of a uniform ${\cal U}[-a, a]$ distribution.} \label{fig:boxmoments}
\end{figure}

The same calculation from Appendix \ref{sec:gpbox-cumulants} can be repeated to get the moments of the non-centered truncated Gaussian $q_n(x_n) \propto \mathbb{I}[|x_n| < a] \, {\cal N}(x_n \, ; \, \mu, \, \lambda_n^{-1})$.
The subscripts $n$ are dropped where evident. The partition function is
\[
Z(\lambda, \mu) = \sqrt{ \frac{\lambda}{2 \pi} }  \int_{-a}^{a} \erm^{ - \frac{1}{2} \lambda (x - \mu)^2} \, \mathrm{d} x = \Phi ( z_{\max} ) - \Phi(z_{\min}) \ ,
\]
where
\[
z_{\max} = \sqrt{\lambda}(\mu + a) \ , \quad z_{\min} = \sqrt{\lambda} (\mu - a) \ .
\]
By again taking increasing derivatives of $Z(\lambda, \mu)$ with respect to $\mu$ and $\lambda$, the moments solved for are
\begin{align*}
\< x \> & = \mu + \frac{1}{\sqrt{\lambda}} \frac{  {\cal N} ( z_{\max} ) - {\cal N} ( z_{\min} ) }{  \Phi ( z_{\max} ) - \Phi ( z_{\min} ) } \ ,
\\
\< x^2\> & = 2 \<x \> \mu + \frac{1}{\lambda}   - \mu^2 -
\frac{1}{\lambda} \frac{  z_{\max} {\cal N} ( z_{\max} ) - z_{\min} {\cal N} ( z_{\min} )  }{  \Phi ( z_{\max} ) - \Phi ( z_{\min} )  }
\ , \\
\< x^3\> & = 3 \< x^2\> \mu + \<x\> \left[ \frac{3}{\lambda} - 3 \mu^2  \right] - \frac{3}{\lambda} \mu + \mu^3
\ , \\
& \qquad \qquad - \frac{1}{\lambda^{3/2}} \, \frac{ (1 - z_{\max}^{\phantom{0}2} ) \, {\cal N} ( z_{\max} )  - (1 - z_{\min}^{\phantom{0}2} ) \, {\cal N} ( z_{\min} ) }{ \Phi ( z_{\max} ) - \Phi ( z_{\min}) }
\ , \\
\< x^4 \> & = 4 \< x^3 \> \mu + \< x^2 \> \left[ \frac{2}{\lambda} - 6 \mu^2 \right]  + \< x \> \left[4  \mu^3 - \frac{4}{\lambda}\mu \right]
+ \frac{2}{\lambda} \mu^2  - \mu^4 + \frac{1}{\lambda^2}
\nonumber \\
& \qquad \qquad
- \frac{1}{\lambda^2} \, \frac{
z_{\max} (1 + z_{\max}^2) \, {\cal N} ( z_{\max} ) -  z_{\min} (1 + z_{\min}^2) \, {\cal N} ( z_{\min} )
}{\Phi( z_{\max} ) - \Phi( z_{\min} )} \ .
\end{align*}
Finally,
\begin{align*}
\< x^5 \> & =
5  \< x^4 \> \mu
+ \< x^3 \> \left[ \frac{6}{\lambda} - 10 \mu^2  \right] + \< x^2 \> \left[10 \mu^3 - \frac{18}{\lambda} \mu \right]
+ \< x \> \left[ \frac{18}{\lambda} \mu^2 - 5 \mu^4 - \frac{3}{\lambda^2} \right]
+ \frac{3}{\lambda^2} \mu
\nonumber \\
& \quad  - \frac{6}{\lambda} \mu^3 + \mu^5
- \frac{1}{\lambda^{5/2}}
\, \frac{
(1 + 2 z_{\max}^2 - z_{\max}^4) \, {\cal N} ( z_{\max} ) -  (1 + 2 z_{\min}^2 - z_{\min}^4) \, {\cal N} ( z_{\min} )
}{\Phi( z_{\max} ) - \Phi( z_{\min} )}  \ .
\end{align*}
As Figure \ref{fig:boxmoments} illustrates, these moments will converge to that of a uniform distribution as the Gaussian's variance grows large.

\subsection{Probit Link Cumulants} \label{sec:gpc-cumulants}

\begin{figure}[t]
\centering
\epsfig{file=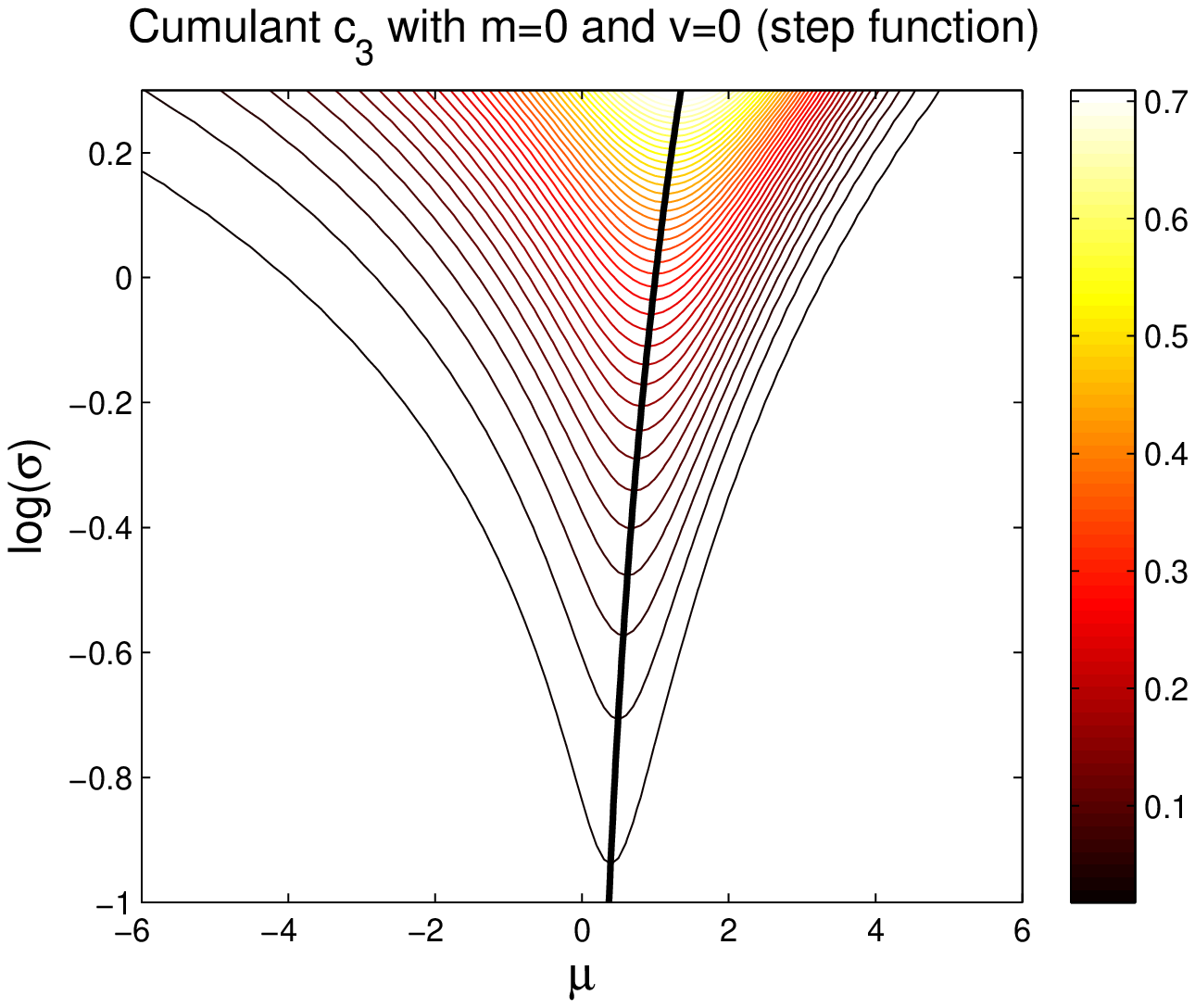, width=0.48\linewidth}
\epsfig{file=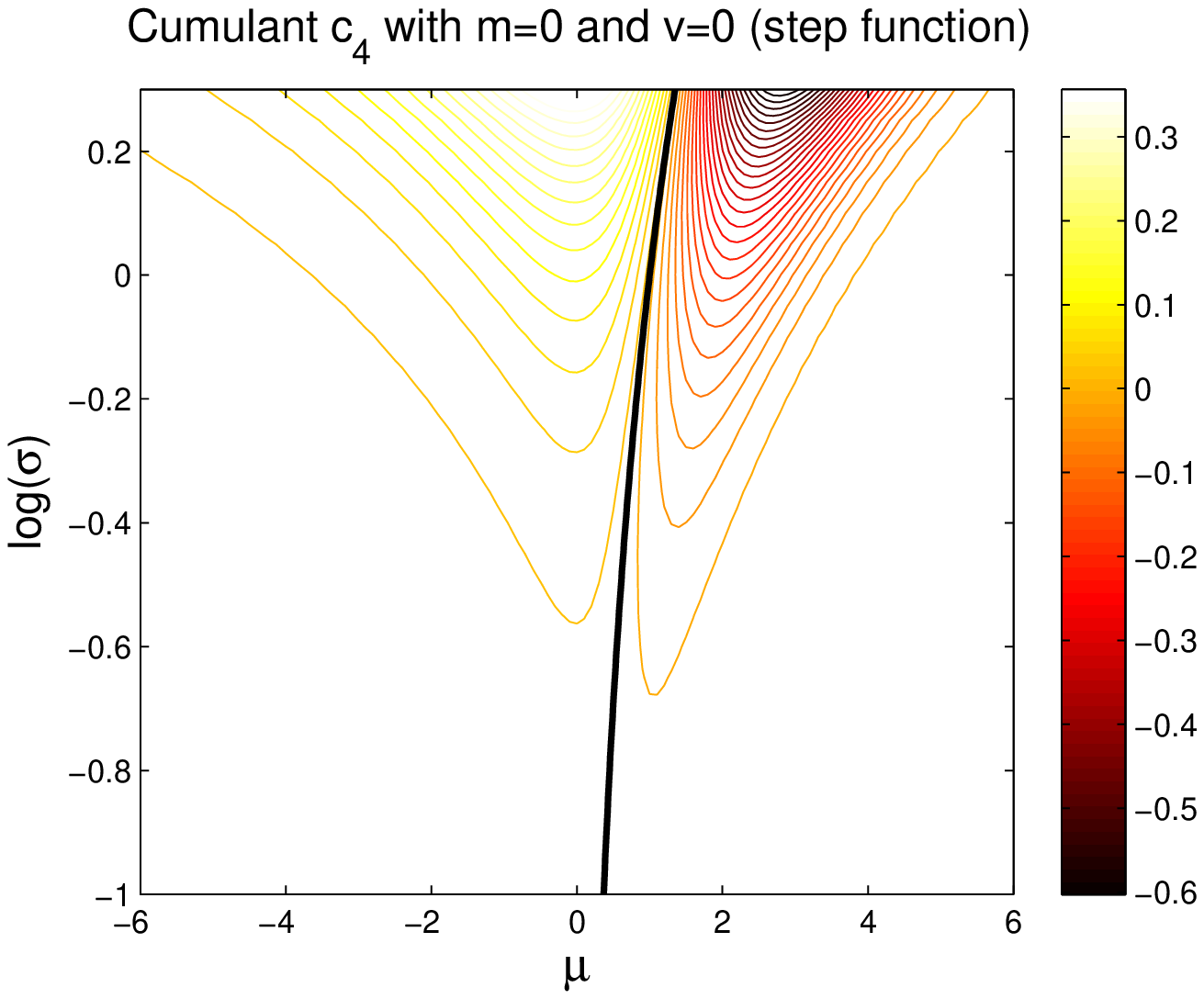, width=0.48\linewidth}
\caption{The third and fourth cumulants of the density $q_n(x) \propto \Phi((x - m) / v) \, {\cal N}(x; \mu, \sigma^2)$ in Appendix \ref{sec:gpc-cumulants}. The step function $\Theta(x)$, with $m = v = 0$, is taken as an example here. The third cumulant is always positive, while the fourth cumulant is positive only when $\sigma > \mu$.} \label{fig:gpc-cumulants}
\end{figure}

EP approximations to Probit regression models, and Gaussian process classification models in general (see Section \ref{sec:gpc-results}), depend on the moments of $q_{n}(x) \propto \Phi ( (x - m) / v ) \, {\cal N}(x ; \mu, \sigma^{2})$.
We introduce $v \ge 0$ so that the likelihood can become a step function at $v = 0$, for example.
We shall obtain the cumulants by taking derivatives of the characteristic function.
The characteristic function of $q_{n}(x)$, as described by Equation (\ref{eq:chiintermsofq}), is
\[
\chi_{n}(k) = \bigE{\erm^{i k x}}_{q_{n}(x)} = \exp\left\{i k \mu - \frac{1}{2} k^{2} \sigma^{2} \right\} \frac{ \Phi(z_{k}) }{ \Phi(z) } \ ,
\]
with
\[
z = \frac{\mu - m }{ \sqrt{v^2 + \sigma^{2}}} \ , \quad z_{k} = \frac{\mu + i k \sigma^{2} - m }{ \sqrt{v^2 + \sigma^{2}}} \ . \]
The cumulants $c_{ln}$ are determined
from the derivatives of $\log \chi_n(k)$ at zero; a lengthy calculation shows that they are
\begin{align*}
c_{3n} & =
\alpha^{3} \beta \big[ 2 \beta^{2} + 3 z \beta + z^{2} - 1 \big] \ , \\
c_{4n} & =
- \alpha^{4} \beta \big[ 6 \beta^{3} + 12 z \beta^{2} + 7 z^{2} \beta + z^{3} - 4 \beta - 3 z \big] \ ,
\end{align*}
where $\alpha = \sigma^{2} / \sqrt{v^2 + \sigma^{2}}$ and $\beta = \Ncal(z; 0, 1) / \Phi(z)$.

\subsection{Two-Variable Ising Model Cumulants} \label{sec:two-variable-ising}

We need some third and fourth order two-variable cumulants and thus generalize the results of Section \ref{sec:cumulants} to the bivariate case. To do this we can exploit the cumulant generating property of $\log \chi_a(\k_a)$. Let $c_{(l,l')}$ denote the joint $l, l'$ order cumulant of variable one and two, respectively. We can generate this cumulant from derivatives of $\log \chi_a(\k_a)$:
$$
c_{(l,l')} = \left. \left( \frac{\partial}{\partial ik_1} \right)^l  \left( \frac{\partial}{\partial ik_2} \right)^{l'} \log \chi_a(\k_a) \right|_{\k=\mathbf{0}} \ .
$$
We can also express this as a recursion in terms of cumulants:
$$
c_{(l+n,l'+n')}= \left. 
\left( \frac{\partial}{\partial ik_1} \right)^n  \left( \frac{\partial}{\partial ik_2} \right)^{n'} c_{(l,l')}(\k) 
\right|_{\k=\mathbf{0}} \ .
$$
By explicit calculation for a bivariate binary distribution we get the first two orders' cumulants: $c_{(1,0)} = m_1$,  $c_{(0,1)}=m_2$, $c_{(2,0)}=1-m_1^2$, $c_{(0,2)}=1-m_2^2$ and $c_{(1,1)}$ is equal to the covariance between the two variables (to be matched with $q(x)$). The fact that we can write $c_{(2,0)}$ in terms of the first order cumulant shows that we can express all order cumulants in terms of the first and second order cumulant for example:
$$
c_{(2,1)} = \left.
\frac{\partial}{\partial ik_2} c_{(2,0)}(\k)
\right|_{\k=\mathbf{0}} = \left.
\frac{\partial}{\partial ik_2} ( 1 - c_{(1,0)}^2(\k) ) \right|_{\k=\mathbf{0}} = - 2 c_{(1,0)} c_{(1,1)} \ . 
$$
Using the same recursion it is easy to show: $c_{(3,0)}=-2c_{(1,0)}c_{(2,0)}$,  $c_{(4,0)}=-2c^2_{(2,0)}-2c_{(1,0)}c_{(3,0)}$, 
$c_{(3,1)}=-2c_{(2,0)}c_{(1,1)}-2c_{(1,0)} c_{(2,1)}$ and $c_{(2,2)}=-2c^2_{(1,1)}-2c_{(1,0)} c_{(1,2)}=-2c^2_{(1,1)}+4c_{(1,0)}c_{(0,1)}c_{(1,1)}$.

\bibliography{bibliography}

\end{document}